\documentclass[sn-chicago,iicol]{sn-jnl}% Default with double column layout
\usepackage{graphicx}%
\usepackage{multirow}%
\usepackage{amsmath,amssymb,amsfonts}%
\usepackage{amsthm}%
\usepackage{mathrsfs}%
\usepackage[title]{appendix}%
\usepackage{xcolor}%
\usepackage{textcomp}%
\usepackage{manyfoot}%
\usepackage{booktabs}%
\usepackage{algorithm}%
\usepackage{algorithmicx}%
\usepackage{algpseudocode}%
\usepackage{listings}%
\usepackage[inline]{enumitem}
\usepackage[caption=false,font=footnotesize]{subfig}
\usepackage{tabularx}
\PassOptionsToPackage{hyphens}{url}\usepackage{hyperref}
\usepackage{url}
\usepackage{adjustbox}
\usepackage{multirow}
\usepackage{makecell}

\theoremstyle{thmstyleone}%
\theoremstyle{thmstyletwo}%

\theoremstyle{thmstylethree}%

\newcommand{\etal}{\textit{et al}.}

\newcommand{\eg}{\textit{e}.\textit{g}.}

\usepackage{xcolor}

\begin{document}

\title[Article Title]{Principles of Designing Robust Remote Face Anti-Spoofing Systems}


\author{\fnm{Xiang} \sur{Xu}}\email{xiangx@amazon.com}
\author{\fnm{Tianchen} \sur{Zhao}}\email{tianchz@amazon.com}
\author{\fnm{Zheng} \sur{Zhang}}\email{zhnzhe@amazon.com}
\author{\fnm{Zhihua} \sur{Li}}\email{zhihuaa@amazon.com}
\author{\fnm{Jon} \sur{Wu}}\email{jonwu@amazon.com}
\author{\fnm{Alessandro} \sur{Achille}}\email{aachille@amazon.com}
\author{\fnm{Mani} \sur{Srivastava}}\email{manibsri@amazon.com}

\affil{\orgname{AWS AI Labs}, \orgaddress{\street{410 Terry Ave N}, \city{Seattle}, \postcode{98109}, \state{WA}, \country{USA}}}

% \affil[2]{\orgdiv{Department}, \orgname{Organization}, \orgaddress{\street{Street}, \city{City}, \postcode{10587}, \state{State}, \country{Country}}}

%\affil[3]{\orgdiv{Department}, \orgname{Organization}, \orgaddress{\street{Street}, \city{City}, \postcode{610101}, \state{State}, \country{Country}}}


\abstract{

Protecting digital identities of human face from various attack vectors is paramount, and face anti-spoofing plays a crucial role in this endeavor. 
Current approaches primarily focus on detecting spoofing attempts within individual frames to detect presentation attacks. 
However, the emergence of hyper-realistic generative models capable of real-time operation has heightened the risk of digitally generated attacks.
In light of these evolving threats, this paper aims to address two key aspects. 
First, it sheds light on the vulnerabilities of state-of-the-art face anti-spoofing methods against digital attacks. 
Second, it presents a comprehensive taxonomy of common threats encountered in face anti-spoofing systems. 
Through a series of experiments, we demonstrate the limitations of current face anti-spoofing detection techniques and their failure to generalize to novel digital attack scenarios. 
Notably, the existing models struggle with digital injection attacks including adversarial noise, realistic deepfake attacks, and digital replay attacks.
To aid in the design and implementation of robust face anti-spoofing systems resilient to these emerging vulnerabilities, the paper proposes key design principles from model accuracy and robustness to pipeline robustness and even platform robustness. 
Especially, we suggest to implement the proactive face anti-spoofing system using active sensors to significant reduce the risks for unseen attack vectors and improve the user experience.

}




\maketitle

\section{Introduction}
\label{sec:intro}
\begin{figure}[t]
    \centering
    \includegraphics[width=1.0\linewidth]{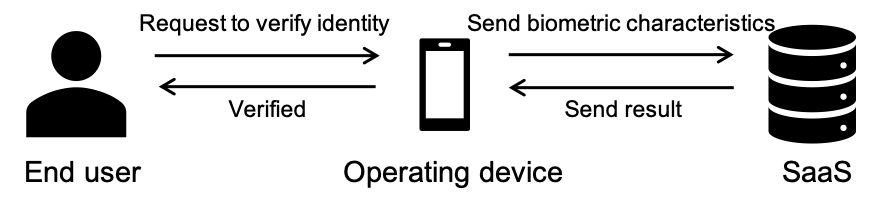}
    \caption{
    A common work flow of remote face verification system.
    }
    \label{fig:face_verification_saas}
\end{figure}

This paper primarily concentrates on exploring the security landscape and principles behind designing robust \textbf{remote face anti-spoofing techniques}.
As remote face verification systems gain popularity, they also become targets for exploitation by malicious actors to fool the system by spoofing the identity of the victim. 
By integrating face anti-spoofing techniques, these systems gain the ability to verify that the provided facial characteristics originate from a living individual who is physically and actively present in front of a camera-equipped client device. 
This significantly enhances the security measures, as illustrated in Figure~\ref{fig:face_verification_saas}. 
%but we believe that our insights and conclusions can be adapted to systems using other forms of biometrics.
% In this paper, we primarily focus on the threats in face face anti-spoofing in the realm of SaaS, as depicted in Figure~\ref{fig:face_verification_saas}. 
% Nonetheless, as the utilization of online face verification systems experiences a meteoric rise in popularity, malicious actors are presented with opportunities to exploit vulnerabilities within these verification systems, either for personal gain or to inflict harm. 
%\footnote{\url{https://www.scmp.com/tech/tech-trends/article/3127645/chinese-government-run-facial-recognition-system-hacked-tax}}\footnote{\url{https://www.nbcnews.com/tech/security/fbi-warns-deepfake-porn-scams-rcna88190}}.
% % ERIC: very hard to parse the following sentences
% This escalating demand for robust authentication solutions has paralleled a surge in security threats, underscoring the imperative need for a comprehensive comprehension of these risks and a state of readiness to effectively mitigate them.
% The potential repercussions of successful breaches encompass an array of outcomes, spanning from unauthorized access to financial fraud, thereby underscoring the pivotal significance of face anti-spoofing as an impregnable bastion safeguarding the integrity of the identity verification module and preempting fraudulent activities. 
% In this context, we envision a scenario where malicious entities have procured a victim's facial attributes and are determined to circumvent the facial face anti-spoofing mechanism.
Building a face anti-spoofing system that can be trusted in the face of diverse potential attackers is important, and requires designing the system to be robust to attacks and equipping them with defense mechanisms.

Developing a face anti-spoofing system robust to a range of potential attackers is crucial, necessitating robust design and fortified defense mechanisms. 
As discussed by \citet{zhang2023understanding}, vulnerabilities at the system level, inadvertently introduced during design, can be exploited by malicious actors subsequently. 
Consequently, the face anti-spoofing models must remain alert and adaptable to counter both existing and emerging threats, engaging in an ongoing battle against adversaries.
%While building a face face anti-spoofing system is essential, potential attackers can still breach its defense measures. 
%One potential threat lies at the system level, as discussed by \citet{zhang2023understanding}, where any imperfect system implementation could inadvertently create vulnerabilities that malicious actors could exploit.
%One vector of potential threats, as discussed by \citet{zhang2023understanding}, arise from system level vulnerabilities that are inadvertently created during system design and subsequently exploited by malicious actors.
%Going beyond pre-existing system level vulnerabilities, the face anti-spoofing models must also remain highly vigilant and quickly adaptive to new threats, thus resulting in a continual battle with adversaries.
%On the other hand, a robust face liveness system must remain highly vigilant and adaptable to confront both known and unknown threats, resulting in an ongoing and ever-evolving adversarial battle. 

%Currently, physical domain presentation attacks, such as the use of masks or displaying images/videos, are the most prevalent threats \citep{hernandez2019introduction}. These attacks involve mimicking the victim's facial features to the sensor. 
%Additionally, the emergence of Artificial Intelligence Generated Content (AIGC) introduces digital injection attacks, where malicious actors inject harmful digital content directly into the data stream \citep{carta2022video}.

At present, most prevalent attacks are physical domain presentation attacks~\citep{hernandez2019introduction} where the adversary impersonate the victim's identity by presenting physical objects (\eg, masks, photographs, images/videos on a display) to the sensor that mimic the victim's facial geometry and texture.
%involving the submission of mimicking the victim's facial geometry and texture using physical objects/presentation medium to impersonate the victim's identity and physical presentation in front of the sensor. 
Additionally, with the rise of Artificial Intelligence Generated Content (AIGC), digital injection attacks represent another substantial and growing threat to face anti-spoofing systems.
In digital injection attacks, the malicious actors must bypass the sensor and directly inject malicious digital content into the data stream~\citep{carta2022video}. 
In scenarios involving a remote face anti-spoofing system, the operating systems (OS) and software are incapable of authenticating the genuineness of the camera source, and device registration can be easily manipulated to deceive the OS if the attacker has the control to the device. An attacker can effortlessly establish a virtual camera and inject digital content into a secured network, making the attack feasible at the capture level and difficult to counteract with network security alone.

Among digital injection attacks, Deepfake attacks pose a significant threat. 
With increasing access to many high-quality public generative models, facilitated by platforms such as mobile applications and open-source code repositories,
%(\textit{e.g.} \texttt{ZAO} and \texttt{FaceAPP})
%(\textit{e.g.}, \texttt{GitHub} and \texttt{Hugging Face})
%, and interactive deep-learning model interfaces (\textit{e.g.}, \texttt{Gradio}), 
even individuals without extensive computer vision expertise can create convincing Deepfakes. 
Specifically, an attacker can either replicate the identity of the targeted victim by employing a calibrated personalized generative model to synthesize image or video that reproduce biometric characteristics with sufficient fidelity to allow authorization, or generate an entirely new identity utilizing cutting-edge techniques like Stable Diffusion~\citep{rombach2022high} in order to compromise the authentication process. These two types of attack are known as \emph{face-swap} and \emph{entire face synthesis} respectively in the deepfake detection literature.
% More specifically, the synthesized content can contain an unknown identity or mimic the specific identity of a targeted victim, a category of threat commonly referred as deepfakes attack.
% % For instance, malicious actors could use Stable Diffusion \citep{rombach2022high} to generate unknown identity impacting the authentication use case. 
% Moreover, attacker could further fine-tune for victim-targeted personalized model can further improve the quality of the model's output and generate realistic synthetic biometric characteristics to pass authorization use case.

Beside deepfakes, there are two additional digital attack categories which are relevant but have been less well studied in the context of face anti-spoofing.
One attack is adversarial noise and patches~\citep{goodfellow2014explaining} which pose a significant threat to modern machine learning models processing by exploiting the vulnerability of deep neural networks to minor input perturbations that can flip model predictions without raising any suspicion among human observers and auditors. Another attack is digital replay attack which refers to the injection of previously captured data replicating biometric characteristics of the victim, such as photos sourced online or captured physically, directly into the verification system.
While \citet{li2022seeing} delved into the risks associated with deepfake attack on remote face anti-spoofing production solutions, there remains a significant gap in comprehensively understanding the impact of the other two threat categories from a machine learning perspective.
This gap extends to understanding the conditions under which these threats could succeed and their implications for the underlying face anti-spoofing models, highlighting the need for further research.

\begin{table*}[htbp]
\caption{Glossary in the face anti-spoofing literature.}
\label{tab:glossary}
\begin{center}
\begin{tabularx}{\textwidth}{p{15em}p{5em}X}
 \hline
 Term & Abbreviation & Meaning \\
 \hline\hline
 %Liveness & - & the state of the signals collected from an subject who is physically, actively, and closely operating the system. \\ \hline
 Anti-spoofing & - & The task to detect the presence of spoofing attempt; computer's ability of an algorithm to verify that it is interacting with a non-living or artificially created representation \\  \hline
 %Passive Anti-spoofing & - & \\ \hline
 %Active Anti-spoofing & - & \\ \hline
 Presentation Attack & PA & The attack attempts using simulated presentation media such as prints, digital devices, and even 3D mask;
 % heart rate, breathing, movement
 % Authentification: liveness, identity (face match, only face not DNA)
 \\  \hline
 Presentation Attack Detection & PAD & The task to detect the presentation attack;\\  \hline
 Digital Injection Attack & DIA & The attack attempts by injecting the digital content, which might be the copy/forgery to a genuine biometric characteristics or fully synthesis, to a face anti-spoofing system;\\  \hline
 %Digital Injection Attack Detection & DIAD & The task to detect the digital attack;\\  \hline
 Attack Presentation Classification Error Rate & APCER & The proportion of attack presentations attack incorrectly classified as bona fide presentations;\\  \hline
 Bina-fide Presentation Classification Error Rate & BPCER & The proportion of bona-fide presentations incorrrectly classified as presentation attack;\\  \hline
 Average Presentation Classification Error Rate & ACER & The average of APCER and BPCER;\\  \hline
 False Acceptance Rate & FAR & In biometric systems, a false match rate in verification systems that factors the occurrence of multiple attempts and of failures to acquire. A higher value indicates higher vulnerability.\\  \hline
 False Rejection Rate & FRR & In biometric systems, a false non-match rate in verification systems that factors the occurrence of multiple attempts and of failures to acquire. A higher value indicates higher user friction.\\  \hline
\end{tabularx}
\end{center}
\end{table*}

By Liebig's law of the minimum, the overall security level of a system is dictated by the most vulnerable module within it.
In this work, we investigate the security level of a system from various angles and propose the best practices to mitigate the risks.
%We start by offering precise definitions for the terminology employed in this paper, particularly clarifying concepts such as liveness and anti-spoofing, which are often misused in the literature.
We then conduct a systematic analysis of threat models, with a specific focus on the security aspects of remote face verification systems.
% We consider risks ranging from physical-world mimicry attack to digital-world content manipulations, with a particular emphasis on an intricate examination of threats impacting face face anti-spoofing.
By this analysis, physical presentation attacks and digital injection attacks are the two main categories of threats to the face anti-spoofing system, especially to the underlying models that process the data captures by the sensor.
The other potential vulnerabilities can often be addressed through careful system design choices and the implementation of best practices in both machine model accuracy and robustness, and system design, and engineering for security. 
However, as noted earlier, currently the attention of face anti-spoofing research community is for the most part still on methods that passively classify whether the face of a person in a single image (a frame a of a video) corresponds to that of a live person or not. The reasons for this state of affairs is largely availability of training data and the need to keep the models low complexity.
%However, current research community's attention is still on single-image passive face livenes detection due to the data availability and model simplicity.

To call out the security risks present in these single-image passive models, we use publicly available datasets and recent proposed models to evaluate them under various settings based on our threat model.
Based on experimental results, we point out that a fundamental issue with the single-image passive models is that these models only focus on artifacts or anomalies in the input feature space, and do not verify the presentation of the subject.
%Through a series of simulated attack scenarios, we identify vulnerabilities within the passive face anti-spoofing system.
Such approaches are therefore suitable for deployment only use cases where the transactions are low valued and the sensor platform is trusted and secured.
%can not be deployed in the use case with low value transaction and secure hardware. 

To achieve high security for medium and high value transactions, we propose a list of fundamental principles crucial for the design of robust face anti-spoofing systems, spanning from ML model robustness to ML system and platform robustness. 
Notably, instead of solely developing a passive model or designing a new active challenge, we advocate a proactive approach that harnesses the capabilities of sensors for active sensing and imaging, thereby enhancing prevention against novel, previously unseen threats.

\begin{figure*}[htbp]
    \centering
    \includegraphics[width=1.0\linewidth]{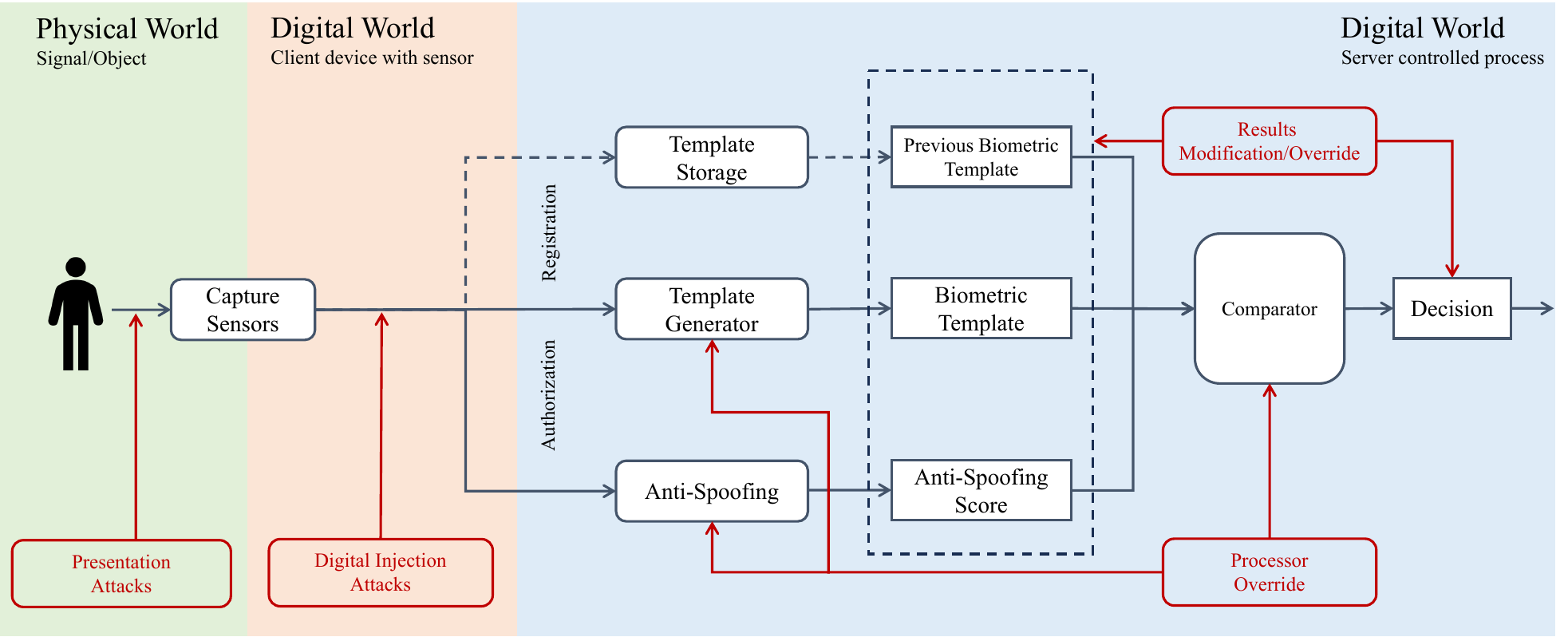}
    \caption{
    The points of threat attack in a biometric system (best viewed in colored). The attack can happen anywhere in the biometric system including physical and digital worlds: 
    In physical world (marked in green), the attackers can present the victim's photo on a presentation media or produce an adversarial environment to a biometric system.
    In transformed digital world (marked in red) on devices, the attackers can hi-jack the digital content to the data stream. 
    In server-controlled processing device, the attackers can modify/override any part of the processing system that can flip the final decision to bypass the biometric check.
    In the paper, we trust the protection on the server side and only focus on the discussion on the input from the physical and digital worlds.
    }
    \label{fig:threat_in_biometric_system}
\end{figure*}

The contribution of this paper can be summarized in four folds:
\begin{enumerate}[leftmargin=*]
    \item It provides a thorough overview of the security landscape for remote face anti-spoofing systems, highlighting the primary threats of presentation attacks (such as prints, physical replays, and 3D masks) and digital injection attacks (including Deepfakes, adversarial, and digital replay), along with their respective attack scenarios.
    \item It reviews the latest machine learning models for countering these threats in face anti-spoofing academic research community and summarizes their performance on the public available benchmarks as well as the general research trends.
    \item It presents evidence through extensive experiments that current state-of-the-art models are vulnerable to both unseen domains and novel attack types including the unseen attack within the same category or unseen more advanced attack vectors.
    \item It recommends that to enhance security and user experience, face face anti-spoofing systems need to be carefully systematic designed, considering the performance of the machine learning models, robustness of the machine learning models, inference pipeline, and service platform.
\end{enumerate}

\section{Glossary}

Recent works in face anti-spoofing have leveraged advancements from a multitude of research domains: face presentation attack detection ~\citep{wang2023wild}, image splicing ~\citep{huh2018fighting}, adversarial robustness~\citep{papernot2016limitations}, deepfake detection~\citep{tolosana2020deepfakes}, analog sensor robustness~\citep{fu2018risks}, and many more.
%
%Face face anti-spoofing leverages advancements from various dynamic research domains, including face anti-spoofing detection (FAS)~\citep{wang2023wild}, image splicing detection~\citep{huh2018fighting}, adversarial robustness~\citep{papernot2016limitations}, deepfakes detection~\citep{tolosana2020deepfakes}, analog sensor robustness~\citep{fu2018risks}, and more.
%
%To mitigate the confusion arising from the disparate terminologies found in recent literature, we initiate a process of establishing a more coherent glossary in Table~\ref{tab:glossary}.
% In the pursuit of enhancing community services, and recognizing the prevalent ambiguity surrounding certain terminologies in recent literature, our endeavor is to furnish a more coherent reference point.
In an effort to clarify disparate and overloaded terminology across these cross-functional domains, we provide a glossary of terms that we use in this paper in Table~\ref{tab:glossary}.
%Firstly, similar to the definition in ISO/IEC 30107-1:2016 \citep{iso30107}, we define \textit{spoofing} as:
%The most important terminology definitions are the foundational concepts of \textit{liveness} and \textit{face anti-spoofing}. Similar to the definition in ISO/IEC 30107-1:2016 \citep{iso30107}, we define \textit{liveness} in the field of face anti-spoofing as:
%\begin{quote}
%a measurable state of a human subject who is alive, and physically and actively engaged with a system
%the state of a human subject being alive can be measured who is physically and actively engaged with the system's operation.
%\end{quote}
%We define anti-spoofing as: 
%\begin{quote}
%the process to measure whether a biometric sample is captured from a live subject, who is physically present at the point of capture
%the measurement process to determine the biometric sample is being captured from a living subject present physically at the point of capture.
%\end{quote}
%From the preceding context, the attributes denoting the state of being alive that can be measured by face anti-spoofing or anti-spoofing encompass the following characteristics:
The face anti-spoofing score is determined by evaluating two key attributes: 
\begin{itemize}[leftmargin=*]
\item Physical Engagement: the subject is actively engaged in a physical interaction or operation within the system, which entails the occurrence of physical actions or movements that signify a living presence. This also indicates their proximity to the operational device.
\item Active Participation: the subject's engagement when operating the system is marked by intentional and purposeful actions within the certain time period, indicating a conscious and responsive involvement with the system's functionalities.
\end{itemize}
These attributes collectively define the concept of ``anti-spoofing" emphasizing the dynamic, intentional, and immediate nature of the subject's interaction with the system, distinct from the conventional medical interpretation of the term.

It is imperative to address a persistent misconception prevalent within the community surrounding the term ``anti-spoofing".
The concept of ``spoofing" involves a purposeful and methodical endeavor wherein the attacker strives to reproduce or imitate the distinctive characteristics inherent to a victim's identity. This emulation can manifest through a variety of spoofing methodologies, which include but not limited to physical presentation attack.
In direct contrast, the term ``anti-spoofing" encompasses a multifaceted spectrum of methodologies and strategies meticulously devised to counteract and nullify the efficacy of these aforementioned spoofing endeavors. 
These countermeasures are intricately designed to recognize the subtle yet crucial disparities between genuine interactions and fabricated simulations, thereby ensuring the veracity and authenticity of the face verification process.

\section{Threat Model}
\label{sec:threat_model}
In this section, we elucidate the threat model for a biometric system integrated with cloud computing, as depicted in Figure~\ref{fig:threat_in_biometric_system}. This model encompasses a spectrum of potential incursions spanning all stages of the system's operation, including biometric registration and authorization processes. In the event that an adversary successfully circumvents the biometric enrollment phase fortified by anti-spoofing models, the biometric characteristics enrolled therein become susceptible during the sequential identity verification phase. As both phases fundamentally contain analogous procedures, characterized by anti-spoofing as a common module, our analytical concentration naturally gravitates towards the authorization stage due to additional template comparison. Within this phase, signals from a subject in the physical environment are transformed into digital representations, facilitated by hardware sensors. Such data can be further processed by data processors running in the application and the server. Malicious actors possess the capability to orchestrate attacks at various junctures within this data propagation pipeline, spanning from the manipulation of physical objects (Presentation Attack) to tampering with the physical environment (Physical Adversarial Attack), digital content transmission (Digital Injection Attack), processor compromise (Processor Override), and the distortion or usurpation of result outcomes (Results Modification/Override).

Considering the intricacies of real-world applications and the typically heightened security measures surrounding models, applications, and servers, we choose to focus on \textbf{presentation attack} and \textbf{digital injection attack} as common threats to face anti-spoofing systems. This focus assumes that the application running on the device and the server handling the data are securely protected through various security engineering approaches.

\subsection{Presentation Attack}

The attacker can produce the presentation medium to mimic the presentation of the victim in front of the face anti-spoofing system.
In a presentation attack, the attacker can bypass the biometric verification system by presenting a biometric trait, which is referred to as a presentation attack instrument. 
The common presentation attack instrument include but are not limited to:
\begin{itemize}[leftmargin=*]
    \item Print: In this attack, the targeted individual's biometric features are printed on a tangible medium, such as paper, and presented to the biometric system as a means of impersonation.
    \item Physical Replay: This attack involves displaying the victim's biometric characteristics on a physical device, such as a tablet or screen, with the intention of deceiving the biometric system into recognizing the imitated traits as authentic.
    \item 3D Mask: In this type of attack, the perpetrator employs the victim's biometric attributes to create a three-dimensional mask. The production of such masks can range from low-cost methods to more precise and expensive approaches involving sophisticated 3D modeling techniques.
\end{itemize}

It is important to note that the aforementioned attack serve as representative examples and are not an exhaustive list of all possible presentation attack methodologies. For example, one possible attack is using Wax: This attack involves the fabrication of the victim's biometric characteristics using wax or similar materials, attempting to deceive the biometric system through the presentation of these replicated features.

Another threat that could happen on the data capture side but less been studied is physical adversarial patches or patterns. The attacker can generate such adversarial patches or patterns by modifying the object \citep{brown2017adversarial, xiao2021improving} or the signals to the sensor \citep{han2023dont} to fool the processor in the physical world.
% the physical adversarial patch can be deployed to produce the adversarial samples in physical world. 
This attack usually requires domain expertise and the algorithm usually took multiple inferences to the victim model and efficient computation power to produce the adversarial patches or patterns.
The prerequisite is usually the adversarial computation from the digital world with the object to improve the transferability from digital to physical world.
  % modify objects in the scene, light shed, optics hack (VR), transparent mask with scar, modify light-rays (physical world), modify signal (digital)
  % why are we focus attack before the camera? Because the pipeline from video feed to the execution system can be made secure (in an ideal world!)
  
%\noindent \textbf{3. Digital Injection Attack}: 
\subsection{Digital Injection Attack}
In an injection attack, the attacker inserts pre-recorded media content (including image or videos) or employs deepfakes technology to create realistic images and videos. Injection attack have gained popularity due to the easiness of producing the content containing biometric characteristics with the rapid growth of the development on AIGC, allowing individuals with limited or no development knowledge to carry out such attack. 
%In fact, the CEO of iProov reported that injection attack occur six times more frequently than presentation attack\footnote{\url{https://www.biometricupdate.com/202210/digital-injection-6x-more-frequent-than-biometric-presentation-attack}}, highlighting their growing prevalence.
The digital injection attack can be divided into two steps: crafting attack contents and bypassing the sensor. 

The injected content can be a digital synthesis (deepfakes), digital adversarial samples, or even digital copy of the victim's biometric characteristics:
\begin{itemize}[leftmargin=*]
    \item Deepfakes attack: It could be generated using off-the-shelf or self-developed model or algorithm to produce the content containing information of the victim. The quality of this synthesis can be varies depending on the generation method and the information of the victim exposed to the attacker.
    \item Digital adversarial attack: If the attacker have the access (either unlimited or limited) to call the model, there is also a possibility that the attacker can craft the adversarial samples (including adversarial noise) to bypass the system by iterating with the target face anti-spoofing models (\eg, calling the API multiple times) or inferencing the surrogate model. It usually serves the one step before producing physical adversarial attack.
    \item Digital Replay attack: the attacker can source the digital copy of the victim's biometric characteristics and replay to the system by injection. It would not contain any artifacts and can be easily pass the system. It is also called honeypot risk or video injection attack in some context.
\end{itemize}

The method of bypassing the sensor depends on the target, and it can vary based on the following factors:
\begin{itemize}[leftmargin=*]
    \item Device Type: The type of device, such as mobile phones, desktop computers, tablets, smart TVs, etc., can influence how the sensor is bypassed.
    \item Device Model: Different models of the same device type may have variations in hardware, which can affect the approach used to bypass the sensor.
    \item Operating System: Different operating systems, like iOS or Android, have their own security strengths and weaknesses.
    \item Application Versions: different versions of the target application might have varying levels of security or vulnerabilities.
\end{itemize}
The complexity of the bypassing step can range from relatively simple methods like using a virtual camera to capture video feeds, to more advanced approaches such as application source code hacks.

\begin{table*}[htb]
\caption{Likelyhood of attack.}
\label{tab:attack_likelyhood}
\begin{adjustbox}{width=\linewidth,center}
\begin{tabular}{llc}
\toprule %
Attack & Condition & Likelihood \\
\midrule
Presentation Attack & Access to victim's identity & High\\
\midrule
Digital Adversarial Attack & Access to victim's identity; Access to (call) victim model; & Low\\
Deepfakes Attack & Access to the victim's identity; Access to the sensor or network; & High \\
Digital Replay Attack & Access to the victim's identity; Access to the sensor or network; & Medium \\
\midrule
Processor Override & Access to the processor of the application or server; & Low \\
\midrule
Results Override & Access to the the application or server; & Low\\
\botrule
\end{tabular}
\end{adjustbox}
\end{table*}

%\noindent \textbf{4. Processor Override}: 
\subsection{Processor Override}
The attackers can modify or override the processing module on the sensed data, which can happened on the edge devices.
For example, some data compression operations on untrusted device (\eg, mobile) to manipulate the data source.
Moreover, the attacker have the access to the processing container and the system is compromised by the attacker.
If the attacker has the control to the server, the attackers can modify or override any processing module including template generator, face anti-spoofing, and template comparator. 
Even more, there is a supply chain / backdoor attack to the trained model, which might exists when there is ungratefully data processing during the training process \citep{gao2020backdoor}. 

%\noindent \textbf{5. Results Override}:
\subsection{Results Modification/Override}
If attackers gain access to the application, server, or hijack the network communication channel, they can potentially modify part of or override complete results at a specific stage to bypass the anti-spoofing check and gain unauthorized access.

face anti-spoofing is a multifaceted process that encompasses physical elements within the scene, hardware sensing and imaging, data transmission, and the robust inference models hosted in a cloud service. Safeguarding against various attack necessitates a holistic approach throughout the entire development cycle of machine learning operations and the implementation of safety engineering practices, including tamper-resistant packaging and electromagnetic shielding.
Furthermore, the probability of encountering the aforementioned attack vectors is not uniform and depends on the system's configuration. For instance, presentation attack methods can be relatively easy to execute in physical environments, whereas digital injection attack may occur when end users have open access to capture devices. We have summarized the likelihood of specific attack and their conditions in Table~\ref{tab:attack_likelyhood}.

\begin{table*}[thb]
\caption{A summary of publicly available dataset for presentation attack detection.}
\label{tab:dataset_pad}
\begin{adjustbox}{width=\linewidth}
\begin{tabular}{lccccccp{10mm}p{10mm}}
\toprule %
Dataset & \#Live/Spoof & \#Sub. & Modality & Setup Variations & Attack Types \\
\midrule
\makecell[l]{CASIA-MFSD \\\citep{zhang2012face}} & 150/450 & 50 & RGB Video & 3 image quality & \makecell{Print (flat, wrapped, cut), \\Replay (tablet)} \\ \midrule
\makecell[l]{IDIAP-Replay \\\citep{chingovska2012effectiveness}} & 200/1000 & 50 & RGB Video & - & Print (flat), Replay (tablet, phone) \\ \midrule
\makecell[l]{MSU-MFSD \\\citep{wen2015face}} & 70/210 & 35 & RGB Video & \makecell{Indoor,\\ 2 Types of cameras} & Print (flat), Replay (tablet, phone) \\ \midrule
\makecell[l]{OULU-NPU \\\citep{OULU_NPU_2017}} & 990/3960 & 55 & RGB Video & \makecell{Indoor,\\ 6 mobile devices} & Print (flat), Replay (phone) \\ \midrule
\makecell[l]{SiW \\\citep{liu2018learning}} & 1320/3300 & 165 & RGB Video & \makecell{Indoor,\\ variations of distance, pose, \\ illumination, and expression} & \makecell{Print (flat, wrapped), \\Replay (phone, tablet, monitor)} \\ \midrule
\makecell[l]{SiW-M \\\citep{liu2019deep}} & 660/968 & 493 & RGB Video & \makecell{Indoor,\\ variations of distance, pose, \\ illumination, and expression} & \makecell{Print (flat), \\Replay (phone, tablet, monitor), \\Mask (hard resin, plastic, \\silicone, paper, Mannequin),\\ Makeup (cusmetics, \\impersonation, obfuscation),\\ Partial (glasses, cut paper)} \\ \midrule
\makecell[l]{CelebA-Spoof \\ \citep{zhang2020celeba}} & 156384/469153 & 10177 & RGB Image & \makecell{Indoor \& outdoor, \\4 Illumination conditions, \\rich annotations} & \makecell{Print (flat, wrapped), \\Replay (phone, tablet, monitor), Mask (Paper)}\\ \midrule
\makecell[l]{CASIA-SURF\\ 3DMask \citep{zhang2020celeba}} & 288/864 & 48 & RGB Images & 3 decorations, 6 environments & Mask (mannequin with 3D Print)\\ \midrule
\makecell[l]{CASIA-SURF\\ CeFA \citep{liu2019static}} & 2100/9300 & 1500 & RGB, IR, \& Depth & Indoor,3 ethnicties & Print, Replay, Mask (Print, Silicone)\\ \midrule
\makecell[l]{CASIA-SURF\\ HiFiMask \citep{liu2022contrastive}} & 13650/40950 & 75 & RGB Videos & \makecell{3 mask decorations, \\7 recording devices, \\6 lighting conditions, \\7 scenes} & Mask (transparent, plaster, resin)\\ \midrule
\makecell[l]{CASIA-SURF\\ SuHiFiMask \citep{fang2023surveillance}} & 10195/10195 & 101 & RGB Video & \makecell{Long distance, Outdoor, \\ Variations in lighting and whether}  & Print, replay, Mask \\
\botrule
\end{tabular}
\end{adjustbox}
\end{table*}

\section{Related Work}
\label{sec:related_work}
In this section, we highlight some recent work including literature reviews for presentation attack detection, digital injection attack detection and the other defense approaches.

\begin{table*}[thb]
\caption{Comparison of different presentation attack detection approaches with single domain evaluation.}
\label{tab:pad_survey}
\begin{adjustbox}{width=\linewidth}
\begin{tabular}{lccccccccc}
\toprule %
\multirow{2}{*}{Method} & \multicolumn{4}{c}{SiW} & \multicolumn{4}{c}{OULU}\\ \cmidrule(lr){2-5} \cmidrule(lr){6-9}
 &Protocol & APCER & BPCER & ACER & Protocol &APCER & BPCER & ACER \\ \midrule

\multirow{4}{*}{\makecell[l]{LMFD-PAD\\ \citep{fang2022learnable}}} & 1 & - & - & - & 1 & 1.4 & 1.6 & 1.5  \\
& 2 & - & - & - & 2 & 3.1 & 0.8 & 2.0  \\
& 3 & - & - & - & 3 & 3.5 $\pm$ 3.2 & 3.3 $\pm$ 3.2 & 3.4 $\pm$ 3.1  \\
& - & - & - & - & 4 & 4.5 $\pm$ 5.3 & 2.5$\pm$4.1 & 3.3$\pm$3.1  \\ \midrule

\multirow{4}{*}{\makecell[l]{PatchNet\\ \citep{wang2022patchnet}}} & 1 & 0.00 & 0.00 & 0.00 & 1 & 0.00 & 0.00 & 0.00  \\
 & 2 & 0.00 $\pm$ 0.00 & 0.00 $\pm$ 0.00 & 0.00 $\pm$ 0.00 & 2 & 1.1 & 1.2 & 1.2  \\
 & 3 & 3.06 $\pm$ 1.10 & 1.83 $\pm$ 0.83 & 2.45 $\pm$ 0.45 & 3 & 1.8 $\pm$ 1.47 & 0.56 $\pm$ 1.24 & 1.18 $\pm$ 1.26  \\
 & - & - & - & - & 4 & 2.5 $\pm$ 3.81 & 3.33$\pm$3.73 & 2.9$\pm$3.0  \\ \midrule

\multirow{4}{*}{\makecell[l]{TransFAS\\ \citep{wang2022face}}} & 1 & 0.00 & 0.00 & 0.00 & 1 & 0.8 & 0.0 & 0.4  \\
 & 2 & 0.00 $\pm$ 0.00 & 0.00 $\pm$ 0.00 & 0.00 $\pm$ 0.00 & 2 & 1.5 & 0.5 & 1.0  \\
 & 3 & 1.95 $\pm$ 0.40 & 1.92 $\pm$ 0.11 & 1.94 $\pm$ 0.26 & 3 & 0.6 $\pm$ 0.7 & 1.1 $\pm$ 2.5 & 0.9 $\pm$ 1.1  \\
 & - & - & - & - & 4 & 2.1 $\pm$ 2.2 & 3.8$\pm$3.5 & 2.9$\pm$2.4  \\ \midrule

\multirow{4}{*}{\makecell[l]{SRENet\\ \citep{guo2022multi}}} & 1 & 0.00 & 0.00 & 0.00 & 1 & 0.2 & 0.6 & 0.4  \\
 & 2 & 0.00 $\pm$ 0.00 & 0.00 $\pm$ 0.00 & 0.00 $\pm$ 0.00 & 2 & 1.4 & 0.8 & 1.1  \\
 & 3 & 6.3 $\pm$ 1.3 & 2.9 $\pm$ 0.4 & 4.6 $\pm$ 0.9 & 3 & 1.6 $\pm$ 1.6 & 1.2 $\pm$ 1.4 & 1.4 $\pm$ 1.5  \\
 & - & - & - & - & 4 & 2.2 $\pm$ 1.9 & 3.8$\pm$4.1 & 3.0$\pm$3.0  \\ \midrule

\multirow{4}{*}{\makecell[l]{LDCNet\\ \citep{huang2022learnable}}} & 1 & - & - & - & 1 & 0.0 & 0.0 & 0.0  \\
& 2 & - & - & - & 2 & 0.8 & 1.0 & 0.9  \\
& 3 & - & - & - & 3 & 4.55 $\pm$ 4.55 & 0.58 $\pm$ 0.91 & 2.57 $\pm$ 2.67  \\
& - & - & - & - & 4 & 4.50 $\pm$ 1.48 & 3.17 $\pm$ 3.49 & 3.83 $\pm$ 2.12  \\ \midrule

\multirow{4}{*}{\makecell[l]{PDLE\\ \citep{kwak2023liveness}}} & 1 & - & - & - & 1 & - & - & 0  \\
& 2 & - & - & - & 2 & - & - & 1.2 \\
& 3 & - & - & - & 3 & - & - & 0.96 $\pm$ 1.03  \\
& - & - & - & - & 4 & - & - & 0.63 $\pm$1.04  \\ \midrule

\botrule
\end{tabular}
\end{adjustbox}
\end{table*}

\subsection{Presentation Attack Detection}
Presentation attack \citep{ramachandra2017presentation, walia2021contemporary,yu2022deep, wan2023advances} refers to an individual presents a physical artifact, such as a photograph or replay~\citep{zhang2012face}, a mask~\citep{erdogmus2014spoofing}, or a 3D model of a face~\citep{zhang2020casia,wang2023wild}, to a facial recognition system with the intent of deceiving it into recognizing the artifact as a genuine live face.
In the compilation of recent literature, several datasets have been summarized in Table~\ref{tab:dataset_pad}, from which a few distinct trends emerge:
\begin{itemize}[leftmargin=*]
    \item Increased Data Volume: The size of datasets has grown significantly over time, evolving from hundreds to tens of thousands of samples. This expansion indicates a trend towards more substantial, richer datasets to improve the training and evaluation of models.
    \item Diversity in Attack Types: Initially, datasets predominantly included basic presentation attacks, such as printed paper or physical replay attacks. However, more recent compilations have begun to encompass a broader spectrum of sophisticated attack methods. For instance, the CASIA-SURF-series datasets introduce 3D masks constructed from various materials, and the SiW-M dataset provides a finer classification of attack types, demonstrating an increased focus on capturing a wide range of fraudulent activities.
    \item Variability in Data Capture Conditions: There is a clear effort in recent datasets to include a greater diversity of capture conditions, including variations in lighting, environmental settings, devices used for capturing data, and distance from the subject. Additionally, these datasets aim to represent a broader range of subject conditions, such as different poses, expressions, ethnic, and skin-tones, to ensure the models developed can operate effectively across varied real-world scenarios.
    \item Incorporation of Multiple Modalities: The latest datasets, such as the CASIA-SURF CeFA, have started to integrate data from additional sensors, including infrared (IR) and depth sensors, alongside traditional RGB camera information. This multi-modal approach is designed to enrich the dataset with more comprehensive information about each subject, aiding in the development of more robust and accurate detection systems.
\end{itemize}

These trends reflect the evolving landscape of research in presentation attack detection, highlighting the community's efforts to address the increasing complexity and sophistication of attacks through the development of more diverse and comprehensive datasets. 

Limited by the dataset and main use cases, in the domain of defense mechanisms against presentation attacks, a significant portion of research has traditionally focused on utilizing RGB-based inputs. 
This approach, while foundational, has seen a variety of innovations in terms of supervision techniques and loss function designs, as evidenced by notable contributions in the field \citep{wen2015face,li2016generalized,liu2018learning,liu2019deep,wang2020cross,yu2020searching,liu2021face,xu2021improving,liu2022contrastive,guo2022multi}. These studies have laid the groundwork for subsequent advancements in detecting and countering presentation attacks. For readers interested in a comprehensive overview of the field's evolution up to 2022, we refer them to thorough literature surveys available in \citep{ramachandra2017presentation, walia2021contemporary,yu2022deep}. These reviews provide valuable insights into the methodologies, challenges, and progress made in presentation attack detection over the years.
In this section, however, our focus shifts to encapsulating the more recent endeavors within this research area. 
The intention is to highlight advancements and innovations post-2022, reflecting the ongoing efforts to refine detection capabilities and address the ever-evolving landscape of presentation attacks. 

%This recent work represents the cutting-edge in devising strategies to counteract sophisticated fraud attempts, particularly those exploiting weaknesses in systems reliant on RGB data. Through this summary, we aim to offer a window into the latest methodologies and their potential to fortify defenses against an array of presentation attacks, underscoring the dynamic and forward-moving trajectory of research in this vital area of cybersecurity.

\begin{table*}[thb]
\caption{Comparison of different presentation attack detection approaches in recent years with four popular benchmark datasets: CASIA (C), IDIAP-Replay (I), MSU-MFSD (M), and OULU-NPU (O). All metric are presented in percentage.}
\label{tab:pad_lr_cross_domain}
\begin{adjustbox}{width=\linewidth}
\begin{tabular}{lccccccccccc}
\toprule %
\multirow{2}{*}{Method} & \multirow{2}{*}{Key idea} & \multicolumn{2}{c}{OCI$\rightarrow$M} & \multicolumn{2}{c}{OMI$\rightarrow$C} & \multicolumn{2}{c}{OCM$\rightarrow$I} & \multicolumn{2}{c}{ICM$\rightarrow$O} & \multicolumn{2}{c}{Average} \\ \cmidrule(lr){3-4} \cmidrule(lr){5-6} \cmidrule(lr){7-8} \cmidrule(lr){9-10} \cmidrule(lr){11-12}
& & HTER & AUROC & HTER & AUROC & HTER & AUROC & HTER & AUROC & HTER & AUROC \\
\midrule

\makecell[l]{LMFD-PAD \\ \citep{fang2022learnable}} & \makecell{Dual stream network for both RGB \\and multi-level  frequency decompositions;\\ Hierachical spatial attentions\\ and one channel attention} & 10.48 & 94.55 & 12.50 & 94.17 & 18.49 & 84.72 & 12.41 & 94.95 & 13.47 & 92.10 \\ \midrule
\makecell[l]{FGHV\\ \citep{liu2022feature}} & \makecell{Separate feature generation process for\\ real and spoof; Enforce different similarities \\ among raw and two generated features} & 9.17 & 96.92 & 12.47 & 93.47 & 16.29 & 90.11 & 13.58 & 93.55 & 12.88 & 93.51\\ \midrule
\makecell[l]{SSAN-R\\ \citep{wang2022domain}} & \makecell{Disentangle features into content and \\ multiscale styles; Constrastive learning \\ between raw features and reassembled ones} & 6.67 & 98.75 & 10.00 & 96.67 & 8.88 & 96.79 & 13.72 & 93.63 & 9.82 & 96.46\\ \midrule
\makecell[l]{PatchNet\\ \citep{wang2022patchnet}} & \makecell{Formulate as patch type recognition \\problem; Asymmetric (real/spoof) \\angular margin softmax} & 7.10 & 98.46 & 11.33 & 94.58 & 13.4 & 95.67 & 11.82 & 95.07 & 10.91 & 95.95 \\ \midrule
\makecell[l]{TransFAS\\ \citep{wang2022face}} & \makecell{Employ graph to model the relationship\\ between patches in adjacent layers \\and fuse hierarchically} & 7.08 & 96.69 & 9.81 & 96.13 & 10.12 & 95.53 & 15.52 & 91.10 & 10.63 & 94.86\\ \midrule
\makecell[l]{AMEL\\ \citep{zhou2022adaptive}} & \makecell{Extract and aggregate domain-specific\\ information as the complement to \\domain-invariant one} & 10.23 & 96.62 & 11.88 & 94.39 & 18.60 & 88.79 & 10.23 & 96.62 & 12.74 & 94.11 \\ \midrule
\makecell[l]{EBDG\\ \citep{du2022energy}} & \makecell{Smaller energy for real faces and\\ larger for spoof; Mimic domain shifts \\ with energy to augment both} & 9.56 & 97.17 & 18.34 & 90.01 & 18.69 & 92.28 & 15.66 & 92.02 & 15.56 & 92.87\\ \midrule
\makecell[l]{GDA\\ \citep{zhou2022generative}} & \makecell{UDA by generation work which translate \\ target domain data into source\\ domain styles with neural statistic\\ and semantic preserved} & 9.2 & 98.0 & 12.2 & 93.0 & 10.0 & 96.0 & 14.4 & 92.6 & 11.45 & 94.90 \\ \midrule
\makecell[l]{ViT\\ \citep{huang2022adaptive}} & \makecell{Vision transformer backbone; \\ Ensembled multiple feature adapters; \\ Feature-wise transformation layer} & 1.58 & 99.68 & 5.70 & 98.91 & 9.25 & 97.15 & 7.47 & 98.42 & 6.0 & 98.54 \\ \midrule
\makecell[l]{LDCNet\\ \citep{huang2022learnable}} & \makecell{Learnable descriptive convolution; \\ Contrastive learning and dedicated attention} & 9.29 & 96.86 & 12.00 & 95.67 & 9.43 & 95.02 & 13.51 & 93.68 & 11.06 & 95.31\\ \midrule
\makecell[l]{DiVT-M\\ \citep{liao2023domain}} & \makecell{Small feature norm for real face; Classification\\ of real and all different spoof types } & 2.86 & 99.14 & 8.67 & 96.92 & 3.71 & 99.29 & 13.06 & 94.04 & 7.07 & 97.35 \\ \midrule
\makecell[l]{OSMTL\\ \citep{chuang2023generalized}} & \makecell{Learning with depth estimation and\\ face parsing; Meta learning to \\simulate domain shift; Triplet \\loss with live as anchor only } & 7.38 & 96.66 & 13.2 & 94.27 & 8.07 & 96.85 & 8.75 & 95.95 & 9.35 & 95.93 \\ \midrule
\makecell[l]{NDA-FAS\\ \citep{wang2023domain}} & \makecell{No real spoof data needed;\\ Real-face samples to be closer\\ to both local and global centers;\\ Pushing away synthetic spoof samples} & 4.29 & 99.27 & 13.22 & 92.89 & 5.50 & 96.90 & 17.45 & 89.93 &10.12	& 94.75 \\ \midrule
\makecell[l]{CRFAS\\ \citep{zheng2023learning}} & \makecell{Disentangle into semantic and domain\\ features; Only features with the \\same domain and category are taken\\ as the positive samples} & 6.90 & 98.30 & 9.33 & 97.31 & 7.78 & 97.14 & 15.52 & 92.18 & 9.88 & 96.23 \\ \midrule
\makecell[l]{PDLE\\ \citep{kwak2023liveness}} & \makecell{Cutmix real and spoof images \\ and regress the discretized label} & 5.41 & 98.85 & 10.05 & 94.27 & 8.62 & 97.60 & 11.42 & 95.52 & 8.875 & 96.56\\ \midrule
\makecell[l]{IADG\\ \citep{zhou2023instance}} & \makecell{Style-diversified samples for each\\ instance; Eliminates the style-\\sensitive feature correlation} & 5.41 & 98.19 & 8.70 & 96.44 & 10.62 & 94.50 & 8.86 & 97.14 & 8.40 & 96.57 \\ \midrule
% \makecell[l]{\\ \citep{}} & \makecell{} &  &  &  &  &  &  &  &  &  &  \\ \midrule
\makecell[l]{SAFAS\\ \citep{sun2023rethinking}} & \makecell{Separate features from different \\domains and classes; Alignment on \\the live-to-spoof transitions} & 5.95 & 96.55 & 8.78 & 95.37 & 6.58 & 97.54 & 10.00 & 96.23 & 7.83 & 96.42 \\ \midrule
\makecell[l]{DFANet\\ \citep{huang2023towards}} & \makecell{Disentangle into real/spoof features\\ and domain features; Augment \\real/spoof features by affine transformations;\\ Augment domain features by adversarial learning} & 5.24 & 97.98 & 8.78 & 97.03 & 8.21 & 96.84 & 9.34 & 96.43 & 7.89 & 97.07 \\ \midrule
\makecell[l]{FLIP-MCL \\ \citep{srivatsan2023flip}} & \makecell{CLIP pretraining; Constrastive learning \\between two image-text views } & 4.95 & 98.11 & 0.54 & 99.98 & 4.25 & 99.07 & 2.31 & 99.63 & 3.01 & 99.20 \\ \midrule
\makecell[l]{CF-PAD\\ \citep{fang2024face}} & \makecell{Mixstyling within the same category; \\ Random feature augmentation} & 8.11 & 96.43 & 11.78 & 95.64 & 16.50 & 91.50 & 9.87 & 95.13 & 11.57 & 94.68 \\ 
%\midrule
\botrule
\end{tabular}
\end{adjustbox}
\end{table*}

% Aside from ResNet, vision transformer also exhibits promising performance~\citep{huang2022adaptive,liu2023fm}.
% However, adapting existing methods to real-world scenarios is still difficult due to the inadequate diversity and scarcity of publicly available datasets~\citep{kose2013vulnerability,erdogmus2014spoofing,bhattacharjee2018spoofing,george2019biometric}. 
% To reduce the unreliability from RGB camera only and improve the security, multi-modality sensors such as IR and Depth camera have been deployed \citep{liu2021face}.
% To benchmark and accelerate the defense model improvement, several anti-spoofing datasets and challenges have been released \citep{zhang2020celeba,purnapatra2021face,liu2021cross,wang2023wild}. 
% Recently, research trends have increasingly focused on improving the model generalization for unseen domain, which involves developing techniques such as domain adaptation~\citep{gu2021pit,zhou2023self} and domain generalization~\citep{guo2022multi,zhou2023instance,sun2023rethinking} for face anti-spoofing.
% In addition, there are other forms of attacks in the physical world, such as analog sensor attacks~\citep{kune2013ghost,yan2020sok,man2020ghostimage,kohler2022signal,han2023dont}, and physical adversarial perturbations~\citep{sharif2016accessorize,eykholt2018robust,komkov2021advhat}. 
% However, these attacks either require special instruments or sophisticated domain knowledge, and therefore have received comparatively less attention in the defense literature. 

Continuing the trend predating 2022, certain approaches such as~\cite{zhou2022generative, du2022energy, zhou2022adaptive, wang2022face,chuang2023generalized,zhou2023instance} still rely on additional supervision, such as depth or face parsing tasks, with the aim of enhancing the model's understanding of spoofing artifacts at a pixel level. However, this reliance on off-the-shelf models for generating supplementary supervision often proves less effective in achieving generalization~\cite{wang2023wild}. 
While intra-dataset evaluations consistently yield nearly perfect results, as demonstrated in Table~\ref{tab:pad_survey}, researchers have noted a decline in performance when models are tested across different datasets, indicating a crucial deficiency in generalization for practical applications. Many studies, listed in Table~\ref{tab:pad_lr_cross_domain}, have shifted their focus towards improving cross-domain generalization, drawing from techniques in general domain generalization research area. Strategies such as image~\cite{zhou2022generative,wang2023domain, kwak2023liveness} or feature-level augmentation~\cite{liu2022feature, zhou2023instance, huang2023towards, fang2024face}, and meta-learning~\cite{zhou2022adaptive, du2022energy, chuang2023generalized} have gained popularity. The former entails manual intervention in feature design, while the latter prompts networks to autonomously adapt from multiple datasets. 

Addressing the tendency of deep networks to exploit spurious correlations while overlooking generalizable artifacts when discerning real from spoof images, several works~\cite{wang2022domain, sun2023rethinking, huang2023towards, zheng2023learning} advocate for disentangling features into domain-invariant and domain-specific ones during the learning process. Domain-invariant features typically encapsulate generalizable artifacts useful for distinguishing live and spoof examples across domains, while domain-specific features capture domain-specific information such as sensor or environmental factors. 
Disentanglement serves as the initial step, followed by the reassembling of these features. Domain-invariant features are then combined with shuffled~\cite{wang2022domain, zheng2023learning} or diversified (augmented~\cite{zhou2023instance, huang2023towards, fang2024face} using techniques like adaptive instance normalization or network-generated~\cite{liu2022feature, huang2022adaptive}) domain-specific features. These diversified domain-specific features help simulate data from unseen domains, thereby improving generalization.
Subsequently, contrastive learning is employed~\cite{wang2022domain, huang2022learnable, wang2022patchnet, liu2022feature, chuang2023generalized, zheng2023learning, wang2023domain} between original and reassembled features to obtain domain-invariant features. Moreover, leveraging the hypothesis that real features should exhibit more compact representations than spoof ones can further enhance model performance. This asymmetric hypothesis is explicitly enforced by~\cite{wang2022patchnet, du2022energy, zhou2023instance, liao2023domain}. Additionally, \cite{zhou2022adaptive} proposes dynamically aggregating domain-specific information based on its relevance to the unseen domain.

Rather than traditional convolutional neural networks, recent studies~\cite{wang2022face, liao2023domain, srivatsan2023flip, huang2022adaptive} have also explored transformer-based backbones for face anti-spoofing tasks. These vision transformers partitions input images into patches and models their relationships globally through self-attention mechanisms, thereby demonstrating improved generalization. Moreover, \cite{srivatsan2023flip} demonstrates that contrastive pretraining between image-text pairs will further enhances the model's generalization capabilities.

\begin{table*}[thb]
\caption{A summary of publicly available dataset for deepfake attack detection.}
\label{tab:deepfake dataset}
\begin{adjustbox}{width=\linewidth}
\begin{tabular}{lcccccc}
\toprule %
Dataset & \#Live/Spoof & \#Sub. & Modality & Setup & Attack Types \\
\midrule
DeepfakeTIMIT \citep{korshunov2018deepfakes}  & 320/620 & 32 & Video & In-the-wild & FaceSwap, DeepFake \\ \midrule
FaceForensics++ \citep{rossler2019faceforensics++} & 1,000/2,000 & - & Video & In-the-wild & \makecell{Face2Face, FaceSwap, \\DeepFakes, and NeuralTextures} \\ \midrule
Celeb-DF \citep{li2020celeb} & 590/5,639 & 59 & Video  & Internet & Improved DeepFakes \\ \midrule
DFDC Preview \citep{dolhansky2019deepfake}  & 1,140/4,074 & 66 & Video and audio & Volunteer Actors & Unknown \\ \midrule
DFDC \citep{dolhansky2020deepfake} & 23,654/ 104,500 & 960 & Video and audio & Volunteer Actors & \makecell{DFAE, MM/NN face swap\\, NTH, and FSGAN} \\ \midrule
DeeperForensics \citep{jiang2020deeperforensics} &50,000/10,000  & 100 & Video & Paid actors & DeepFake Variational Auto-Encoder \\ \midrule
WildDeepfake \citep{zi2020wilddeepfake} & 3,805/3,509 & - & Video and audio & Internet & Unknown \\ \midrule
FakeAVCeleb \citep{khalid2021fakeavceleb} & 500/19,500 & 500 & Video and audio & Youtube celebrity & \makecell{FaceSwap, FSGAN, \\Wav2Lip, and SV2TTS} \\ \midrule
FFIW \citep{zhou2021face} & 10,000/10,000 & - & Video & In-the-wild & FaceSwap, DeepFaceLab, and FSGAN \\ \midrule
UniAttackData \citep{fang2024unified} & 1800/27906 & 1800 & RGB Video & Indoor lab & \makecell{Print, Replay, Mask, 6 adversarial \\attacks, and 6 deepfake attacks}\\
\botrule
\end{tabular}
\end{adjustbox}
\end{table*}

\subsection{Digital Injection Attack Detection}

Injection attack refers to an attacker inserts digitally content into the input stream of a facial recognition system. 
This type of attack is highly accessible on PC as it's supported by virtual camera in a straightforward fashion. 
There are also sophisticated injection attack against mobile devices through either hardware~\citep{chen2019biometric}, or software~\citep{carta2021pitfalls,carta2022video}, and even network \citep{francillon2011relay,jie2019tradeoff,liu2023time}.
The scope of this work is limited to accessible attack that are most likely to occur in practical scenarios, particularly targeting potential attackers who do not possess exceedingly advanced expertise in the specific domain. For example, executing an analog sensor attack against a face verification system requires the utilization of professional instruments~\citep{han2023dont}; well-established protocols to safeguard against man-in-the-middle attack have been developed over the past decades~\citep{rescorla2001ssl}. 

This type of attack commonly occurs in situations where cloud-based biometric verification systems are used on un-trusted devices \citep{li2022seeing, zhang2023understanding}. 
This is because the existing operating systems (OS) and software lack the ability to verify the authenticity of the camera source. 
Furthermore, device registration can be manipulated easily, allowing an attacker with device access to trick the OS. An attacker can simply create a virtual camera and insert digital content into a secure network (for example, through the HTTPS protocol). 
This makes the attack possible at the capture stage and challenging to thwart with just network security measures.
Given the inherent difficulty in completely mitigating security risks on client-side devices, this work operates under the assumption that all data streams sent to the cloud server are susceptible, and the focus lies in the examination of the data content.

In this category of digital injection attacks, we have identified three sub-categories: deepfake attacks, digital adversarial attacks, and digital replay attacks. In these scenarios, a malicious user is capable of creating and injecting different types of content into the system to manipulate the outcome of its predictions.

\subsubsection{Deepfakes Attack Detection}

\begin{table*}[thb]
\caption{A summary of recent approaches for deepfake attack detection.}
\label{tab:pad_oulu}
\begin{adjustbox}{width=\linewidth}
\begin{tabular}{lcccccccc}

\toprule %
\multirow{2}{*}{Method} & \multirow{2}{*}{Key idea} & \multirow{2}{*}{Input} & \multirow{2}{*}{Modalities} & \multirow{1}{*}{AUC (seen test set)} & \multicolumn{3}{c}{AUC (unseen test set)} \\ 
\cmidrule(lr){5-5} \cmidrule(lr){6-8} & & & & FF++ & CelebDF & DFD & DFDC \\

\midrule

DPNet \citep{trinh2021interpretable}  & Temporal inconsistencies & Multiple frames & Video & 99.20  & 68.2 & 92.44 & -- \\ \midrule

MADD \citep{zhao2021multi}  & Attention mechanism & Single frame & Video & 99.29 & 67.44 & --&-- \\ \midrule

I2G \citep{zhao2021learning}  & Pair-wise self-consistency & Single frame & Video & 99.79 & -- & 99.07 & 67.52 \\ \midrule

Joint \citep{zhou2021joint}  & \makecell{Video-audio multimodal \\ cross-attention} & Multiple frames & Video + audio & 99.65 & -- & --\\ \midrule

Self-supervised \citep{chen2022self}  & \makecell{ Synthesize the most \\ challenging forgeries for \\ self-supervised training} & Single frame & Video & 98.4 & 79.7 & --&-- \\ \midrule

Implicit-ID \citep{huang2023implicit}  & \makecell{Utilize difference between \\ explicit
and implicit identity} & Single frame & Video & 99.32 & 83.80 & 93.92 & 81.23 \\ \midrule

ID-unaware \citep{dong2023implicit}  & \makecell{Reduce the influence of the \\ ID representation} & Single frame & Video & 99.79 & --&--&73.74  \\ \midrule

UCF \citep{yan2023ucf}  & \makecell{Decomposes information into: \\ forgery irrelevant, method-specific, \\ and common forgery
features} & Single frame & Video & 99.6 & 82.4 & 94.5 & 80.5 \\ \midrule

Anomaly \citep{feng2023self} & \makecell{Capture the temporal synchronization \\ between video frames and sound} & Multiple frames & Video + audio &  -- & -- & -- & -- \\ \midrule

SFDG \citep{wang2023dynamic}  & \makecell{Exploit the relation-aware features \\ in spatial and frequency domains \\ via dynamic graph learning} & Single frame & Video & 99.53 & 75.83 & 88.00 & 73.64 \\ \midrule

AltFreezing \citep{wang2023altfreezing}  & \makecell{divide the weights of a spatio-temporal \\network into two groups: \\spatial-related and temporal-related} & Multiple frames & Video & 99.70 &  -- & 89.50 & 98.5 \\ %\midrule
\botrule
\end{tabular}
\end{adjustbox}
\end{table*}

\begin{figure*}[t]
  \centering
  \includegraphics[width=\linewidth]{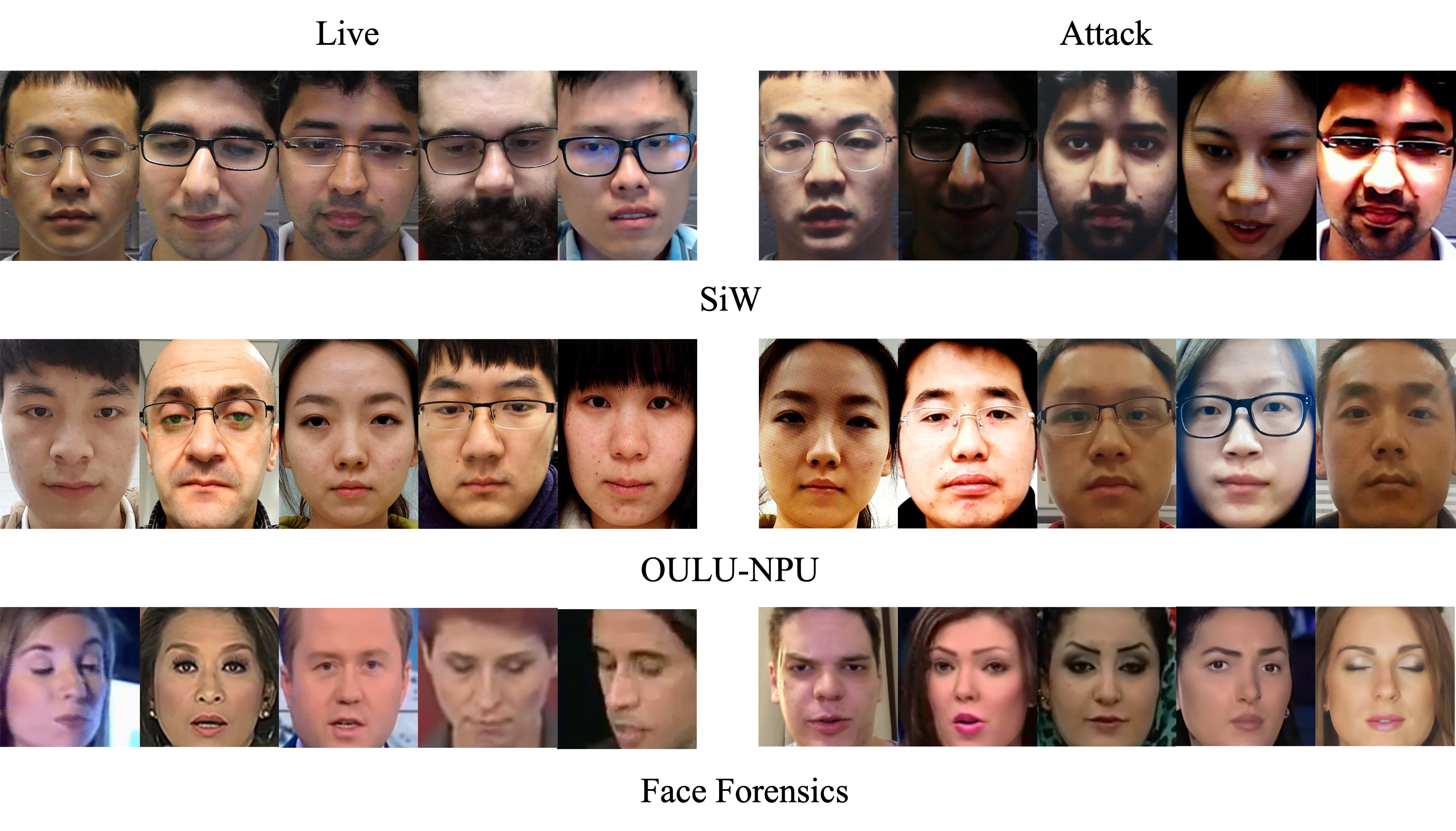}
  \caption{Example of images of live and attacks category in SiW, OUlU-NPU, and Face Forensics datasets.}
  \label{fig:example}
\end{figure*}

\noindent{\textbf{Deepfake Generation:}}
Face anti-spoofing is currently facing challenges due to the rapid advancements in realistic synthetic media techniques taking advantage of Generative Adversarial Networks (GANs) \citep{goodfellow2014generative} and recent prevalent diffusion models \citep{ho2020denoising}. Some representative face editing methods proposed in academia such as face-swap technologies~\citep{bitouk2008face,korshunova2017fast,bao2018towards,natsume2018rsgan,natsume2019fsnet,nirkin2019fsgan,li2019faceshifter,chen2020simswap,zhu2021one,li2021faceinpainter}, and reenactment technologies~\citep{wiles2018x2face,thies2019deferred,siarohin2019first,zakharov2019few,prajwal2020lip} are being maliciously employed within internet communities. 
These synthetic approaches are always well-known as Deepfakes \citep{mirsky2021creation, rana2022deepfake, nguyen2022deep}.
In particular, reenactment allows attackers to create videos with specific head and lip movements based on a single image of the victim synthesized by GANs~\citep{karras2019style,karras2020training} or diffusion models~\citep{ho2020denoising,ho2022classifier,rombach2022high}. 
Diffusion models also exhibit the exceptional capability to swap face through in-painting~\citep{nichol2021glide} on the image level, and there are also efforts on discovering text-to-video generation~\citep{zhang2023controlvideo}, posing additional potential threats to the face anti-spoofing system. In response, academia has introduced numerous deepfake detection datasets as summerized in Table \ref{tab:deepfake dataset}. As deepfake methods advance, so does the scale of accompanying datasets. Take the DFDC \citep{dolhansky2020deepfake}, for example. It's one of the largest and most challenging deepfake datasets, featuring over 100,000 clips from 3,426 paid actors and 8 manipulation methods (e.g., Deepfake Autoencoder (DFAE), MM/NN face swap, NTH, FSGAN, StyleGAN, Refinement, and audio swaps). These clips, generated using various techniques, closely mimic real-world scenarios, posing significant challenges to current deepfake detection methods.
Another noteworthy dataset is FakeAVCeleb \citep{khalid2021fakeavceleb}, pioneering in video-audio multimodal face synthesis. It uses AI algorithms to synthesize audio and adjust lip movements, effectively impersonating someone else's speech and gestures. These developments highlight the need for more robust detection and generalizable strategies against deepfakes in multiple modalities.

\noindent{\textbf{Deepfake Detection:}}
In recent years, many deepfakes detection methods have been proposed~\citep{li2018exposing,korshunov2018deepfakes,tripathy2020icface,ciftci2020fakecatcher,dong2020identity,li2020face,wang2020cnn,haliassos2021lips,li2021exploring,zhao2021learning,li2021frequency,liu2021spatial,luo2021generalizing,wang2022m2tr,huang2023implicit,dong2023implicit}. In Table 7, we present a collection of key ideas, modalities, and performance metrics from recent prominent works in deepfake detection. Our taxonomy of detection methods is based on three primary perspectives: (1) identification of spatial and frequency artifacts (e.g., I2G and SFDG), (2) assessment of temporal consistencies (e.g., DPNet), and (3) synchronization across modalities (e.g., Anomaly and Joint). Evaluation typically involves testing on both seen datasets, such as FF++, where manipulation methods are present in the training set, and unseen datasets including CelebDF, DFD, and DFDC.

Analysis of intra-dataset performance on FF++ reveals consistently high AUC scores exceeding 99\%, with marginal improvement over recent years. However, cross-dataset evaluation, where manipulation methods are unseen during training, results in a dramatic decrease in AUC scores to the 70s and 80s. This underscores the persistent challenge of achieving cross-dataset generalization in deepfake detection, posing significant hurdles for both academia and industry. Notably, this vulnerability exposes large-scale pre-trained deepfake models to new manipulation methods, potentially compromising their claimed generalizability. This situation also poses risks to current face anti-spoofing systems, necessitating vigilant monitoring of emerging manipulation methods and the adaptation of deepfake detection models to cover these unseen techniques.

Additionally, the research community has been focusing on training with relatively smaller datasets such as FaceForensics$++$~\citep{rossler2019faceforensics++}, and the cross-domain performance is intricate as the existing datasets are noisy~\citep{rossler2019faceforensics++,li2020celeb,dong2020identity,dolhansky2020deepfake,he2021forgerynet,le2021openforensics,narayan2023df}. The effectiveness of the proposed defense methods against deepfakes in real-world scenarios is left as an open question. Moreover, while diffusion models exhibit promising capabilities in realistic face in-painting and generation, limited research has focused on researching detection mechanisms for them.
Figure~\ref{fig:example} depicts the examples of live and attack category from presentation detection and deepfake attack detection datasets.

\subsubsection{Adversarial Attack Detection}

Modern deep networks are susceptible to small perturbations in their input, raising a serious security concern when applying to critical domains~\citep{szegedy2013intriguing}. \citet{papernot2016limitations} distinguished four types of adversaries depending on the information the attackers have access to; we focus on the black-box scenario where only input-output pairs are known to the attacker. The black-box attack is the major threat to practical applications~\citep{cherepanova2021lowkey} due to the the transferability of adversarial attack~\citep{liu2016delving}. Extensive efforts have been made to defend against black-box attack~\citep{dong2019efficient,yang2020patchattack,deng2022understanding,deb2023faceguard}, as well as their variants like physical perturbation attack~\citep{sharif2019general,hu2022protecting}, and back-door attack~\citep{chen2017targeted,wang2019neural}, with the latter poses additional security risks to federated learning~\citep{bagdasaryan2020backdoor} and foundation models~\citep{bommasani2021opportunities}. In general, improving adversarial robustness of a deep learning model usually comes at a cost of deteriorating its performance on clean data~\citep{weng2020trade,wang2020once}.

\subsubsection{Digital Replay Attack Detection}
The concept of a replay attack has its roots in network security \citep{singh2018revisiting} and control systems \citep{mo2009secure}, where legitimate data transmissions are illicitly repeated. 
In the context of biometric verification transactions, it's possible for biometric characteristics to be recorded and subsequently replayed. 
To distinguish this from physical replay attacks in presentation attack scenarios, we refer to it as digital replay attacks \citep{ratha2001enhancing}. 
In certain cases such as survellance system, it might also be known as a video injection attack \citep{carta2022video, nagothu2019detecting}.
Differently, video injection attack can be covered but can be addressed by network security since the hardware, network, and software are owned or maintained by the surveillance checker.
In this scenario of remote biometric verification using the end-user's devices, the integrity of hardware cannot be guaranteed. 
An attacker could potentially use the victim's facial images obtained from the internet or real-life encounters to spoof the system. 
The existing defenses against presentation attacks, deepfakes, and adversarial attacks are ineffective in this case. 
This is because the biometric data used in the spoofing are genuine, captured from the actual victims, and do not contain any artificial anomalies or artifacts.

This threat could be potentially harmful for remote face anti-spoofing system that the attacker do not need to have powerful machine to generate deepfakes or perform adversarial attacks. 
The optimal solution for this issue is to cryptographically sign the biometric data at the hardware level, with verification carried out by the hardware manufacturer to confirm the authenticity of the captured content. 
However, this approach necessitates collaboration among hardware manufacturers, operating system providers, software developers, and face anti-spoofing service providers, which has yet to occur.

However, the defense approach to this threat is less studied in the academic field, especially in biometric community, resulting in limited literature addressing this issue.
With that being said, there is no silver bullet to solve this attack completely.
Using machine learning, a more feasible method to ensure the authenticity of the information is to embed a digital watermark into the content needing verification. 
For instance, research \citep{smith2015face, farrukh2020facerevelio, zhang2021aurora} has introduced the use of random color flashing on the device's screen as a digital watermark for the video sequence. 
This approach is distinct from light-based face anti-spoofing methods \citep{tang2018face, ebihara2020specular, liu2020livescreen}, which rely on specific light patterns and possess limited randomness or entropy. 
The key innovation in thread of methods is the generation and verification of a random color pattern on the screen, which acts as a way to authenticate the content.
However, the integrity of these color signals are not been verified leading to potential manipulation as well.

\subsection{Other Defenses}

To enhance the security of face anti-spoofing systems, current practices involve integrating supplementary modalities that engage with users. These modalities involve voice recognition~\citep{thullier2017text}, iris detection~\citep{bowyer2016handbook}, active gestures or eye blinking detection~\citep{li2022seeing}, blood flow detection~\citep{guzman2012thermal}, flashing lights reflection detection~\citep{chan2017face,tang2018face,farrukh2020facerevelio,liu2020livescreen,zhang2021aurora}. However, these challenges can still be potentially simulated using contemporary technologies, and thereby being susceptible to digital injection attack, \eg, real-time interactive deepfakes emerges recently~\citep{horvitz2022horizon}. Alternative strategies integrate watermarks~\citep{cox2007digital} into data originating from authentic cameras, enabling recognition by cloud servers; Trusted Execution Environments \citep{jauernig2020trusted} offers a means to guarantee the secure execution of data in-use. \citet{ma2022totems} introduces a heightened level of complexity to the physics within the scene, thereby enhancing the challenge of real-time synthesis.

\section{The Security Analysis on Single-Frame Passive Face Anti-spoofing}
\label{sec:security_analsis_on_passive_face_liveness_detection}
There is a prevailing trend within the research community to employ single-image-based passive face anti-spoofing models for their simplicity and reduced user friction. Nonetheless, when integrated as the security component in a biometric system, they inherently introduce greater vulnerability compared to more complex approaches.
To quantitatively demonstrate this vulnerability, we conducted a series of experiments by evaluating the recent publicly available single-image-based presentation attack models on the public datasets in the different settings based on our threat model.

\subsection{Experimental Settings}
\subsubsection{Datasets \& Metrics}
The experiments conducted in various scenarios make use of two primary public datasets:
\begin{itemize}[leftmargin=*]
    \item \textbf{SiW}\citep{liu2018learning}: The dataset contains a collection of videos featuring 165 subjects, including both live and spoof recordings. Each subject has a total of 8 live videos and a maximum of 20 spoof videos, resulting in a dataset comprising 4,478 videos in total. All videos are captured at a frame rate of 30 frames per second (fps), with an approximate duration of 15 seconds and a resolution of 1080P HD. The live videos were recorded during four sessions, exhibiting variations in distance, pose, illumination, and expression. On the other hand, the spoof videos were gathered using various attack methods such as printed paper and replay.
    \item \textbf{OULU-NPU}\citep{OULU_NPU_2017}: This dataset encompasses 4950 videos that include both genuine access and attack scenarios. These videos were captured utilizing the front cameras of six mobile devices, across three sessions featuring distinct illumination conditions and background scenes. The OULU-NPU database focuses on two types of presentation attack: print and video-replay. These attack were generated by employing two printers and two display devices.
    \item \textbf{DeeperForensics} \citep{jiang2020deeperforensics}: DeeperForensics is a significant dataset for detecting real-world face forgeries, comprising 60,000 videos and 17.6 million frames. The source videos were gathered in a controlled laboratory setting, while the target videos were derived from FaceForensics++ \citep{rossler2019faceforensics++} to create a substantial volume of attack data using the face-swapping method proposed by the authors.
    %\item \textbf{DFDC} \citep{dolhansky2020deepfake}: Deepfake detection challenge is created by Meta with eight facial modification algorithms. The full dataset consists of 124K videos and the preview dataset consists of 5k videos.
    % \item \textbf{Stable-Diffusion}: This dataset is generated containing 320 images using Stable Diffusion v1.5 \citep{rombach2022high}.
\end{itemize}
In the following experiments, we will use the APCER, BPCER, ACER, FRR@FAR, and FAR@FRR (Table~\ref{tab:glossary}) as the metrics to evaluate the performance of the model in each setting.

\subsubsection{Benchmark Models}
The current face presentation attack detection models in the existing literature mainly work with the single frame or image passively.
We chose three recent public available face presentation attack detection models to perform the experiments:
\begin{itemize}[leftmargin=*]
    \item \textbf{CDCN} \citep{yu2020searching} is introduced to capture fine-grained information from the input image. Since the model pre-trained weights is not provided, we used the publicly available code to train model with the default setting on training set from the OULU-NPU dataset.
    \item \textbf{MDFAS} \citep{guo2022multi} is introduced to tackles the issue of forgetting when incorporating new domain data, all while maintaining a high level of adaptability. To perform the experiments, we used the public available tensorflow-based model to perform the experiments.
    \item \textbf{SAFAS} \citep{sun2023rethinking} is proposed to encourage domain separability while aligning the live-to-spoof transition to be the consistent for all domains. Due to non-existent of the pre-trained model, three datasets (CASIA-MFSD, MSU-MFSD, IDAIP-Replay) are used to train the model based on the public available code repository.
    \item \textbf{CADDM} \citep{dong2023implicit} is a recent deepfake detection approach that has designed an artifact detection module and multi-scale facial swap to improve the deepfake detection performance.
    %\item \textbf{TruFor} \citep{guillaro2023trufor} is one of recent image forensics frameworks that extracts both high-level and low-level traces through a transformer-based fusion architecture that combines the RGB image and a learned noise-sensitive fingerprint to detect and localization the forgeries.
\end{itemize}
For public available models, all the RGB data is preprocessed by the preprocessing function provided by Sun \etal \citep{sun2023rethinking} and the depth map is estimated by PRNet \citep{feng2018joint}.
In addition to these public models, we trained another four models based on vallina ResNet-50 \citep{he2016deep} by replacing the last fully connected layer and initializing using ImageNet pretrained weights \citep{deng2009imagenet} as follows:
\begin{itemize}[leftmargin=*]
    \item \textbf{R50-OULU}: The model is trained using OULU-NPU dataset \citep{OULU_NPU_2017} only for protecting against presentation attacks only, which achieved o AUROC of 99.97\% on OULU testset;
    \item \textbf{R50-FF}: The model is trained using FaceForensics++ \citep{rossler2019faceforensics++} only for protecting against Deepfake attacks only, which achieved a AUROC of 99.46\% on the FaceForensics++ testset;
    \item \textbf{R50-ONLU-FF}: The model is trained using both OULU-NPU dataset \citep{OULU_NPU_2017} and FaceForensics++ \citep{rossler2019faceforensics++} for protecting against presentation and deepfake attacks.
    \item \textbf{R50-ONLU-FF-FUSE}: The individual models R50-OULU and R50-FF are used to inference to predict live scores which are aggregated to generate a final score. The default aggregation function we used is simply score production.
\end{itemize}
In the evaluation, we used the threshold that achieve the best ACER as the decision threshhold and used the sigmoid function calibrated on OULU-NPU dataset to map the decision threshold close to 0.5.

\begin{table*}[thb]
\caption{The performance of the presentation attack detection and deepfake attack detection models tested on OULU-NPU Dataset. Note that the metrics in this table are in percentage. PA presents Presentation Attacks and DA presents Deepfake Attacks.}
\label{tab:oulu}
\begin{adjustbox}{width=\linewidth}
\begin{tabular}{l|c|c|ccc|ccccc|ccc}
\toprule%
\multirow{2}{*}{Method} & \multirow{2}{*}{Scenario} & \multirow{2}{*}{Target} & \multirow{2}{*}{APCER} & \multirow{2}{*}{BPCER} & \multirow{2}{*}{ACER} & \multicolumn{5}{c|}{FRR@FAR=} & \multicolumn{3}{c}{FAR@FRR=}\\ \cmidrule{7-11} \cmidrule{12-14}
& & & & & & 1E-3 & 5E-3 & 1E-2 & 5E-2 & 1E-1 & 1E-2 & 5E-2 & 1E-1\\
\midrule
CDCN   & In-domain & PA & 11.2 & 6.4 & 8.8 & 74.2 & 55.6 & 39.7 & 16.9 & 8.9 & 31.5 & 15.0 & 9.1\\
MDFAS  & In-domain & PA & 4.2 & 1.4 & 2.8 & 57.5 & 12.8 &7.5 & 1.1 & 0.6 & 4.7 & 1.3 & 0.6\\
SAFAS  & Out-domain & PA & 15.1 & 6.9 & 11.0 & 65.0 & 57.5 & 45.8 & 22.2 & 13.6 & 31.9 & 17.7 & 12.7\\ 
R50-OULU & In-domain & PA & 1.5 & 0.0 & 0.7 & 8.3 & 1.9 & 0.8 & 0.0 & 0.0 & 0.8 & 0.2 & 0.0 \\
\midrule
CADDM  & Out-domain & DA &  64.2& 28.1 & 46.1 & 98.9 & 98.1 & 97.5 & 91.4 & 85.3 & 97.4 & 91.9 & 85.6\\
%TruFor  & Out-domain & DA &  &  &  &  &  & \\ 
R50-FF & Out-domain & DA & 38.3 & 48.3 & 43.3 & 76.9 & 76.9 & 76.9 & 76.9 & 76.9 & 92.9 & 89.3 & 81.5\\
\midrule
R50-OULU-FF & In-domain & PA, DA & 0.6 & 2.8 & 1.7 & 9.2 & 3.3 & 2.8 & 2.8 & 0.0 & 2.8 & 0.3 & 0.1\\
%\midrule
R50-OULU-FF-FUSE & In-domain & PA, DA & 1.5 & 0.0 & 0.7 & 9.2 & 1.9 & 0.8 & 0.0 & 0.0 & 0.8 & 0.2 & 0.1\\
\botrule
\end{tabular}
\end{adjustbox}
\end{table*}

\begin{table*}[ht]
\caption{The performance of the presentation attack detection and deepfake attack detection models tested on SiW Dataset. Note that the metrics in this table are in percentage. PA presents Presentation Attacks and DA presents Deepfake Attacks.}
\label{tab:siw}
\begin{adjustbox}{width=\linewidth}
\begin{tabular}{l|c|c|ccc|ccccc|ccc}
\toprule %
\multirow{2}{*}{Method} & \multirow{2}{*}{Scenario} & \multirow{2}{*}{Target} & \multirow{2}{*}{APCER} & \multirow{2}{*}{BPCER} & \multirow{2}{*}{ACER} & \multicolumn{5}{c|}{FRR@FAR=} & \multicolumn{3}{c}{FAR@FRR=}\\ \cmidrule{7-11} \cmidrule{12-14}
& & & & & & 1E-3 & 5E-3 & 1E-2 & 5E-2 & 1E-1 & 1E-2 & 5E-2 & 1E-1\\
\midrule
CDCN   & Out-domain & PA & 16.1 & 46.2 & 31.2 & 100.0 & 99.7& 99.5 & 90.7 &64.94 & 86.4 & 69.5 & 53.5 \\
MDFAS  & In-domain & PA & 1.2 & 1.0 & 1.1 & 8.4 & 2.3 &1.4 & 0.3 &0.1 & 1.2 & 0.3 & 0.1\\
SAFAS  & Out-domain & PA & 1.0 & 25.4 & 13.2 & 50.1 & 34.9 &25.8 & 13.5 & 9.8 & 78.6 & 26.8 & 9.2\\ 
R50-OULU & Out-domain & PA & 5.0 & 6.3 & 5.6 & 37.2 & 20.4 & 16.4 & 6.3 & 4.4 & 29.4 & 7.2 & 2.8\\
\midrule
CADDM  & Out-domain & DA & 100.0 & 0.0 & 50.0 & 100.0 & 99.9 & 99.6 & 96.4 & 92.9 & 99.8 & 98.5 & 96.5\\
%TruFor  & Out-domain & DA &  &  &  &  &  & \\ 
R50-FF & Out-domain & DA & 61.4 & 25.7 & 43.6 & 77.1 & 77.1 & 77.1 & 77.1 & 77.1 & 95.3 & 87.4 & 79.8\\ \midrule
R50-OULU-FF & Out-domain & PA, DA & 6.6 & 10.0 & 8.3 & 50.0 & 32.6 & 26.0 & 12.4 & 6.9 & 26.9 & 12.4 & 6.6\\
R50-OULU-FF-FUSE & Out-domain & PA, DA & 4.9 & 6.3 & 5.6 & 42.7 & 22.2 & 16.6 & 6.3 & 4.4 & 29.3 & 7.1 & 2.8\\
\botrule
\end{tabular}
\end{adjustbox}
\end{table*}

\subsection{Presentation Attack}
\label{sec:exp_presentation_attack_detection}
Tables \ref{tab:oulu} and \ref{tab:siw} provides a comprehensive view of the defense capabilities against presentation attack, evaluated across various scenarios:
We examine the performance of CDCN, R50-OULU, and R50-OULU-FF in the in-domain setting, where both the training and testing data originate from the same dataset. 
Conversely, SAFAS is assessed in an out-domain scenario due to the publicly available model was trained using four datasets other than OULU-NPU dataset. 
The evaluation marks the instances where there is a disconnect between the training and testing datasets, categorizing them as out-domain. 
As for MDFAS, it is designated as unknown due to limited information about its training set.

It is worth noting that these models exhibit significant variations in performance across these diverse scenarios. CDCN, in particular, demonstrates subpar performance overall, especially when tested in the out-domain setting on the SiW dataset. Notably, employing a calibrated threshold yields an error rate (APCER) as high as 10\%+ in both datasets (in-domain and out-domain), which raises concerns.

Similarly, SAFAS exhibits a high rejection rate, primarily because its threshold is computed based on the OULU-NPU dataset. 
Regrettably, these models currently struggle to achieve high accuracy in detecting presentation attack using a single image.
Moreover, this is the model that target to improve the domain generalization but observed limited robustness to the unknown domains, which raises another concern.

Even for MDFAS, which has an unknown training set, the APCER can vary from 1\% to 4\% depending on the test set. This suggests that the performance of single-frame models is less precise and robust. Furthermore, it's noteworthy that attackers can potentially bypass the system by showing the presentation attack instrument to the system only a limited number of times (ranging from 25 to 100 times).

Interestingly, our vanilla trained model, R50-OULU, exhibits the best performance in the in-domain setting and remains relatively competitive in the out-domain setting. When comparing these four models across both datasets, several clear trends emerge:
\begin{enumerate*}
    \item In the in-domain setting, where the domain and attack vectors are the same as those seen during training, the model achieves its best performance;
    \item In the out-domain setting, which is often the case when the model is used for remote biometric verification systems, the performance of the model tends to decrease significantly, especially when encountering different domains and previously seen attack vectors.
\end{enumerate*}
These findings underscore the importance of collecting data from multiple sources to encompass various deployment environments and attack vectors. 
This approach is crucial to bridging the gap and enhancing the security level.

\begin{table*}[thb]
\caption{The performance of the presentation attack detection models tested on FaceForensics Dataset. Note that the metrics in this table are in percentage.}
\label{tab:faceforensics}
\begin{adjustbox}{width=\linewidth}
\begin{tabular}{l|c|c|ccc|ccccc|ccc}
\toprule %
\multirow{2}{*}{Method} & \multirow{2}{*}{Scenario} & \multirow{2}{*}{Target} & \multirow{2}{*}{APCER} & \multirow{2}{*}{BPCER} & \multirow{2}{*}{ACER} & \multicolumn{5}{c|}{FRR@FAR=} & \multicolumn{3}{c}{FAR@FRR=}\\ \cmidrule{7-11} \cmidrule{12-14}
& & & & & & 1E-3 & 5E-3 & 1E-2 & 5E-2 & 1E-1 & 1E-2 & 5E-2 & 1E-1\\
\midrule 
R50-OULU & Out-domain & DA & 31.3 & 65.7 & 48.5& 99.6& 99.1& 98.6& 93.7& 88.2& 99.4& 97.1& 91.3\\
\midrule
CADDM & In-domain & DA & 0.2 & 0.9 & 0.6 & 1.1 & 0.9 & 0.8 & 0.5 & 0.3 & 0.2 & 0.0 & 0.0\\
R50-FF & In-domain & DA & 2.7 & 0.9 & 1.8 & 81.4& 36.0 & 15.1 & 0.0 & 0.0 & 2.6 & 1.7& 1.3\\
\midrule
R50-OULU-FF & In-domain & PA, DA & 1.9& 0.3& 1.1 & 86.8& 54.1& 37.4& 0.0& 0.0& 1.5& 1.4& 1.4\\
R50-OULU-FF-FUSE & In-domain & PA, DA & 15.3& 15.6& 15.5 & 80.6& 68.6& 64.6& 37.3& 22.7& 66.6& 48.1& 25.2\\
\botrule
\end{tabular}
\end{adjustbox}
\end{table*}

\begin{table*}[thb]
\caption{The performance of the presentation attack detection models tested on DeeperForensics Dataset (Standard set). Note that the metrics in this table are in percentage.}
\label{tab:deeperforensics}
\begin{adjustbox}{width=\linewidth}
\begin{tabular}{l|c|c|ccc|ccccc|ccc}
\toprule %
\multirow{2}{*}{Method} & \multirow{2}{*}{Scenario} & \multirow{2}{*}{Target} & \multirow{2}{*}{APCER} & \multirow{2}{*}{BPCER} & \multirow{2}{*}{ACER} & \multicolumn{5}{c|}{FRR@FAR=} & \multicolumn{3}{c}{FAR@FRR=}\\ \cmidrule{7-11} \cmidrule{12-14}
& & & & & & 1E-3 & 5E-3 & 1E-2 & 5E-2 & 1E-1 & 1E-2 & 5E-2 & 1E-1\\
\midrule 
SAFAS & Out-domain & PA & 22.3 & 39.0 & 30.6 & 95.6 & 88.6 & 83.5 & 65.8 & 54.9 & 95.0 & 77.5 & 63.3\\ 
R50-OULU & Out-domain & PA & 91.7 & 5.6 & 48.6 & 99.9 & 99.7 & 99.5 & 97.0 & 94.0 & 97.2 & 92.2 & 88.0\\
\midrule
CADDM  & Out-domain & DA & 42.3 & 48.3 & 45.3 & 100.0 & 99.9 & 99.6 & 96.5 & 91.0 & 99.0 & 94.9 & 89.2\\
%TruFor & Out-domain & DA &  &  &  &  &  & \\ 
R50-FF & Out-domain & DA & 13.3 & 79.5 & 46.4 & 95.6 & 95.6 & 95.6 & 89.4 & 83.0 & 98.8 & 91.8 & 88.0\\
\midrule
R50-OULU-FF & Out-domain & PA, DA & 99.8 & 0.0 & 49.9 & 99.9 & 99.4 & 99.2 & 97.6 & 96.1 & 99.7 & 99.3 & 98.8\\
R50-OULU-FF-FUSE & Out-domain & PA, DA & 91.6 & 5.6 & 48.6 & 99.8 & 99.5 & 99.2 & 96.7 & 93.8 & 97.2 & 92.1 & 87.9\\
\botrule
\end{tabular}
\end{adjustbox}
\end{table*}

\subsection{Digital Injection Attacks}
We assess the threat of digital injection attacks through three types of experimental setups, each designed to evaluate the models' security against deepfake, digital adversarial attack, and digital replay attack.
Given that these threat vectors target different aspects of the model's robustness, they necessitate distinct experimental setups. These setups are detailed in the subsequent subsections.

\subsubsection{Deepfakes Attack}
\label{sec:exp_deepfake_detection}

Tables~\ref{tab:faceforensics} and~\ref{tab:deeperforensics} present the performance of evaluated models on the Face Forensics dataset (using the raw protocol) and the Deeper Forensics dataset (using the standard protocol), respectively. Similar to findings in presentation attack detection, the models (CADDM and R50-FF) perform well in in-domain settings but show significant performance drops in out-domain settings. 
This indicates that models trained on a single domain might have security vulnerabilities against novel deepfake attacks.

To assess robustness of the models in cross-attack scenarios, we trained models on either presentation attack detection or deepfake attack detection datasets and then tested them on the alternate dataset. 
In Tables~\ref{tab:oulu} and~\ref{tab:siw}, the deepfake detection models (CADDM and R50-FF) achieved an ACER close to 50\%, indicating that models trained for deepfake attack detection do not generalize well to presentation attacks. 
A similar observation is made for presentation attack detection models (SAFAS and R50-OULU), as shown in Tables~\ref{tab:faceforensics} and~\ref{tab:deeperforensics}, where their performance in deepfake detection contexts is also significantly regressed.
This observation is aligned well with the findings from \citet{li2022seeing}, who has evaluated the robustness of the existing commercial face anti-spoofing system in the market by injecting the face-swapped samples as Deepfakes attack. 

To showcase the effectiveness of including a broad range of threats in model training, we displayed the experimental outcomes for R50-OULU-FF and R50-OULU-FF-FUSE. Several notable observations emerged:
\begin{itemize}[leftmargin=*]
    \item R50-OULU-FF exhibited slightly inferior performance compared to the individual models trained on separate datasets (R50-OULU and R50-FF), as evidenced by its ROC curve.
    \item R50-OULU-FF-FUSE showed improved results on the presentation attack detection datasets but significantly underperformed on the FaceForensics dataset.
    \item No model performed well on the unseen deepfake attack (Table~\ref{tab:deeperforensics}), which raised significant security concern on the existing deepfake detection models.
\end{itemize}
We believe the findings still hold for other unseen advanced generative models such as stable diffusion \citep{rombach2022high}.

% Finally, we evaluate another category of Deepfakes named whole image synthesis as additional advanced unseen face synthesis threat to the defense model. 
% In the era of AIGC, the attacker could ultilize the public available models to generate such attack content and inject into the face anti-spoofing system.
% To perform such attack, we generated a limited dataset comprising 320 images using Stable Diffusion v1.5 \citep{rombach2022high} and input them to evaluating models to generate liveness scores. 
% In Table \ref{tab:stable_diffusion}, it becomes evident that models designed specifically for countering presentation attack prove ineffective in preventing digital deepfakes attack. This observation aligns with findings in the literature \citep{li2022seeing}, where it was shown that even face-swap deepfakes attack pose a challenge to existing presentation attack models.
% Interestingly, our verification process revealed that Stable Diffusion can achieve an alarmingly high attack success rate of up to 84.44\%, while maintaining a 10\% false rejection rate. 
% This outcome underscores that existing presentation attack models lack the capacity to defend against images generated through stable diffusion, often mis-classifying them as genuine user inputs.

Based on the experimental findings in Sec.~\ref{sec:exp_presentation_attack_detection} and~\ref{sec:exp_deepfake_detection}, it is clear that face anti-spoofing models based on single-frame analysis requires more diverse data coverage and careful model design. 
This enhancement is necessary to bridge the performance gap between in-domain and out-domain settings, as well as between known and unknown attack vectors, ensuring both accuracy and robustness.

\begin{figure*}[htbp]
    \centering
    \subfloat[PGD attack]{%
      \includegraphics[width=\textwidth]{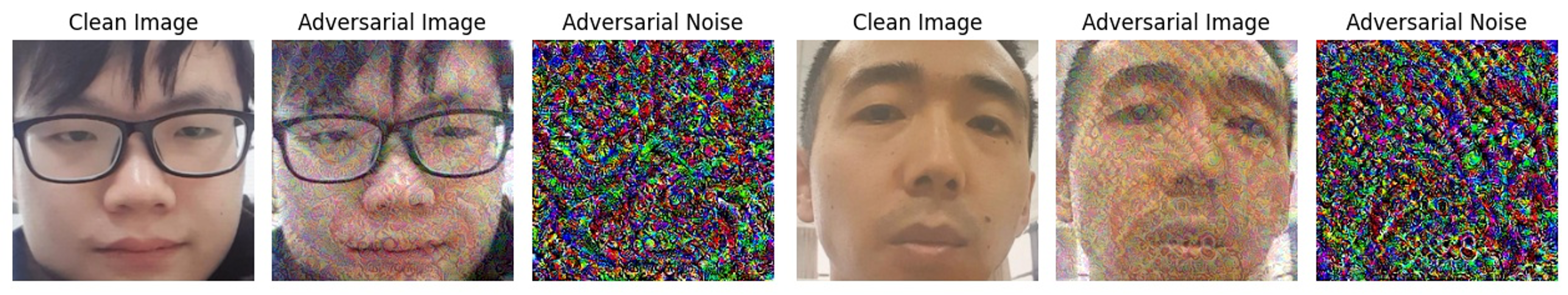}
    }
    \hfill
    \subfloat[Simple black-box attack]{%
      \includegraphics[width=\linewidth]{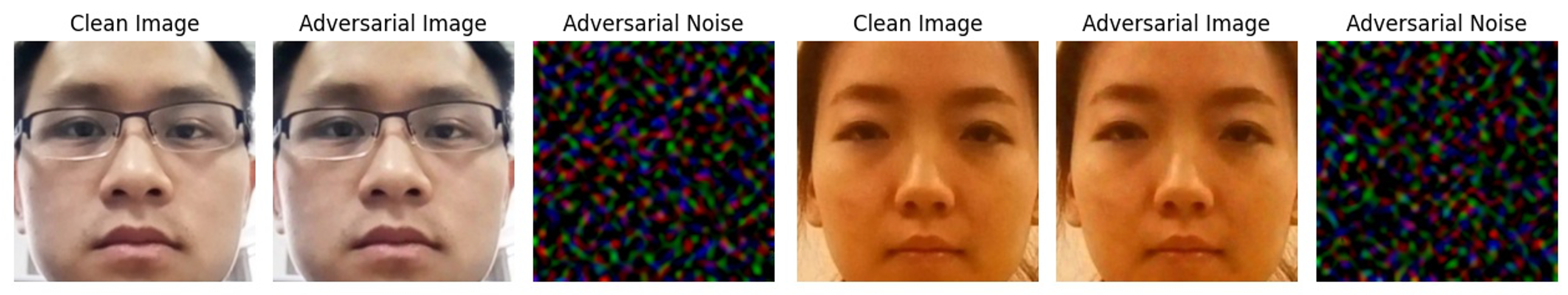}
    }
    \caption{Adversarial attack illustration by (a) PGD attack and (b) Simple black-box attack, in the order of clean spoof image, generated adversarial image, and generated adversarial noise. 
    }
    \label{fig:adversarial_samples}
\end{figure*}

\subsubsection{Digital Adversarial Attack}
In this subsection, we use the adversarial noise as the one attack representation of the digital adversarial samples in three scenarios: 
\begin{enumerate}[leftmargin=*]
    \item White-box setting: the attacker have the access to the face anti-spoofing model;
    \item Black-box setting: the attacker only can query the server model and get face anti-spoofing score from it;
    \item Black-box setting with a surrogate model: the attacker can have a surrogate model running in the local and use that to generate adversarial noise to attack the server model.
\end{enumerate}
In our setup, we employed presentation attack detection models as the victim models on the server, which demonstrated reasonable performance in detecting presentation attack mediums when tested on the OULU dataset, as shown in Table~\ref{tab:oulu}. 
Our objective is to flip the model's prediction by modifying the input image using a presentation medium, such as print or physical replay, in a limited number of queries.

To simulate the adversarial attack in both open-box and black-box settings, we chose PGD attack \citep{madry2017towards} and Simple Black-box Attack (Simba) \citep{narodytska2017simple} as the attack methods and evaluate the robustness of the presentation attack detection models on OULU-NPU dataset.
Figure~\ref{fig:adversarial_samples}(a) shows the adversarial noise generated by PGD attack with the epsilon control of 8 pixels and 10 queries to the model while Figure~\ref{fig:adversarial_samples}(b) shows the adversarial noise generated by Simba with 100 queries to the model. 
The experimental results depicted in the Table~\ref{tab:did_adv_pgd} and Table~\ref{tab:did_adv_simba} indicate that the existing public available models are vulnerable to the adversarial samples, which has been limited explored in face anti-spoofing literature.
By adding such adversarial noise to the clean spoof image, the score from the model could be flipped easily from spoof to real. 

In the open-box setting, PGD attack can generate adversarial noise extremely easy (APCER increased from 15.14\% to 93.05\%) on the resnet-18-based SAFAS with only a single query.
CDCN shows some a better resistance than SAFAS to PGD attack with it's special central difference convolutional operation but would not be robust enough when the number of queries increases to 10 (APCER increases from 11.18\% to 40.76\%).

Interestingly, in the black-box setting, Simba only need to query once to the CDCN model and can generate the adversarial examples that can increase APCER from 11.18\% to 36.11\%.
When the number of queries reaches to 1000, the APCER can be increased up to 62.29\%.
Compared with CDCN, SAFAS is robust to Simba, which only lead APCER increase from 15.15\% to 22.85\%.

\begin{table}[htbp]
\caption{The performance of the presentation attack detection models in white-box setting tested on adversarial samples generated by PGD attack using OULU-NPU dataset.}
\label{tab:did_adv_pgd}
%\begin{adjustbox}{width=\textwidth}
%\begin{tabular}{l|c|ccc|ccccc|ccc}
\begin{tabular}{l|c|ccc}
\toprule %
Method & \#Queries & APCER & BPCER & ACER \\
%\multirow{2}{*}{Method} & \multirow{2}{*}{\#Queries} & \multirow{2}{*}{APCER} & \multirow{2}{*}{BPCER} & \multirow{2}{*}{ACER} & \multicolumn{5}{c|}{FRR@FAR=} & \multicolumn{3}{c}{FAR@FRR=}\\ \cmidrule{6-10} \cmidrule{11-13}
%& & & & & 1E-3 & 5E-3 & 1E-2 & 5E-2 & 1E-1 & 1E-2 & 5E-2 & 1E-1\\
\midrule
\multirow{2}{*}{CDCN} & 1 & 14.5 & 6.1 & 10.3 \\
   & 10 & 40.8 & 6.1 & 23.4 \\
\midrule
\multirow{2}{*}{SAFAS} & 1 & 93.1 & 6.7 & 49.9 \\
   & 10 & 100.0 & 6.7 & 53.3 \\
%\midrule
% \multirow{2}{*}{R50} & 1 &  &  & \\
%    & 10 & & & \\
\botrule
\end{tabular}
%\end{adjustbox}
\end{table}

\begin{table}[htbp]
\caption{The performance of the presentation attack detection models in black-box setting tested on adversarial samples generated by Simba using OULU-NPU dataset.}
\label{tab:did_adv_simba}
%\begin{adjustbox}{width=\linewidth,center}
%\begin{tabular}{l|c|ccc|ccccc|ccc}
\begin{tabular}{l|c|ccc}
\toprule %
Method & \#Queries & APCER & BPCER & ACER \\
%\multirow{2}{*}{Method} & \multirow{2}{*}{\#Queries} & \multirow{2}{*}{APCER} & \multirow{2}{*}{BPCER} & \multirow{2}{*}{ACER} & \multicolumn{5}{c|}{FRR@FAR=} & \multicolumn{3}{c}{FAR@FRR=}\\ \cmidrule{6-10} \cmidrule{11-13}
%& & & & & 1E-3 & 5E-3 & 1E-2 & 5E-2 & 1E-1 & 1E-2 & 5E-2 & 1E-1\\
\midrule
\multirow{4}{*}{CDCN} & 1 & 36.1 & 6.1 & 21.1 \\
   & 10 & 45.6 & 6.1 & 25.8 \\
   & 100 & 57.2 & 6.1 & 31.7\\
   & 1000 & 62.3 & 6.1 & 34.2\\
\midrule
\multirow{4}{*}{SAFAS} & 1 & 15.1 & 6.9 & 11.0\\
   & 10 & 16.9 & 6.9 & 11.9 \\
   & 100 & 17.8 & 6.9 & 12.6 \\
   & 1000 & 22.9 & 6.9 & 14.9 \\
%\midrule
% \multirow{4}{*}{R50} & 1 &  & & \\
%    & 10 & & & \\
%    & 100 & & & \\
%    & 1000 & & & \\
\botrule
\end{tabular}
%\end{adjustbox}
\end{table}

\begin{table}[htbp]
\caption{The performance of the presentation attack detection models (Vic. Model) in the black-box setting with the surrogate model (Sur. Model) tested on adversarial samples generated by Simba using OULU-NPU dataset.}
\label{tab:did_adv_transfer}
%\begin{adjustbox}{width=\linewidth,center}
%\begin{tabular}{l|c|c|ccc|ccccc|cc}
\begin{tabular}{l|c|ccc}
\toprule %
Victim  & Surrogate & APCER & BPCER & ACER \\
%\multirow{2}{*}{Sur. Model} & \multirow{2}{*}{Vic. Model} & \multirow{2}{*}{Dataset} & \multirow{2}{*}{APCER} & \multirow{2}{*}{BPCER} & \multirow{2}{*}{ACER} & \multicolumn{5}{c|}{FRR@FAR=} & \multicolumn{2}{c}{FAR@FRR=}\\ \cmidrule{7-11} \cmidrule{12-13}
%& & & & & & 1E-3 & 5E-3 & 1E-2 & 5E-2 & 1E-1 & 1E-2 & 5E-2 \\
\midrule
\multirow{3}{*}{CDCN} & SAFAS & 16.53 & 6.94 & 11.04 \\
 & MDFAS & 13.26 & 1.39 & 7.33 \\
 % & R50 & \\
 \midrule
\multirow{3}{*}{SAFAS} & CDCN & 72.43 & 6.39 & 39.41 \\
 & MDFAS & 17.78 & 1.39 & 9.58 \\
 % & R50 & \\
\botrule
\end{tabular}
%\end{adjustbox}
\end{table}

In the third setting, we can generate adversarial samples using the Simba on a surrogate model and test them on the target model. 
Since in this experimental setting, the attacker can do the perturbation as much as he can, we use the adversarial samples after 100 steps, which achieved 100\% APCER.
Table~\ref{tab:did_adv_transfer} shows that the adversarial samples generated using a strong surrogate model can achieve high transferability rate on a weaker victim model (SAFAS as surrogate model and CDCN as the victim model) while achieve limited attack successful rate on a even stronger victim model (CDCN as surrogate model and the other two as the victim model).

%artificat detection expecially presentation, digital deepfake, adversarial, and raw image.

\begin{table}[htbp]
\caption{The performance of the presentation attack detection models tested on OULU-NPU Dataset.}
\label{tab:dra_oulu}
\begin{tabular}{l|c|ccc}
\toprule %
Method & Scenario & APCER & BPCER & ACER\\
\midrule
CDCN   & In-domain & 93.61 & 6.39 & 50.00\\
MDFAS  & Unknown & 98.91 & 1.39 & 50.00\\
SAFAS  & Out-domain & 93.06 & 6.94 & 50.00\\
\botrule
\end{tabular}
\end{table}

\subsubsection{Digital Replay Attack}
In this experiment, we aimed to simulate a digital replay attack by sending real images to the model twice assuming the first-time face anti-spoofing check is real and the second-time check is attack (The raw image can be found online or captured in real life). 
%The first set of images provided to the system was genuine, while the second set was a part of the attack. 
Due to the stateless nature of the existing single-frame passive model, it produced identical scores for the same image when validated in both instances. As a result, we observed alarmingly high error rates, as illustrated in Table \ref{tab:dra_oulu}.
Considering that the model's primary objective is to detect artifacts or irregularities in the presented image, a digital replay attack, as described, poses a particularly insidious threat. This is because such an attack doesn't introduce any discernible artifacts or abnormalities, making it difficult for the model to distinguish between genuine and replayed images. 
What makes this threat even more concerning is the relative ease with which an attacker can obtain a victim's face image, often available through various means. 
Consequently, this type of attack could indeed be one of the most harmful, exploiting the model's limitations and bypassing its security measures without leaving any telltale signs.

\subsection{Discussion}
From the experiments above, it becomes evident that single image-based face anti-spoofing models, while offering an excellent user experience, have significant vulnerabilities under the common threats in the face face anti-spoofing as discussed in Sec.~\ref{sec:threat_model}. These vulnerabilities make them less suitable for scenarios where security is of paramount concern.
The foundation issue is that these models only check the artifacts from the image, which is less reliable and can be easily perturbed or eliminated for miss-detection and do not have ability to determine the subject is actually \textbf{physically} closed to the sensor and \textbf{actively} operating the system.
This fundamental issue of underlying the existing passive models make them vulnerable to out-domain or unseen presentation attack from the physical world and various attacks from digital world.

Considering their simplicity and minimal user friction, these passive face anti-spoofing methods find their most suitable applications in scenarios where hardware integrity can be ensured, and low-value transactions are involved. Examples include public kiosks within surveillance environments, where the risk of fraudulent activities is relatively low, and the primary focus lies on user convenience and accessibility.
While some researchers are exploring the integration of presentation and deepfake detection into a unified model \citep{deb2023unified}, we contend that, based on insights gained from prior deepfake detection studies \citep{zhao2021learning,liu2021spatial,huang2023implicit,dong2023implicit}, single-image-based models are still susceptible to novel or unseen attacks. This susceptibility is attributed to the following reasons:
First, during the model training phase, it is impractical to cover an infinite number of combinations of hyper-parameters and architectures of known generative models.
Second, in the stage of testing, artifacts can be manipulated through various data compression techniques or post-processing functions applied by the generative approach and potentially could be completely different if it is from the new proposed generative models.
video-based methods may introduce additional complexity in the temporal dimension, but it remains a matter of time and processing power for attackers to compromise them, given the continued advancement of deepfake and video generation models. Furthermore, all these models are generally ineffective in protecting against digital replay attacks.

\section{The Best Practices and Principles in Face Anti-spoofing}
\label{sec:best_practice_principles_in_liveness_detection}
As the discussion in the preceding sections shows, face anti-spoofing is a complex multi-stage process with information flowing from objects in the physical scenes to models running in a cloud service. 
Some of these stages, such as the physical scene and the sensed signal channel, are exogenous to the system and therefore inherently more vulnerable to adversarial actions. 
Others are endogenous to the system, and thus can be hardened against adversarial actions via appropriate engineering employing methods from cryptography, trusted computing enclaves, tamper-resistant packaging, electro-magnetic shielding, \textit{etc}. 
Given the complexity of face anti-spoofing task, there is no silver bullet for creating a trustworthy face anti-spoofing system and instead it is important to take a multi-layered end-to-end approach while distinguishing between problems that present fundamental challenges vs. those that arise from inadequate engineering practice. 
Below we distill the insights from the earlier sections into a set of principles for guiding the design of rubost face anti-spoofing systems from model, machine learning (ML) system, and platform perspectives.

\subsection{Model Performance}

\noindent \textbf{Data Coverage}: 
A sophisticated face anti-spoofing system comprises multiple components and functionalities designed to address a wide array of potential attacks spanning both physical and digital realms. Achieving comprehensive data coverage in both training and testing phases is essential for ensuring the highest level of security in such systems \citep{yang2019face}.
First, data coverage on various devices and attack vectors plays a pivotal role in the accuracy of deep learning models underpinning anti-spoofing systems, as these models rely heavily on the data they are trained on. 
Second, ensuring broad coverage across various domains, ethnicities, and environmental conditions is crucial for model fairness and robustness across different demographic groups and device configurations.
Third, in addressing the spectrum of attack classes, existing literature suggests that unseen attack types consistently exhibit lower accuracy compared to known attack categories \citep{yu2022deep,le2024sok}. 
While designing a more sophisticated model specifically for generalizing across unseen attack categories is helpful, it is not comparable to training directly with those unseen attack classes \citep{chen2022ost,liu2023towards}. Hence, prioritizing exhaustive data coverage is paramount, as face anti-spoofing systems demand the highest level of security with little tolerance for bypass attempts, exceeding the security requirements of other application scenarios.
At last, for evaluation purposes, it is imperative to gather a diverse range of attack classes to simulate real-world scenarios comprehensively.
With that being said, it is beneficial to expand the red team \citep{diogenes2018cybersecurity} network to attack the system to identity any potential risks before model release and understand the model behavior. 
This allows for stress testing of the anti-spoofing system and demonstrates its effectiveness in handling various types of attacks.

\noindent \textbf{Modeling}:
Another aspect to improve the accuracy in general is on modeling including improving better model architecture, selecting compatible training strategy, and training more robust model weights.
Model architecture has been developed several revolutions: from literature, the models mainly are based on the backbone of ResNet \citep{he2016deep} and EffientNet \citep{tan2019efficientnet}. 
Recent trend is to build the model based on the ViT-based architectures \citep{han2022survey} on pure image or clip-based model \citep{radford2021learning} with additional text supervision.
Training strategy is associated with the dataset, model, and objective. 
In general, a suitable training strategy can benefit the model accuracy or robustness depending on the objective function.
The recent foundational models \citep{radford2021learning, li2023multimodal} with large capacity trained on the internet scaled multi-modality data shows the potentials to be applied to a broad spectrum of tasks \citep{liu2023fm, sarafianos2019adversarial} and to achieved improved accuracy.

\noindent \textbf{Multi-Modality}:
On presentation attack detection, for instance, additional sensors like depth and infrared cameras can enrich the data pool. 
Incorporating multi-modal sensors, beyond the traditional single RGB cameras, can enhance security as indicated by research such as \citep{zhang2020casia,liu2023fm}. 
On deepfake detection, recent work \citep{zhou2021joint, yang2023avoid} utilize audio and visual track to determine the deepfakes and \citet{shao2023detecting} detect and ground the medium manipulation on both text and vision domains.
This strategy is predicated on the notion that multi-modality complicates and increases the cost for adversaries aiming to tamper with data streams, especially in less common modalities. 
By utilizing data from various sensors, systems can perform more robust face anti-spoofing and cross-modality verification. 
However, this integration of multiple sensors may elevate the cost and complexity, potentially restricting their adoption to only high-end mobile devices (\eg, laptops and desktops often lack such advanced sensory capabilities).

\noindent \textbf{Temporal Dimension}: Existing digital and physical spoof detection models have primarily concentrated on employing single images as input, leveraging 2D spatial artifacts. However, a limited number of deepfake models have explored the use of multiple frames to capture sequential inconsistencies as a key indicator for enhancing detection accuracy and generalizability \citep{gu2021spatiotemporal,zhao2023istvt,yin2023dynamic}. Intrinsic to this endeavor is the extension of input data to the temporal dimension, transitioning from 2D to 3D, which significantly augments the information available for model analysis. Temporal inconsistencies and unrealism can unveil more artifacts than spatial variations, providing the model with additional cues for detection.
For instance, generating temporally consistent deepfakes necessitates additional considerations in the generation pipeline, posing a considerable challenge for attackers. Designing temporally coherent attack media requires substantial computational resources and effort \citep{liu2021generative}, introducing computation delays that hinder bypassing face anti-spoofing systems demanding real-time feedback.

In the context of physical attacks, prevailing methods predominantly adopt frame-level spoof detection, aggregating predictions from one or multiple frames to assess the spoofing score of a video instance. However, temporal geometric information, such as motion trajectories, offers a more universally applicable clue \citep{wang2020deep,chang2023closer}. For example, a paper-based attack commonly induces global motions such as translation or rotation.

Effectively harnessing temporal information presents challenges. Existing learning-based methods often focus on superficial spatial information, which may lead to over-fitting to spatial artifacts \citep{meyer2018improving,zhao2023istvt}, given their relative ease of capture. Nonetheless, there is considerable potential to enhance accuracy through spatial-temporal models, necessitating well-designed architectures to effectively utilize temporal information.

\subsection {Model Robustness}

The various digital and physical world attack described earlier all seek adversarial tampering of the digital data that is then input to the downstream machine learning models. In digital attack the adversary directly modifies the digitized biometric signal, while in physical world attack the adversary leverages physical proximity to tamper with the scene, signal, or the sensor so that the eventual digitized biometric signal is maliciously corrupted. 
While a combination of existing software and physical security measures such as trusted computing, attested applications, shielded sensors, tamper-resistant hardware, and human supervision can certainly help reduce or even prevent these attack, in many face anti-spoofing application settings such measures are not possible as the end devices are consumer-grade hardware such as smartphones and computers that are owned and operated by the potentially adversarial user in a space that they control. Moreover, technological advances are making it continually easier for adversaries to tamper with inputs to the downstream algorithms and machine learning models.
Some of the upstream subversive actions can be defended against via cryptographic and fingerprinting methods to authenticate the source device, and verify integrity and freshness of data received. 
In light of the preceding, it is important to have a multi-pronged strategy which not only harden the client device software against digital injection attack, the sensing mechanisms against physical-world attack, and crucially, hardens the machine learning models to adversarial inputs.

\noindent \textbf{Training data integrity}: One potential attack we did not mention in our threat model but may impact the model robustness is the injection of hidden backdoors in the model training data (\eg, via training data poisoning or tampering of model weights by a third party). 
It introduces vulnerability that is triggered only when specific adversarial inputs are presented during inference ~\citep{wang2019neural, gao2020backdoor, saha2020hidden}. 
face anti-spoofing systems that rely on models whose training cannot be trusted (\eg, foundation models from a model zoo, or models trained using third party data) must put in place suitable model testing procedures to check the integrity of the data.

\noindent \textbf{Robustness and Certification}: As shown earlier in the paper, the existing face anti-spoofing models are vulnerable to adversarial attack. 
While efficient defense against adversarial input attack remain very much an active area of research, several training-time and test-time approaches have emerged that face anti-spoofing systems can benefit from. 
Training time approaches seek to improve the model robustness by adversarial training \citep{bai2021recent,wong2020fast} or efficiently learn models whose output is provably robust to input perturbations up to certain magnitude \citep{raghunathan2018certified}.
Inference-time approaches seek to detect adversarial perturbations by looking for artifacts in the inputs \citep{metzen2016detecting}, deploying a image purification model to clean the data \citep{xiao2022densepure} or 
thwarting adversarial inputs via ensemble and moving target defenses \citep{song2022sardino}.

\noindent \textbf{Domain Generalization}: Another aspect of the model robustness is its generalization to unseen domains. 
The model can try to learn a generalized enough feature space for live categories and can detect the different image filtering styles of attack categories using domain adaptation \citep{xu2019d} or generalization techniques \citep{jia2020single, wang2022domain}.
Another specific approach is using test time context learning approach \citep{shu2022test} to provide vision prompts to the models and adjust the model behavior with such prompts.

\noindent \textbf{Anomaly detection}: 
Another approach is to integrate the anomaly detection into face anti-spoofing system \citep{perez2019deep,nikisins2018effectiveness,arashloo2017anomaly} to identify anomalous behaviors or pattern within the data given the fact that the common face face anti-spoofing system usually use predefined gesture (\eg. presenting the face in front of the camera). 
Anomalies within a video can manifest across both spatial and temporal dimensions, and it is anticipated that models should be capable of detecting these irregularities when the characteristics of the data in the stream deviate from those of authentic input.
Meanwhile, the abnormal pattern can be found in the network traffic or the data stream that targets to attack the models \citep{Ye2024}.

% \begin{figure}[htbp]
%   \centering
%   \includegraphics[width=\linewidth]{figures/category.png}
%   \caption{The category of three different type of face anti-spoofing system.\eric{this image does not deserve so much space}}
%   \label{fig:category_liveness_detection_system}
% \end{figure}

These methods mentioned above target to improve the model robustness in terms of data perturbation, unseen domains, unseen attacks.
However, it still requires additional effort to increase robustness of the ML pipeline to further protect the underlying models.

\subsection{ML Pipeline Robustness}
%In current face face anti-spoofing systems, we usually classify them into two main categories: passive and active approaches based on the type of challenge they pose.
We introduce a new methodology called ``proactive" by adding another dimension involving the sensor to enhance system security against the presentation attacks and digital injection attacks. 
Especially, it introduce the effectiveness that can detect the digital replay attacks that common face anti-spoofing cannot identify.
To explain this methology, we need to break down the categories as the combination of challenge and sense:

\noindent \textbf{Passive-Passive}: face anti-spoofing systems operate quietly without demanding any specific user actions or gestures, which is known as passive face anti-spoofing. They assess spoofing score solely based on the presented image or data. The representative approach is the single image-based face anti-spoofing models evaluated in the Sec.~\ref{sec:security_analsis_on_passive_face_liveness_detection}.

\noindent \textbf{Active-Passive}: face anti-spoofing systems typically engage the user by prompting them to perform a specific challenge, such as blinking or moving their head, which is known as active face anti-spoofing. Subsequently, the system verifies whether the user successfully completes this challenge. The active approaches take the advantage of using the motion information in the video data to spot out the artifacts. However, the underlying models are still passively verify the signal and challenge.

\begin{figure}[tbp]
  \centering
  \includegraphics[width=\linewidth]{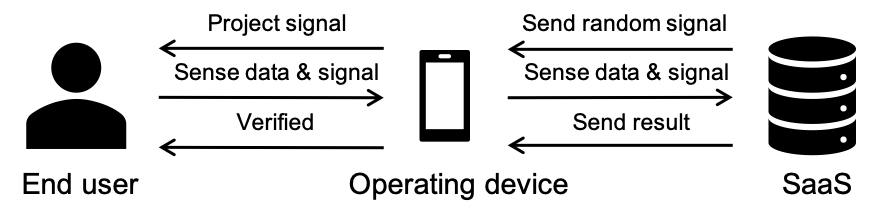}
  \caption{Illustration of one type of proactive approaches by projecting the random signals and verifying the signals.}
  \label{fig:proactive_liveness_detection_system}
\end{figure}

\noindent \textbf{Proactive}: face anti-spoofing systems employ sensors to actively emit signals into the physical environment to stay ahead of attack and enhance security. 
These signals can take various forms, such as infrared light or emissions specific to the sensor's technology. 
The fundamental criteria for these hardware sensors are their capability to project signals that are challenging to replicate. Such signals can either be randomly generated by a secure server or unique signals projected by a secure imaging sensor. 
These signals induce physical environmental changes, which are then captured by sensors. 
Subsequently, the face anti-spoofing system can validate these signals either through imaging on the sensor itself or by verifying the signal in a cloud-based model, as illustrated in Figure~\ref{fig:proactive_liveness_detection_system}.
This approach allows the system to closely collaborate with hardware sensors, significantly bolstering the protection of data integrity against digital injection attack and increasing the system's ability to detect presentation attack within the physical environment.
Within the proactive category, two subcategories exist:
\begin{enumerate}[leftmargin=*]
    \item Passive-Active Approaches: Users are only required to present their face without any specific action. This method relies on the hardware sensor to add more protection layers, such as multi-modality approaches using depth and IR images to challenge the ability of modification from the attackers simultaneously.
    \item Active-Active Approaches: These require users to execute a noticeable action as part of the challenge. This method achieves the maximum protection in our setting by verifying the information not only from the spatial and temporal dimensions, but also the signal variations projected from the sensors.
\end{enumerate}
Therefore, beside of analyzing the artifacts in the media, the proactive face anti-spoofing system can analyze these changes to validate the user's authenticity to make sure the genuineness of the time, location, and human, effectively preventing common digital injection attack and even digital replay attack.
To compare these three categories of approaches, we will summarize them based on six dimensions, as depicted in Figure~\ref{fig:comparison_liveness_systems}:

\begin{figure*}[htbp]
  \centering
  \subfloat[Passive face anti-spoofing]{\includegraphics[width=0.3\linewidth]{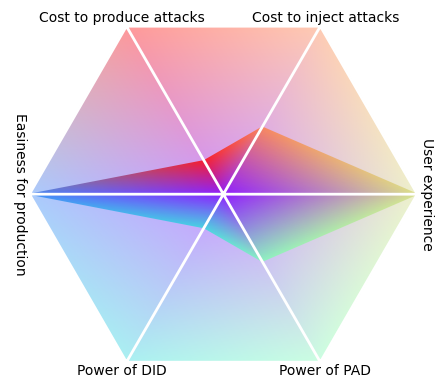}}
  \hfill
  \subfloat[Active face anti-spoofing]{\includegraphics[width=0.3\linewidth]{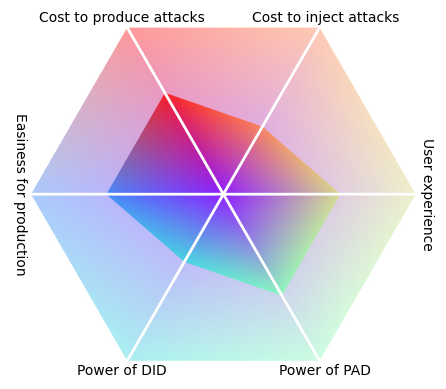}}
  \hfill
  \subfloat[Proactive face anti-spoofing]{\includegraphics[width=0.3\linewidth]{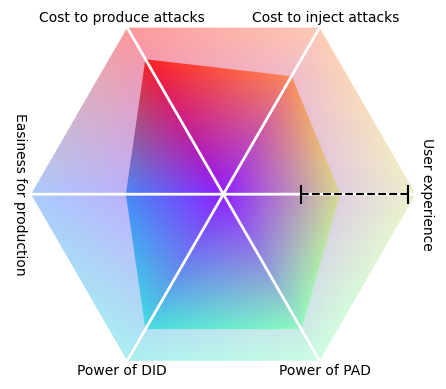}}
  \caption{The comparison of three different type of face anti-spoofing system. Passive face anti-spoofing has the advantage of best user experience and being most easy to deploy for production. Proactive approach has the best security that increase significant cost to attackers and have more power to detect the attacks. The user experience of proactive approach depends on type of the sensor, which could be as high as passive approach by deploying more advanced sensor or more carefully-designing patterns.}
  \label{fig:comparison_liveness_systems}
\end{figure*}

\noindent \textbf{Cost to Generate Attack}: This dimension focuses on the expenses incurred by attackers in creating fraudulent attempts. In a passive face anti-spoofing system, the cost for attackers is extremely low. They can utilize online photos of the target victim or employ existing deepfakes generation methods to fabricate a convincing likeness of the target or a new identity. Similarly, active approaches present a low barrier to attackers but require them to invest more resources in generating data with fewer discernible anomalies. However, proactive approaches involving active signals from sensors interacting with the physical environment are significantly more challenging to replicate, thus increasing the cost for attackers.

\noindent \textbf{Cost to Inject Attack}: This dimension considers the expense for attackers to inject fraudulent attempts into the anti-spoofing system. The cost varies depending on use cases and may require prior knowledge, such as hardware devices and operating systems. In passive or active face anti-spoofing, attackers can inject a single frame or a sequence of frames if they find a way to physically or digitally inject the data into the system. Proactive approaches, on the other hand, demand attackers to simulate the active signals projected from sensors for successful attack, resulting in higher attack costs.

\noindent \textbf{User Experience}: This dimension assesses the user experience during interactions with the face anti-spoofing system. Common metrics include the first-time user pass rate and the ease of performing actions measured by time. Passive face anti-spoofing offers the best user experience as it requires no interaction from the end user. In most cases, users may not even notice the face anti-spoofing check, and the entire process can take less than a second. Active face anti-spoofing systems require end users to perform specific gestures, which involve user understanding and the model's capability to recognize the gesture. This interaction typically takes several seconds and may reduce the pass rate. Proactive approaches can have varying user experiences depending on the hardware and signal type. With support for active invisible signals (e.g., Infrared light), the interaction can be minimized (passive-active), maximizing user experience. However, for maximum protection, the active-active approach necessitates projecting active signals and capturing user movements, resulting in decreased user experience and increased overall interaction time.

\noindent \textbf{Power of Presentation Attack Detection}: This dimension assesses the ability of these approaches to detect presentation attack originating from the physical world. In general, passive face anti-spoofing exhibits the lowest robustness compared to the other two approaches because it has fewer clues to make its final decision. Passive face anti-spoofing's performance varies based on factors such as the presence of artifacts in the image, training set coverage, model design, and generalization. 
Compared to the passive approach, the active approach finds it easier to detect presentation media without a high degree of deformation freedom, such as paper or a screen, in certain challenges (e.g., moving the head right/left). However, active systems may still struggle with 3D masks. The proactive approach, with the help of active sensing, can also reconstruct the surface of the presentation media and easily detect paper or screens. Certain signals (e.g., visible/infrared light) can be used to deter 3D masks due to their inherent physical attributes.

\noindent \textbf{Power of Digital Injection Detection}: This dimension evaluates the ability to detect digital injection attack from the digital world. Passive face face anti-spoofing is vulnerable to digital injection attack because it relies solely on the analysis of presented data or images, which can be spoofed or manipulated. Attackers may find it relatively inexpensive to produce spoofing content and carry out attack against passive systems. Consequently, deploying passive face anti-spoofing systems may be limited in untrusted or high-risk environments where the threat of attack is significant.
Active face anti-spoofing systems, on the other hand, are also facing increased vulnerability in the era of advanced generative techniques. Researchers and developers have created tools and models capable of generating highly realistic talking heads and deepfakes videos, which are effective against low-cost presentation attack and low-quality deepfakes attack. Attackers can employ various strategies to compromise active face anti-spoofing, including creating 3D masks or generating digital injection attack.
It's worth noting that even active face anti-spoofing systems struggle to prevent digital replay attack, as discussed earlier. These challenges highlight the evolving nature of security threats in the face of advancing AI and generative technologies, emphasizing the need for continuous research and development in face anti-spoofing and countermeasure techniques.

\noindent \textbf{Easiness for Production}: This dimension evaluates the ease of deploying and scaling these models for production. Passive face anti-spoofing models are simpler and require no cooperation from end-users, making them much easier to deploy and maintain compared to the other two approaches. Active and proactive approaches, on the other hand, require maintaining state (action type/signal sequence) within a secure server and often involve streaming infrastructure to reduce overall end-to-end processing time. Consequently, they require more engineering expertise to deploy the final solution into production.

Therefore, in situations involving medium and high-value transactions, achieving the highest level of protection necessitates a proactive approach that collaborates with active sensors and hardware to perform active defense. 
This approach can defend against a wide spectrum of threats, ranging from physical presentation attack to digital replay attack.

\subsection {Platform Robustness}

face anti-spoofing systems work in a distributed setting substrate with a sensor-equipped client device communicating and coordinating across the Internet via various network services with a server that executes detection algorithms and models. Naturally, face anti-spoofing can be compromised through attack on the platforms and services that form the distributed computing substrate.

\subsubsection {Client Device}
%TO DO
%Can't trust as the device is in uncontrolled deployment.
Operating system and application software in the client  device are part of the data pipeline in face anti-spoofing systems as they capture sensor data, process it to extract information, which subsequently drives further user interaction or is sent to the server for downstream processing by the models. The integrity of the client-side software stack is critical particularly because the malicious users can perform various attacks including digital injection attack and result modification/override.
%The raw sensor data produced by the client device hardware is captured, processed, and transmitted by a combination of operating system and application software

\subsubsection{Network Services}

face anti-spoofing systems depend on the Internet and various network services for communication and coordination between the server and the client. The wide availability of the Transport Layer Security (TLS) cryptographic protocol makes it possible to prevent eavesdropping and tampering of the content of server-client communication through the use of technologies such as HTTPS and WSS (secure web sockets). However, with the remote sensor device in face anti-spoofing systems often not being fully trustworthy (e.g., the client may be a jail-broken smartphone or a user-provided low-cost embedded IoT device with old buggy firmware), the integrity of server-client communication cannot always be assumed.

Moreover, face anti-spoofing systems depend on network services, which if compromised can allow an attacker to subvert the model inference. In particular, while it is computationally very hard for network attackers to break modern encryption, it is easy for a network attacker to inject packet delays which can be used to compromise time synchronization protocols such as NTP, thus making timestamps and measurements of elapsed time at the client untrustworthy. Additionally, the widely used venerable NTP protocol that Internet-connected computers and devices depend on for their sense of time has had many vulnerabilities reported in recent years, both in the protocol itself and in the various implementations. Timestamping errors of several seconds have also been reported on certain versions of Android mobile OS~\citep{sandha2020time}. A properly designed face anti-spoofing system must therefore not rely on the client device platform having a reliable knowledge of absolute or relative time, and must not blindly trust any timestamp and measurements of time reported by the client side.

\subsubsection{Server-side Processing}

The server-side processing in a face anti-spoofing system consist of executing and inferencing models. While the possibility of platform vulnerabilities always exists due to zero-days in software, various micro-architectural side-channel exploits (timing, power, row-hammer, etc.), and software supply chain attack, these risks are not specific to face anti-spoofing systems, and can be managed by adopting various prevention and mitigation strategies available at operating system, programming language, and access control level. 

% Multi-layer protection

% if the device is unprotected, time \& generation complexity.
% if protected, active sensors -> signals noticable/unnoticable.
% The protection methods can be enforced on both server side and client side.

% Cost to attack -> Cost to inject the attack
% Challenge-based face anti-spoofing - Cognitive-interaction -> Languiage -> Instruction In the limited time interaton bandwidth.
% Physical-interaction
% deepfakes is getting realistic -> passive -> realtieme -> hurt user experince -> physical challenge. 

% add the discussion on protection on the video injection.

% \subsection{\textcolor{red}{OLD TEXT}}

% \subsection{Server side}
% \subsubsection{Active Signals}
% Freshness
% Randomness
% \subsubsection{Accuracy}

% \subsubsection{Cryptographic Signature Key}

% \subsection{Client side}

% The client-side security cannot be fully trusted as there is no reliable way in practice to differentiate between injected images/videos and real-time content captured by a genuine user, considering that the hacker has physical or remote control over a device.

% \subsubsection{Mutual Authentication}

% \subsubsection{Secure Channel Protocol}

% % TS 102 484 : Secure channel between a UICC and an end-point terminal, Secure Channel Protocol ’03’

% \subsubsection{Hardware}

% Reasonable solution to protect against digital injection attack under:
% % bug free!
% \textit{e.g.} Trusted Execution Environments (TEEs) or Secure Enclave as a secure subsystem integrated into Apple systems on chip

% Attack?: \url{https://www.macrumors.com/2017/08/18/hacker-releases-decryption-key-secure-enclave/}

% Knox security platform
% https://www.samsungknox.com/en

% 1. Memory issue
% 2. Transition cost

\section{Conclusion}
In this paper, we have presented an extensive threat model for face anti-spoofing within the realm of cloud computing. Our objective has been to pinpoint the prevalent attack associated with machine learning models.
In the course of our experiments, we have employed passive single-image face face anti-spoofing models as representative examples. 
These models have been subjected to rigorous benchmarking across diverse attack scenarios, revealing their vulnerabilities to presentation attack, deepfakes manipulations, adversarial attack, and digital replay attack.
To counter these identified risks and bolster defenses against unforeseen threats, we have proffered a compendium of best practices and foundational principles essential for designing robust face face anti-spoofing systems.
Our recommendations span the entire spectrum, encompassing considerations ranging from sensor hardware intricacies to the intricacies of ML system design and the intricacies of secure server computing.
Especially, we proposed a proactive defending paradigm that harnesses the capabilities of sensors to actively defend the various threat and keep the high top security in the era of AIGC.

\bibliography{sn-bibliography}% common bib file

\begin{thebibliography}{}
\providecommand{\doi}[1]{\url{https://doi.org/#1}}
\bibcommenthead

\bibitem[\protect\citeauthoryear{Arashloo, Kittler, and Christmas}{Arashloo
  et~al.}{2017}]{arashloo2017anomaly}
Arashloo, S.R., J.~Kittler, and W.~Christmas. 2017.
\newblock An anomaly detection approach to face spoofing detection: A new
  formulation and evaluation protocol.
\newblock {\em IEEE Access\/}~5 .

\bibitem[\protect\citeauthoryear{Bagdasaryan, Veit, Hua, Estrin, and
  Shmatikov}{Bagdasaryan et~al.}{2020}]{bagdasaryan2020backdoor}
Bagdasaryan, E., A.~Veit, Y.~Hua, D.~Estrin, and V.~Shmatikov 2020.
\newblock How to backdoor federated learning.
\newblock In {\em International Conference on Artificial Intelligence and
  Statistics}, pp.\  2938--2948.

\bibitem[\protect\citeauthoryear{Bai, Luo, Zhao, Wen, and Wang}{Bai
  et~al.}{2021}]{bai2021recent}
Bai, T., J.~Luo, J.~Zhao, B.~Wen, and Q.~Wang. 2021.
\newblock Recent advances in adversarial training for adversarial robustness.
\newblock {\em arXiv preprint arXiv:2102.01356\/} .

\bibitem[\protect\citeauthoryear{Bao, Chen, Wen, Li, and Hua}{Bao
  et~al.}{2018}]{bao2018towards}
Bao, J., D.~Chen, F.~Wen, H.~Li, and G.~Hua 2018.
\newblock Towards open-set identity preserving face synthesis.
\newblock In {\em Conference on Computer Vision and Pattern Recognition}, pp.\
  6713--6722.

\bibitem[\protect\citeauthoryear{Bitouk, Kumar, Dhillon, Belhumeur, and
  Nayar}{Bitouk et~al.}{2008}]{bitouk2008face}
Bitouk, D., N.~Kumar, S.~Dhillon, P.~Belhumeur, and S.K. Nayar. 2008.
\newblock Face swapping: automatically replacing faces in photographs, {\em ACM
  SIGGRAPH},  1--8.

\bibitem[\protect\citeauthoryear{Bommasani, Hudson, Adeli, Altman, Arora, von
  Arx, Bernstein, Bohg, Bosselut, Brunskill, et~al.}{Bommasani
  et~al.}{2021}]{bommasani2021opportunities}
Bommasani, R., D.A. Hudson, E.~Adeli, R.~Altman, S.~Arora, S.~von Arx, M.S.
  Bernstein, J.~Bohg, A.~Bosselut, E.~Brunskill, et~al. 2021.
\newblock On the opportunities and risks of foundation models.
\newblock {\em arXiv preprint arXiv:2108.07258\/} .

\bibitem[\protect\citeauthoryear{Boulkenafet, Komulainen, Li, Feng, and
  Hadid}{Boulkenafet et~al.}{2017}]{OULU_NPU_2017}
Boulkenafet, Z., J.~Komulainen, L.~Li, X.~Feng, and A.~Hadid 2017.
\newblock {OULU-NPU}: A mobile face presentation attack database with
  real-world variations.
\newblock In {\em 12th IEEE International Conference on Automatic Face Gesture
  Recognition (FG 2017)}, pp.\  612--618.

\bibitem[\protect\citeauthoryear{Bowyer and Burge}{Bowyer and
  Burge}{2016}]{bowyer2016handbook}
Bowyer, K.W. and M.J. Burge. 2016.
\newblock {\em Handbook of iris recognition}.
\newblock Springer.

\bibitem[\protect\citeauthoryear{Brown, Man{\'e}, Roy, Abadi, and Gilmer}{Brown
  et~al.}{2017}]{brown2017adversarial}
Brown, T.B., D.~Man{\'e}, A.~Roy, M.~Abadi, and J.~Gilmer. 2017.
\newblock Adversarial patch.
\newblock {\em arXiv preprint arXiv:1712.09665\/} .

\bibitem[\protect\citeauthoryear{Carta, Barral, El~Mrabet, and Mouille}{Carta
  et~al.}{2021}]{carta2021pitfalls}
Carta, K., C.~Barral, N.~El~Mrabet, and S.~Mouille 2021.
\newblock On the pitfalls of videoconferences for challenge-based face liveness
  detection.
\newblock In {\em World Multi-Conference on Systemics, Cybernetics and
  Informatics}, pp.\  1--6.

\bibitem[\protect\citeauthoryear{Carta, Barral, Mrabet, and Mouille}{Carta
  et~al.}{2022}]{carta2022video}
Carta, K., C.~Barral, N.E. Mrabet, and S.~Mouille 2022.
\newblock Video injection attacks on remote digital identity verification
  solution using face recognition.
\newblock In {\em International Multi-Conference on Complexity, Informatics and
  Cybernetics}.

\bibitem[\protect\citeauthoryear{Chan, Liu, Chen, Yeung, Zhang, Wang, and
  Hsu}{Chan et~al.}{2017}]{chan2017face}
Chan, P.P., W.~Liu, D.~Chen, D.S. Yeung, F.~Zhang, X.~Wang, and C.C. Hsu. 2017.
\newblock Face liveness detection using a flash against 2d spoofing attack.
\newblock {\em IEEE Transactions on Information Forensics and Security\/}~{\em
  13\/}(2): 521--534 .

\bibitem[\protect\citeauthoryear{Chang, Lee, Yao, Chen, Wang, Lai, and
  Chen}{Chang et~al.}{2023}]{chang2023closer}
Chang, C.J., Y.C. Lee, S.H. Yao, M.H. Chen, C.Y. Wang, S.H. Lai, and T.P.C.
  Chen 2023.
\newblock A closer look at geometric temporal dynamics for face anti-spoofing.
\newblock In {\em Proceedings of the IEEE/CVF Conference on Computer Vision and
  Pattern Recognition}, pp.\  1081--1091.

\bibitem[\protect\citeauthoryear{Chen, Zhang, Song, Liu, and Wang}{Chen
  et~al.}{2022}]{chen2022self}
Chen, L., Y.~Zhang, Y.~Song, L.~Liu, and J.~Wang 2022.
\newblock Self-supervised learning of adversarial example: Towards good
  generalizations for deepfake detection.
\newblock In {\em Proceedings of the IEEE/CVF conference on computer vision and
  pattern recognition}, pp.\  18710--18719.

\bibitem[\protect\citeauthoryear{Chen, Zhang, Song, Wang, and Liu}{Chen
  et~al.}{2022}]{chen2022ost}
Chen, L., Y.~Zhang, Y.~Song, J.~Wang, and L.~Liu. 2022.
\newblock Ost: Improving generalization of deepfake detection via one-shot
  test-time training.
\newblock {\em Advances in Neural Information Processing Systems\/}~35:
  24597--24610 .

\bibitem[\protect\citeauthoryear{Chen, Chen, Ni, and Ge}{Chen
  et~al.}{2020}]{chen2020simswap}
Chen, R., X.~Chen, B.~Ni, and Y.~Ge 2020.
\newblock Simswap: An efficient framework for high fidelity face swapping.
\newblock In {\em Proceedings of the 28th ACM International Conference on
  Multimedia}, pp.\  2003--2011.

\bibitem[\protect\citeauthoryear{Chen, Liu, Li, Lu, and Song}{Chen
  et~al.}{2017}]{chen2017targeted}
Chen, X., C.~Liu, B.~Li, K.~Lu, and D.~Song. 2017.
\newblock Targeted backdoor attacks on deep learning systems using data
  poisoning.
\newblock {\em arXiv preprint arXiv:1712.05526\/} .

\bibitem[\protect\citeauthoryear{Chen, Ma, and Ma}{Chen
  et~al.}{2019}]{chen2019biometric}
Chen, Y., B.~Ma, and Z.~Ma. 2019.
\newblock Biometric authentication under threat: Liveness detection hacking.
\newblock {\em Black Hat USA\/} .

\bibitem[\protect\citeauthoryear{Cherepanova, Goldblum, Foley, Duan, Dickerson,
  Taylor, and Goldstein}{Cherepanova et~al.}{2021}]{cherepanova2021lowkey}
Cherepanova, V., M.~Goldblum, H.~Foley, S.~Duan, J.~Dickerson, G.~Taylor, and
  T.~Goldstein 2021.
\newblock Lowkey: Leveraging adversarial attacks to protect social media users
  from facial recognition.
\newblock In {\em International Conference on Learning Representations}.

\bibitem[\protect\citeauthoryear{Chingovska, Anjos, and Marcel}{Chingovska
  et~al.}{2012}]{chingovska2012effectiveness}
Chingovska, I., A.~Anjos, and S.~Marcel 2012.
\newblock On the effectiveness of local binary patterns in face anti-spoofing.
\newblock In {\em 2012 BIOSIG-proceedings of the international conference of
  biometrics special interest group (BIOSIG)}, pp.\  1--7. IEEE.

\bibitem[\protect\citeauthoryear{Chuang, Wang, and Lai}{Chuang
  et~al.}{2023}]{chuang2023generalized}
Chuang, C.C., C.Y. Wang, and S.H. Lai 2023.
\newblock Generalized face anti-spoofing via multi-task learning and one-side
  meta triplet loss.
\newblock In {\em 2023 IEEE 17th International Conference on Automatic Face and
  Gesture Recognition (FG)}, pp.\  1--8. IEEE.

\bibitem[\protect\citeauthoryear{Ciftci, Demir, and Yin}{Ciftci
  et~al.}{2020}]{ciftci2020fakecatcher}
Ciftci, U.A., I.~Demir, and L.~Yin. 2020.
\newblock Fakecatcher: Detection of synthetic portrait videos using biological
  signals.
\newblock {\em IEEE Transactions on Pattern Analysis and Machine
  Intelligence\/} .

\bibitem[\protect\citeauthoryear{Cox, Miller, Bloom, Fridrich, and Kalker}{Cox
  et~al.}{2007}]{cox2007digital}
Cox, I., M.~Miller, J.~Bloom, J.~Fridrich, and T.~Kalker. 2007.
\newblock {\em Digital watermarking and steganography}.
\newblock Morgan Kaufmann Publishers.

\bibitem[\protect\citeauthoryear{Deb, Liu, and Jain}{Deb
  et~al.}{2023a}]{deb2023faceguard}
Deb, D., X.~Liu, and A.K. Jain 2023a.
\newblock Faceguard: A self-supervised defense against adversarial face images.
\newblock In {\em International Conference on Automatic Face and Gesture
  Recognition}, pp.\  1--8.

\bibitem[\protect\citeauthoryear{Deb, Liu, and Jain}{Deb
  et~al.}{2023b}]{deb2023unified}
Deb, D., X.~Liu, and A.K. Jain 2023b.
\newblock Unified detection of digital and physical face attacks.
\newblock In {\em International Conference on Automatic Face and Gesture
  Recognition}, pp.\  1--8.

\bibitem[\protect\citeauthoryear{Deng, Dong, Socher, Li, Li, and Fei-Fei}{Deng
  et~al.}{2009}]{deng2009imagenet}
Deng, J., W.~Dong, R.~Socher, L.J. Li, K.~Li, and L.~Fei-Fei 2009.
\newblock Imagenet: A large-scale hierarchical image database.
\newblock In {\em 2009 IEEE conference on computer vision and pattern
  recognition}, pp.\  248--255. Ieee.

\bibitem[\protect\citeauthoryear{Deng, Chen, Meng, Zhang, Xu, and Cheng}{Deng
  et~al.}{2022}]{deng2022understanding}
Deng, Z., K.~Chen, G.~Meng, X.~Zhang, K.~Xu, and Y.~Cheng 2022.
\newblock Understanding real-world threats to deep learning models in android
  apps.
\newblock In {\em ACM Conference on Computer and Communications Security}, pp.\
   785--799.

\bibitem[\protect\citeauthoryear{Diogenes and Ozkaya}{Diogenes and
  Ozkaya}{2018}]{diogenes2018cybersecurity}
Diogenes, Y. and E.~Ozkaya. 2018.
\newblock {\em Cybersecurity-attack and defense strategies: Infrastructure
  security with red team and blue team tactics}.
\newblock Packt Publishing Ltd.

\bibitem[\protect\citeauthoryear{Dolhansky, Bitton, Pflaum, Lu, Howes, Wang,
  and Ferrer}{Dolhansky et~al.}{2020}]{dolhansky2020deepfake}
Dolhansky, B., J.~Bitton, B.~Pflaum, J.~Lu, R.~Howes, M.~Wang, and C.C. Ferrer.
  2020.
\newblock The deepfake detection challenge (dfdc) dataset.
\newblock {\em arXiv preprint arXiv:2006.07397\/} .

\bibitem[\protect\citeauthoryear{Dolhansky, Howes, Pflaum, Baram, and
  Ferrer}{Dolhansky et~al.}{2019}]{dolhansky2019deepfake}
Dolhansky, B., R.~Howes, B.~Pflaum, N.~Baram, and C.C. Ferrer. 2019.
\newblock The deepfake detection challenge (dfdc) preview dataset.
\newblock {\em arXiv preprint arXiv:1910.08854\/} .

\bibitem[\protect\citeauthoryear{Dong, Wang, Ji, Liang, Fan, and Ge}{Dong
  et~al.}{2023}]{dong2023implicit}
Dong, S., J.~Wang, R.~Ji, J.~Liang, H.~Fan, and Z.~Ge 2023.
\newblock Implicit identity leakage: The stumbling block to improving deepfake
  detection generalization.
\newblock In {\em Conference on Computer Vision and Pattern Recognition}, pp.\
  3994--4004.

\bibitem[\protect\citeauthoryear{Dong, Bao, Chen, Zhang, Yu, Chen, Wen, and
  Guo}{Dong et~al.}{2020}]{dong2020identity}
Dong, X., J.~Bao, D.~Chen, W.~Zhang, N.~Yu, D.~Chen, F.~Wen, and B.~Guo. 2020.
\newblock Identity-driven deepfake detection.
\newblock {\em arXiv preprint arXiv:2012.03930\/} .

\bibitem[\protect\citeauthoryear{Dong, Su, Wu, Li, Liu, Zhang, and Zhu}{Dong
  et~al.}{2019}]{dong2019efficient}
Dong, Y., H.~Su, B.~Wu, Z.~Li, W.~Liu, T.~Zhang, and J.~Zhu 2019.
\newblock Efficient decision-based black-box adversarial attacks on face
  recognition.
\newblock In {\em Conference on Computer Vision and Pattern Recognition}, pp.\
  7714--7722.

\bibitem[\protect\citeauthoryear{Du, Li, Zuo, Zhu, and Lu}{Du
  et~al.}{2022}]{du2022energy}
Du, Z., J.~Li, L.~Zuo, L.~Zhu, and K.~Lu 2022.
\newblock Energy-based domain generalization for face anti-spoofing.
\newblock In {\em Proceedings of the 30th ACM International Conference on
  Multimedia}, pp.\  1749--1757.

\bibitem[\protect\citeauthoryear{Ebihara, Sakurai, and Imaoka}{Ebihara
  et~al.}{2020}]{ebihara2020specular}
Ebihara, A.F., K.~Sakurai, and H.~Imaoka 2020.
\newblock Specular-and diffuse-reflection-based face spoofing detection for
  mobile devices.
\newblock In {\em 2020 IEEE International Joint Conference on Biometrics
  (IJCB)}, pp.\  1--10. IEEE.

\bibitem[\protect\citeauthoryear{Erdogmus and Marcel}{Erdogmus and
  Marcel}{2014}]{erdogmus2014spoofing}
Erdogmus, N. and S.~Marcel. 2014.
\newblock Spoofing face recognition with 3d masks.
\newblock {\em IEEE Transactions on Information Forensics and Security\/}~{\em
  9\/}(7): 1084--1097 .

\bibitem[\protect\citeauthoryear{Fang, Liu, Wan, Escalera, Zhao, Zhang, Li, and
  Lei}{Fang et~al.}{2023}]{fang2023surveillance}
Fang, H., A.~Liu, J.~Wan, S.~Escalera, C.~Zhao, X.~Zhang, S.Z. Li, and Z.~Lei.
  2023.
\newblock Surveillance face anti-spoofing.
\newblock {\em IEEE Transactions on Information Forensics and Security\/} .

\bibitem[\protect\citeauthoryear{Fang, Liu, Yuan, Zheng, Zeng, Liu, Deng,
  Escalera, Liu, Wan, et~al.}{Fang et~al.}{2024}]{fang2024unified}
Fang, H., A.~Liu, H.~Yuan, J.~Zheng, D.~Zeng, Y.~Liu, J.~Deng, S.~Escalera,
  X.~Liu, J.~Wan, et~al. 2024.
\newblock Unified physical-digital face attack detection.
\newblock {\em arXiv preprint arXiv:2401.17699\/} .

\bibitem[\protect\citeauthoryear{Fang and Damer}{Fang and
  Damer}{2024}]{fang2024face}
Fang, M. and N.~Damer 2024.
\newblock Face presentation attack detection by excavating causal clues and
  adapting embedding statistics.
\newblock In {\em Proceedings of the IEEE/CVF Winter Conference on Applications
  of Computer Vision}, pp.\  6269--6279.

\bibitem[\protect\citeauthoryear{Fang, Damer, Kirchbuchner, and Kuijper}{Fang
  et~al.}{2022}]{fang2022learnable}
Fang, M., N.~Damer, F.~Kirchbuchner, and A.~Kuijper 2022.
\newblock Learnable multi-level frequency decomposition and hierarchical
  attention mechanism for generalized face presentation attack detection.
\newblock In {\em Proceedings of the IEEE/CVF winter conference on applications
  of computer vision}, pp.\  3722--3731.

\bibitem[\protect\citeauthoryear{Farrukh, Aburas, Cao, and Wang}{Farrukh
  et~al.}{2020}]{farrukh2020facerevelio}
Farrukh, H., R.M. Aburas, S.~Cao, and H.~Wang 2020.
\newblock Facerevelio: a face liveness detection system for smartphones with a
  single front camera.
\newblock In {\em International Conference on Mobile Computing and Networking},
  pp.\  1--13.

\bibitem[\protect\citeauthoryear{Feng, Chen, and Owens}{Feng
  et~al.}{2023}]{feng2023self}
Feng, C., Z.~Chen, and A.~Owens 2023.
\newblock Self-supervised video forensics by audio-visual anomaly detection.
\newblock In {\em Proceedings of the IEEE/CVF Conference on Computer Vision and
  Pattern Recognition}, pp.\  10491--10503.

\bibitem[\protect\citeauthoryear{Feng, Wu, Shao, Wang, and Zhou}{Feng
  et~al.}{2018}]{feng2018joint}
Feng, Y., F.~Wu, X.~Shao, Y.~Wang, and X.~Zhou 2018.
\newblock Joint 3d face reconstruction and dense alignment with position map
  regression network.
\newblock In {\em Proceedings of the European conference on computer vision
  (ECCV)}, pp.\  534--551.

\bibitem[\protect\citeauthoryear{Francillon, Danev, and Capkun}{Francillon
  et~al.}{2011}]{francillon2011relay}
Francillon, A., B.~Danev, and S.~Capkun 2011.
\newblock Relay attacks on passive keyless entry and start systems in modern
  cars.
\newblock In {\em Network and Distributed System Security Symposium}.

\bibitem[\protect\citeauthoryear{Fu and Xu}{Fu and Xu}{2018}]{fu2018risks}
Fu, K. and W.~Xu. 2018.
\newblock Risks of trusting the physics of sensors.
\newblock {\em Communications of the ACM\/}~{\em 61\/}(2): 20--23 .

\bibitem[\protect\citeauthoryear{Gao, Doan, Zhang, Ma, Zhang, Fu, Nepal, and
  Kim}{Gao et~al.}{2020}]{gao2020backdoor}
Gao, Y., B.G. Doan, Z.~Zhang, S.~Ma, J.~Zhang, A.~Fu, S.~Nepal, and H.~Kim.
  2020.
\newblock Backdoor attacks and countermeasures on deep learning: A
  comprehensive review.
\newblock {\em arXiv preprint arXiv:2007.10760\/} .

\bibitem[\protect\citeauthoryear{Goodfellow, Pouget-Abadie, Mirza, Xu,
  Warde-Farley, Ozair, Courville, and Bengio}{Goodfellow
  et~al.}{2014}]{goodfellow2014generative}
Goodfellow, I., J.~Pouget-Abadie, M.~Mirza, B.~Xu, D.~Warde-Farley, S.~Ozair,
  A.~Courville, and Y.~Bengio. 2014.
\newblock Generative adversarial nets.
\newblock {\em Advances in neural information processing systems\/}~27 .

\bibitem[\protect\citeauthoryear{Goodfellow, Shlens, and Szegedy}{Goodfellow
  et~al.}{2014}]{goodfellow2014explaining}
Goodfellow, I.J., J.~Shlens, and C.~Szegedy. 2014.
\newblock Explaining and harnessing adversarial examples.
\newblock {\em arXiv preprint arXiv:1412.6572\/} .

\bibitem[\protect\citeauthoryear{Gu, Chen, Yao, Ding, Li, Huang, and Ma}{Gu
  et~al.}{2021}]{gu2021spatiotemporal}
Gu, Z., Y.~Chen, T.~Yao, S.~Ding, J.~Li, F.~Huang, and L.~Ma 2021.
\newblock Spatiotemporal inconsistency learning for deepfake video detection.
\newblock In {\em Proceedings of the 29th ACM international conference on
  multimedia}, pp.\  3473--3481.

\bibitem[\protect\citeauthoryear{Guo, Liu, Jain, and Liu}{Guo
  et~al.}{2022}]{guo2022multi}
Guo, X., Y.~Liu, A.~Jain, and X.~Liu 2022.
\newblock Multi-domain learning for updating face anti-spoofing models.
\newblock In {\em European Conference on Computer Vision}, pp.\  230--249.

\bibitem[\protect\citeauthoryear{Guzman, Goryawala, Wang, Barreto, Andrian,
  Rishe, and Adjouadi}{Guzman et~al.}{2012}]{guzman2012thermal}
Guzman, A.M., M.~Goryawala, J.~Wang, A.~Barreto, J.~Andrian, N.~Rishe, and
  M.~Adjouadi. 2012.
\newblock Thermal imaging as a biometrics approach to facial signature
  authentication.
\newblock {\em IEEE Journal of Biomedical and Health Informatics\/}~{\em
  17\/}(1): 214--222 .

\bibitem[\protect\citeauthoryear{Haliassos, Vougioukas, Petridis, and
  Pantic}{Haliassos et~al.}{2021}]{haliassos2021lips}
Haliassos, A., K.~Vougioukas, S.~Petridis, and M.~Pantic 2021.
\newblock Lips don't lie: A generalisable and robust approach to face forgery
  detection.
\newblock In {\em Conference on Computer Vision and Pattern Recognition}, pp.\
  5039--5049.

\bibitem[\protect\citeauthoryear{Han, Wang, Chen, Chen, Guo, Liu, Tang, Xiao,
  Xu, Xu, et~al.}{Han et~al.}{2022}]{han2022survey}
Han, K., Y.~Wang, H.~Chen, X.~Chen, J.~Guo, Z.~Liu, Y.~Tang, A.~Xiao, C.~Xu,
  Y.~Xu, et~al. 2022.
\newblock A survey on vision transformer.
\newblock {\em IEEE transactions on pattern analysis and machine
  intelligence\/}~{\em 45\/}(1): 87--110 .

\bibitem[\protect\citeauthoryear{Han, Chan, Wengrowski, Li, Tippenhauer,
  Srivastava, Zonouz, and Garcia}{Han et~al.}{2023}]{han2023dont}
Han, Y., M.~Chan, E.~Wengrowski, Z.~Li, N.O. Tippenhauer, M.~Srivastava,
  S.~Zonouz, and L.~Garcia. 2023.
\newblock Why don't you clean your glasses? perception attacks with dynamic
  optical perturbations.
\newblock {\em arXiv preprint arXiv:2307.13131\/} .

\bibitem[\protect\citeauthoryear{He, Zhang, Ren, and Sun}{He
  et~al.}{2016}]{he2016deep}
He, K., X.~Zhang, S.~Ren, and J.~Sun 2016.
\newblock Deep residual learning for image recognition.
\newblock In {\em Proceedings of the IEEE conference on computer vision and
  pattern recognition}, pp.\  770--778.

\bibitem[\protect\citeauthoryear{He, Gan, Chen, Zhou, Yin, Song, Sheng, Shao,
  and Liu}{He et~al.}{2021}]{he2021forgerynet}
He, Y., B.~Gan, S.~Chen, Y.~Zhou, G.~Yin, L.~Song, L.~Sheng, J.~Shao, and
  Z.~Liu 2021.
\newblock Forgerynet: A versatile benchmark for comprehensive forgery analysis.
\newblock In {\em Conference on Computer Vision and Pattern Recognition}, pp.\
  4360--4369.

\bibitem[\protect\citeauthoryear{Hernandez-Ortega, Fierrez, Morales, and
  Galbally}{Hernandez-Ortega et~al.}{2019}]{hernandez2019introduction}
Hernandez-Ortega, J., J.~Fierrez, A.~Morales, and J.~Galbally. 2019.
\newblock Introduction to face presentation attack detection.
\newblock {\em Handbook of Biometric Anti-Spoofing\/}: 187--206 .

\bibitem[\protect\citeauthoryear{Ho, Jain, and Abbeel}{Ho
  et~al.}{2020}]{ho2020denoising}
Ho, J., A.~Jain, and P.~Abbeel. 2020.
\newblock Denoising diffusion probabilistic models.
\newblock {\em Advances in Neural Information Processing Systems\/}~33:
  6840--6851 .

\bibitem[\protect\citeauthoryear{Ho and Salimans}{Ho and
  Salimans}{2022}]{ho2022classifier}
Ho, J. and T.~Salimans. 2022.
\newblock Classifier-free diffusion guidance.
\newblock {\em arXiv preprint arXiv:2207.12598\/} .

\bibitem[\protect\citeauthoryear{Horvitz}{Horvitz}{2022}]{horvitz2022horizon}
Horvitz, E. 2022.
\newblock On the horizon: Interactive and compositional deepfakes.
\newblock In {\em International Conference on Multimodal Interaction}, pp.\
  653--661.

\bibitem[\protect\citeauthoryear{Hu, Liu, Zhang, Li, Zhang, Jin, and Wu}{Hu
  et~al.}{2022}]{hu2022protecting}
Hu, S., X.~Liu, Y.~Zhang, M.~Li, L.Y. Zhang, H.~Jin, and L.~Wu 2022.
\newblock Protecting facial privacy: Generating adversarial identity masks via
  style-robust makeup transfer.
\newblock In {\em Conference on Computer Vision and Pattern Recognition}, pp.\
  15014--15023.

\bibitem[\protect\citeauthoryear{Huang, Wang, Yang, Ai, Zou, Wang, and
  Ye}{Huang et~al.}{2023}]{huang2023implicit}
Huang, B., Z.~Wang, J.~Yang, J.~Ai, Q.~Zou, Q.~Wang, and D.~Ye 2023.
\newblock Implicit identity driven deepfake face swapping detection.
\newblock In {\em Conference on Computer Vision and Pattern Recognition}, pp.\
  4490--4499.

\bibitem[\protect\citeauthoryear{Huang, Sun, Liu, Chu, Xiao, Yuan, Adam, and
  Yang}{Huang et~al.}{2022}]{huang2022adaptive}
Huang, H.P., D.~Sun, Y.~Liu, W.S. Chu, T.~Xiao, J.~Yuan, H.~Adam, and M.H. Yang
  2022.
\newblock Adaptive transformers for robust few-shot cross-domain face
  anti-spoofing.
\newblock In {\em European Conference on Computer Vision}, pp.\  37--54.

\bibitem[\protect\citeauthoryear{Huang, Chong, Ni, Chen, and Hsu}{Huang
  et~al.}{2023}]{huang2023towards}
Huang, P.K., J.X. Chong, H.Y. Ni, T.H. Chen, and C.T. Hsu 2023.
\newblock Towards diverse liveness feature representation and domain expansion
  for cross-domain face anti-spoofing.
\newblock In {\em 2023 IEEE International Conference on Multimedia and Expo
  (ICME)}, pp.\  1199--1204. IEEE.

\bibitem[\protect\citeauthoryear{Huang, Ni, Ni, and Hsu}{Huang
  et~al.}{2022}]{huang2022learnable}
Huang, P.K., H.Y. Ni, Y.~Ni, and C.T. Hsu 2022.
\newblock Learnable descriptive convolutional network for face anti-spoofing.
\newblock In {\em BMVC}, pp.\  239.

\bibitem[\protect\citeauthoryear{Huh, Liu, Owens, and Efros}{Huh
  et~al.}{2018}]{huh2018fighting}
Huh, M., A.~Liu, A.~Owens, and A.A. Efros 2018.
\newblock Fighting fake news: Image splice detection via learned
  self-consistency.
\newblock In {\em European Conference on Computer Vision}, pp.\  101--117.

\bibitem[\protect\citeauthoryear{Jauernig, Sadeghi, and Stapf}{Jauernig
  et~al.}{2020}]{jauernig2020trusted}
Jauernig, P., A.R. Sadeghi, and E.~Stapf. 2020.
\newblock Trusted execution environments: properties, applications, and
  challenges.
\newblock {\em IEEE Security \& Privacy\/}~{\em 18\/}(2): 56--60 .

\bibitem[\protect\citeauthoryear{Jia, Zhang, Shan, and Chen}{Jia
  et~al.}{2020}]{jia2020single}
Jia, Y., J.~Zhang, S.~Shan, and X.~Chen 2020.
\newblock Single-side domain generalization for face anti-spoofing.
\newblock In {\em Conference on Computer Vision and Pattern Recognition}, pp.\
  8484--8493.

\bibitem[\protect\citeauthoryear{Jiang, Li, Wu, Qian, and Loy}{Jiang
  et~al.}{2020}]{jiang2020deeperforensics}
Jiang, L., R.~Li, W.~Wu, C.~Qian, and C.C. Loy 2020.
\newblock Deeperforensics-1.0: A large-scale dataset for real-world face
  forgery detection.
\newblock In {\em Proceedings of the IEEE/CVF conference on computer vision and
  pattern recognition}, pp.\  2889--2898.

\bibitem[\protect\citeauthoryear{Jie, Choo, Li, Chen, and Guo}{Jie
  et~al.}{2019}]{jie2019tradeoff}
Jie, Y., K.K.R. Choo, M.~Li, L.~Chen, and C.~Guo. 2019.
\newblock Tradeoff gain and loss optimization against man-in-the-middle attacks
  based on game theoretic model.
\newblock {\em Future Generation Computer Systems\/}~101: 169--179 .

\bibitem[\protect\citeauthoryear{Karras, Aittala, Hellsten, Laine, Lehtinen,
  and Aila}{Karras et~al.}{2020}]{karras2020training}
Karras, T., M.~Aittala, J.~Hellsten, S.~Laine, J.~Lehtinen, and T.~Aila. 2020.
\newblock Training generative adversarial networks with limited data.
\newblock {\em Advances in Neural Information Processing Systems\/}~33:
  12104--12114 .

\bibitem[\protect\citeauthoryear{Karras, Laine, and Aila}{Karras
  et~al.}{2019}]{karras2019style}
Karras, T., S.~Laine, and T.~Aila 2019.
\newblock A style-based generator architecture for generative adversarial
  networks.
\newblock In {\em Conference on Computer Vision and Pattern Recognition}, pp.\
  4401--4410.

\bibitem[\protect\citeauthoryear{Khalid, Tariq, Kim, and Woo}{Khalid
  et~al.}{2021}]{khalid2021fakeavceleb}
Khalid, H., S.~Tariq, M.~Kim, and S.S. Woo. 2021.
\newblock Fakeavceleb: A novel audio-video multimodal deepfake dataset.
\newblock {\em arXiv preprint arXiv:2108.05080\/} .

\bibitem[\protect\citeauthoryear{Korshunov and Marcel}{Korshunov and
  Marcel}{2018}]{korshunov2018deepfakes}
Korshunov, P. and S.~Marcel. 2018.
\newblock Deepfakes: a new threat to face recognition? assessment and
  detection.
\newblock {\em arXiv preprint arXiv:1812.08685\/} .

\bibitem[\protect\citeauthoryear{Korshunova, Shi, Dambre, and Theis}{Korshunova
  et~al.}{2017}]{korshunova2017fast}
Korshunova, I., W.~Shi, J.~Dambre, and L.~Theis 2017.
\newblock Fast face-swap using convolutional neural networks.
\newblock In {\em International Conference on Computer Vision}, pp.\
  3677--3685.

\bibitem[\protect\citeauthoryear{Kwak, Jung, Yoo, Shin, and Kim}{Kwak
  et~al.}{2023}]{kwak2023liveness}
Kwak, Y., M.~Jung, H.~Yoo, J.~Shin, and C.~Kim 2023.
\newblock Liveness score-based regression neural networks for face
  anti-spoofing.
\newblock In {\em ICASSP 2023-2023 IEEE International Conference on Acoustics,
  Speech and Signal Processing (ICASSP)}, pp.\  1--5. IEEE.

\bibitem[\protect\citeauthoryear{Le, Kim, Tariq, Moore, Abuadbba, and Woo}{Le
  et~al.}{2024}]{le2024sok}
Le, B.M., J.~Kim, S.~Tariq, K.~Moore, A.~Abuadbba, and S.S. Woo. 2024.
\newblock Sok: Facial deepfake detectors.
\newblock {\em arXiv preprint arXiv:2401.04364\/} .

\bibitem[\protect\citeauthoryear{Le, Nguyen, Yamagishi, and Echizen}{Le
  et~al.}{2021}]{le2021openforensics}
Le, T.N., H.H. Nguyen, J.~Yamagishi, and I.~Echizen 2021.
\newblock Openforensics: Large-scale challenging dataset for multi-face forgery
  detection and segmentation in-the-wild.
\newblock In {\em International Conference on Computer Vision}, pp.\
  10117--10127.

\bibitem[\protect\citeauthoryear{Li, Gan, Yang, Yang, Li, Wang, and Gao}{Li
  et~al.}{2023}]{li2023multimodal}
Li, C., Z.~Gan, Z.~Yang, J.~Yang, L.~Li, L.~Wang, and J.~Gao. 2023.
\newblock Multimodal foundation models: From specialists to general-purpose
  assistants.
\newblock {\em arXiv preprint arXiv:2309.10020\/}~{\em 1\/}(2): 2 .

\bibitem[\protect\citeauthoryear{Li, Wang, Ji, Zhang, Xi, Guo, and Wang}{Li
  et~al.}{2022}]{li2022seeing}
Li, C., L.~Wang, S.~Ji, X.~Zhang, Z.~Xi, S.~Guo, and T.~Wang 2022.
\newblock Seeing is living? rethinking the security of facial liveness
  verification in the deepfake era.
\newblock In {\em USENIX Security Symposium}, pp.\  2673--2690.

\bibitem[\protect\citeauthoryear{Li, Wang, Fan, and Dong}{Li
  et~al.}{2021}]{li2021exploring}
Li, D., W.~Wang, H.~Fan, and J.~Dong 2021.
\newblock Exploring adversarial fake images on face manifold.
\newblock In {\em Conference on Computer Vision and Pattern Recognition}, pp.\
  5789--5798.

\bibitem[\protect\citeauthoryear{Li, Li, Cao, Song, and He}{Li
  et~al.}{2021}]{li2021faceinpainter}
Li, J., Z.~Li, J.~Cao, X.~Song, and R.~He 2021.
\newblock Faceinpainter: High fidelity face adaptation to heterogeneous
  domains.
\newblock In {\em Conference on Computer Vision and Pattern Recognition}, pp.\
  5089--5098.

\bibitem[\protect\citeauthoryear{Li, Xie, Li, Wang, and Zhang}{Li
  et~al.}{2021}]{li2021frequency}
Li, J., H.~Xie, J.~Li, Z.~Wang, and Y.~Zhang 2021.
\newblock Frequency-aware discriminative feature learning supervised by
  single-center loss for face forgery detection.
\newblock In {\em Conference on Computer Vision and Pattern Recognition}, pp.\
  6458--6467.

\bibitem[\protect\citeauthoryear{Li, Bao, Yang, Chen, and Wen}{Li
  et~al.}{2019}]{li2019faceshifter}
Li, L., J.~Bao, H.~Yang, D.~Chen, and F.~Wen. 2019.
\newblock Faceshifter: Towards high fidelity and occlusion aware face swapping.
\newblock {\em arXiv preprint arXiv:1912.13457\/} .

\bibitem[\protect\citeauthoryear{Li, Bao, Zhang, Yang, Chen, Wen, and Guo}{Li
  et~al.}{2020}]{li2020face}
Li, L., J.~Bao, T.~Zhang, H.~Yang, D.~Chen, F.~Wen, and B.~Guo 2020.
\newblock Face x-ray for more general face forgery detection.
\newblock In {\em Conference on Computer Vision and Pattern Recognition}, pp.\
  5001--5010.

\bibitem[\protect\citeauthoryear{Li, Komulainen, Zhao, Yuen, and
  Pietik{\"a}inen}{Li et~al.}{2016}]{li2016generalized}
Li, X., J.~Komulainen, G.~Zhao, P.C. Yuen, and M.~Pietik{\"a}inen 2016.
\newblock Generalized face anti-spoofing by detecting pulse from face videos.
\newblock In {\em International Conference on Pattern Recognition}, pp.\
  4244--4249.

\bibitem[\protect\citeauthoryear{Li and Lyu}{Li and Lyu}{2018}]{li2018exposing}
Li, Y. and S.~Lyu. 2018.
\newblock Exposing deepfake videos by detecting face warping artifacts.
\newblock {\em arXiv preprint arXiv:1811.00656\/} .

\bibitem[\protect\citeauthoryear{Li, Yang, Sun, Qi, and Lyu}{Li
  et~al.}{2020}]{li2020celeb}
Li, Y., X.~Yang, P.~Sun, H.~Qi, and S.~Lyu 2020.
\newblock Celeb-{DF}: A large-scale challenging dataset for deepfake forensics.
\newblock In {\em Conference on Computer Vision and Pattern Recognition}, pp.\
  3207--3216.

\bibitem[\protect\citeauthoryear{Liao, Chen, Liu, Yeh, Hu, and Chen}{Liao
  et~al.}{2023}]{liao2023domain}
Liao, C.H., W.C. Chen, H.T. Liu, Y.R. Yeh, M.C. Hu, and C.S. Chen 2023.
\newblock Domain invariant vision transformer learning for face anti-spoofing.
\newblock In {\em Proceedings of the IEEE/CVF Winter Conference on Applications
  of Computer Vision}, pp.\  6098--6107.

\bibitem[\protect\citeauthoryear{Liu, Tan, Li, Wan, Escalera, Guo, and Li}{Liu
  et~al.}{2019}]{liu2019static}
Liu, A., Z.~Tan, X.~Li, J.~Wan, S.~Escalera, G.~Guo, and S.Z. Li. 2019.
\newblock Static and dynamic fusion for multi-modal cross-ethnicity face
  anti-spoofing.
\newblock {\em arXiv preprint arXiv:1912.02340\/} .

\bibitem[\protect\citeauthoryear{Liu, Tan, Wan, Liang, Lei, Guo, and Li}{Liu
  et~al.}{2021}]{liu2021face}
Liu, A., Z.~Tan, J.~Wan, Y.~Liang, Z.~Lei, G.~Guo, and S.Z. Li. 2021.
\newblock Face anti-spoofing via adversarial cross-modality translation.
\newblock {\em IEEE Transactions on Information Forensics and Security\/}~16:
  2759--2772 .

\bibitem[\protect\citeauthoryear{Liu, Tan, Yu, Zhao, Wan, Lei, Zhang, Li, and
  Guo}{Liu et~al.}{2023}]{liu2023fm}
Liu, A., Z.~Tan, Z.~Yu, C.~Zhao, J.~Wan, Y.L.Z. Lei, D.~Zhang, S.Z. Li, and
  G.~Guo. 2023.
\newblock Fm-vit: Flexible modal vision transformers for face anti-spoofing.
\newblock {\em IEEE Transactions on Information Forensics and Security\/} .

\bibitem[\protect\citeauthoryear{Liu, Zhao, Yu, Wan, Su, Liu, Tan, Escalera,
  Xing, Liang, et~al.}{Liu et~al.}{2022}]{liu2022contrastive}
Liu, A., C.~Zhao, Z.~Yu, J.~Wan, A.~Su, X.~Liu, Z.~Tan, S.~Escalera, J.~Xing,
  Y.~Liang, et~al. 2022.
\newblock Contrastive context-aware learning for 3d high-fidelity mask face
  presentation attack detection.
\newblock {\em IEEE Transactions on Information Forensics and Security\/}~17:
  2497--2507 .

\bibitem[\protect\citeauthoryear{Liu, Li, Zhou, Chen, He, Xue, Zhang, and
  Yu}{Liu et~al.}{2021}]{liu2021spatial}
Liu, H., X.~Li, W.~Zhou, Y.~Chen, Y.~He, H.~Xue, W.~Zhang, and N.~Yu 2021.
\newblock Spatial-phase shallow learning: rethinking face forgery detection in
  frequency domain.
\newblock In {\em Conference on Computer Vision and Pattern Recognition}, pp.\
  772--781.

\bibitem[\protect\citeauthoryear{Liu, Li, Xie, Jiang, Wang, Guo, and Chen}{Liu
  et~al.}{2020}]{liu2020livescreen}
Liu, H., Z.~Li, Y.~Xie, R.~Jiang, Y.~Wang, X.~Guo, and Y.~Chen 2020.
\newblock Livescreen: Video chat liveness detection leveraging skin reflection.
\newblock In {\em IEEE International Conference on Computer Communications},
  pp.\  1083--1092.

\bibitem[\protect\citeauthoryear{Liu, He, Xiao, Han, and Ren}{Liu
  et~al.}{2023}]{liu2023time}
Liu, J., Y.~He, C.~Xiao, J.~Han, and K.~Ren. 2023.
\newblock Time to think the security of wifi-based behavior recognition
  systems.
\newblock {\em IEEE Transactions on Dependable and Secure Computing\/} .

\bibitem[\protect\citeauthoryear{Liu, Huang, Yu, Wang, and Mallya}{Liu
  et~al.}{2021}]{liu2021generative}
Liu, M.Y., X.~Huang, J.~Yu, T.C. Wang, and A.~Mallya. 2021.
\newblock Generative adversarial networks for image and video synthesis:
  Algorithms and applications.
\newblock {\em Proceedings of the IEEE\/}~{\em 109\/}(5): 839--862 .

\bibitem[\protect\citeauthoryear{Liu, Lu, Xu, Yang, Ding, and Ma}{Liu
  et~al.}{2022}]{liu2022feature}
Liu, S., S.~Lu, H.~Xu, J.~Yang, S.~Ding, and L.~Ma 2022.
\newblock Feature generation and hypothesis verification for reliable face
  anti-spoofing.
\newblock In {\em Proceedings of the AAAI Conference on Artificial
  Intelligence}, Volume~36, pp.\  1782--1791.

\bibitem[\protect\citeauthoryear{Liu, Chen, Liu, and Song}{Liu
  et~al.}{2016}]{liu2016delving}
Liu, Y., X.~Chen, C.~Liu, and D.~Song. 2016.
\newblock Delving into transferable adversarial examples and black-box attacks.
\newblock {\em arXiv preprint arXiv:1611.02770\/} .

\bibitem[\protect\citeauthoryear{Liu, Chen, Gou, Huang, Wang, Dai, and
  Xiong}{Liu et~al.}{2023}]{liu2023towards}
Liu, Y., Y.~Chen, M.~Gou, C.T. Huang, Y.~Wang, W.~Dai, and H.~Xiong 2023.
\newblock Towards unsupervised domain generalization for face anti-spoofing.
\newblock In {\em Proceedings of the IEEE/CVF International Conference on
  Computer Vision}, pp.\  20654--20664.

\bibitem[\protect\citeauthoryear{Liu, Jourabloo, and Liu}{Liu
  et~al.}{2018}]{liu2018learning}
Liu, Y., A.~Jourabloo, and X.~Liu 2018.
\newblock Learning deep models for face anti-spoofing: Binary or auxiliary
  supervision.
\newblock In {\em Conference on Computer Vision and Pattern Recognition}, pp.\
  389--398.

\bibitem[\protect\citeauthoryear{Liu, Stehouwer, Jourabloo, and Liu}{Liu
  et~al.}{2019}]{liu2019deep}
Liu, Y., J.~Stehouwer, A.~Jourabloo, and X.~Liu 2019.
\newblock Deep tree learning for zero-shot face anti-spoofing.
\newblock In {\em Conference on Computer Vision and Pattern Recognition}, pp.\
  4680--4689.

\bibitem[\protect\citeauthoryear{Luo, Zhang, Yan, and Liu}{Luo
  et~al.}{2021}]{luo2021generalizing}
Luo, Y., Y.~Zhang, J.~Yan, and W.~Liu 2021.
\newblock Generalizing face forgery detection with high-frequency features.
\newblock In {\em Conference on Computer Vision and Pattern Recognition}, pp.\
  16317--16326.

\bibitem[\protect\citeauthoryear{Ma, Chai, Huh, Wang, Lim, Isola, and
  Torralba}{Ma et~al.}{2022}]{ma2022totems}
Ma, J., L.~Chai, M.~Huh, T.~Wang, S.N. Lim, P.~Isola, and A.~Torralba 2022.
\newblock Totems: Physical objects for verifying visual integrity.
\newblock In {\em European Conference on Computer Vision}, pp.\  164--180.

\bibitem[\protect\citeauthoryear{Madry, Makelov, Schmidt, Tsipras, and
  Vladu}{Madry et~al.}{2017}]{madry2017towards}
Madry, A., A.~Makelov, L.~Schmidt, D.~Tsipras, and A.~Vladu. 2017.
\newblock Towards deep learning models resistant to adversarial attacks.
\newblock {\em arXiv preprint arXiv:1706.06083\/} .

\bibitem[\protect\citeauthoryear{Metzen, Genewein, Fischer, and
  Bischoff}{Metzen et~al.}{2016}]{metzen2016detecting}
Metzen, J.H., T.~Genewein, V.~Fischer, and B.~Bischoff 2016.
\newblock On detecting adversarial perturbations.
\newblock In {\em International Conference on Learning Representations}.

\bibitem[\protect\citeauthoryear{Meyer, Reudenbach, Hengl, Katurji, and
  Nauss}{Meyer et~al.}{2018}]{meyer2018improving}
Meyer, H., C.~Reudenbach, T.~Hengl, M.~Katurji, and T.~Nauss. 2018.
\newblock Improving performance of spatio-temporal machine learning models
  using forward feature selection and target-oriented validation.
\newblock {\em Environmental Modelling \& Software\/}~101: 1--9 .

\bibitem[\protect\citeauthoryear{Mirsky and Lee}{Mirsky and
  Lee}{2021}]{mirsky2021creation}
Mirsky, Y. and W.~Lee. 2021.
\newblock The creation and detection of deepfakes: A survey.
\newblock {\em ACM computing surveys (CSUR)\/}~{\em 54\/}(1): 1--41 .

\bibitem[\protect\citeauthoryear{Mo and Sinopoli}{Mo and
  Sinopoli}{2009}]{mo2009secure}
Mo, Y. and B.~Sinopoli 2009.
\newblock Secure control against replay attacks.
\newblock In {\em 2009 47th annual Allerton conference on communication,
  control, and computing (Allerton)}, pp.\  911--918. IEEE.

\bibitem[\protect\citeauthoryear{Nagothu, Chen, Blasch, Aved, and Zhu}{Nagothu
  et~al.}{2019}]{nagothu2019detecting}
Nagothu, D., Y.~Chen, E.~Blasch, A.~Aved, and S.~Zhu. 2019.
\newblock Detecting malicious false frame injection attacks on surveillance
  systems at the edge using electrical network frequency signals.
\newblock {\em Sensors\/}~{\em 19\/}(11): 2424 .

\bibitem[\protect\citeauthoryear{Narayan, Agarwal, Thakral, Mittal, Vatsa, and
  Singh}{Narayan et~al.}{2023}]{narayan2023df}
Narayan, K., H.~Agarwal, K.~Thakral, S.~Mittal, M.~Vatsa, and R.~Singh 2023.
\newblock Df-platter: Multi-face heterogeneous deepfake dataset.
\newblock In {\em Conference on Computer Vision and Pattern Recognition}, pp.\
  9739--9748.

\bibitem[\protect\citeauthoryear{Narodytska and Kasiviswanathan}{Narodytska and
  Kasiviswanathan}{2017}]{narodytska2017simple}
Narodytska, N. and S.P. Kasiviswanathan 2017.
\newblock Simple black-box adversarial attacks on deep neural networks.
\newblock In {\em Conference on Computer Vision and Pattern Recognition
  Workshops}, Volume~2, pp.\ ~2.

\bibitem[\protect\citeauthoryear{Natsume, Yatagawa, and Morishima}{Natsume
  et~al.}{2018}]{natsume2018rsgan}
Natsume, R., T.~Yatagawa, and S.~Morishima. 2018.
\newblock Rsgan: face swapping and editing using face and hair representation
  in latent spaces.
\newblock {\em arXiv preprint arXiv:1804.03447\/} .

\bibitem[\protect\citeauthoryear{Natsume, Yatagawa, and Morishima}{Natsume
  et~al.}{2019}]{natsume2019fsnet}
Natsume, R., T.~Yatagawa, and S.~Morishima 2019.
\newblock Fsnet: An identity-aware generative model for image-based face
  swapping.
\newblock In {\em Asian Conference on Computer Vision}, pp.\  117--132.

\bibitem[\protect\citeauthoryear{Nguyen, Nguyen, Nguyen, Nguyen, Huynh-The,
  Nahavandi, Nguyen, Pham, and Nguyen}{Nguyen et~al.}{2022}]{nguyen2022deep}
Nguyen, T.T., Q.V.H. Nguyen, D.T. Nguyen, D.T. Nguyen, T.~Huynh-The,
  S.~Nahavandi, T.T. Nguyen, Q.V. Pham, and C.M. Nguyen. 2022.
\newblock Deep learning for deepfakes creation and detection: A survey.
\newblock {\em Computer Vision and Image Understanding\/}~223: 103525 .

\bibitem[\protect\citeauthoryear{Nichol, Dhariwal, Ramesh, Shyam, Mishkin,
  McGrew, Sutskever, and Chen}{Nichol et~al.}{2021}]{nichol2021glide}
Nichol, A., P.~Dhariwal, A.~Ramesh, P.~Shyam, P.~Mishkin, B.~McGrew,
  I.~Sutskever, and M.~Chen. 2021.
\newblock Glide: Towards photorealistic image generation and editing with
  text-guided diffusion models.
\newblock {\em arXiv preprint arXiv:2112.10741\/} .

\bibitem[\protect\citeauthoryear{Nikisins, Mohammadi, Anjos, and
  Marcel}{Nikisins et~al.}{2018}]{nikisins2018effectiveness}
Nikisins, O., A.~Mohammadi, A.~Anjos, and S.~Marcel 2018.
\newblock On effectiveness of anomaly detection approaches against unseen
  presentation attacks in face anti-spoofing.
\newblock In {\em 2018 International Conference on Biometrics (ICB)}, pp.\
  75--81. IEEE.

\bibitem[\protect\citeauthoryear{Nirkin, Keller, and Hassner}{Nirkin
  et~al.}{2019}]{nirkin2019fsgan}
Nirkin, Y., Y.~Keller, and T.~Hassner 2019.
\newblock Fsgan: Subject agnostic face swapping and reenactment.
\newblock In {\em International Conference on Computer Vision}, pp.\
  7184--7193.

\bibitem[\protect\citeauthoryear{Papernot, McDaniel, Jha, Fredrikson, Celik,
  and Swami}{Papernot et~al.}{2016}]{papernot2016limitations}
Papernot, N., P.~McDaniel, S.~Jha, M.~Fredrikson, Z.B. Celik, and A.~Swami
  2016.
\newblock The limitations of deep learning in adversarial settings.
\newblock In {\em IEEE European Symposium on Security and Privacy}, pp.\
  372--387.

\bibitem[\protect\citeauthoryear{P{\'e}rez-Cabo, Jim{\'e}nez-Cabello,
  Costa-Pazo, and L{\'o}pez-Sastre}{P{\'e}rez-Cabo
  et~al.}{2019}]{perez2019deep}
P{\'e}rez-Cabo, D., D.~Jim{\'e}nez-Cabello, A.~Costa-Pazo, and R.J.
  L{\'o}pez-Sastre 2019.
\newblock Deep anomaly detection for generalized face anti-spoofing.
\newblock In {\em Proceedings of the IEEE/CVF Conference on Computer Vision and
  Pattern Recognition Workshops}, pp.\  0--0.

\bibitem[\protect\citeauthoryear{Prajwal, Mukhopadhyay, Namboodiri, and
  Jawahar}{Prajwal et~al.}{2020}]{prajwal2020lip}
Prajwal, K., R.~Mukhopadhyay, V.P. Namboodiri, and C.~Jawahar 2020.
\newblock A lip sync expert is all you need for speech to lip generation in the
  wild.
\newblock In {\em ACM International Conference on Multimedia}, pp.\  484--492.

\bibitem[\protect\citeauthoryear{Radford, Kim, Hallacy, Ramesh, Goh, Agarwal,
  Sastry, Askell, Mishkin, Clark, et~al.}{Radford
  et~al.}{2021}]{radford2021learning}
Radford, A., J.W. Kim, C.~Hallacy, A.~Ramesh, G.~Goh, S.~Agarwal, G.~Sastry,
  A.~Askell, P.~Mishkin, J.~Clark, et~al. 2021.
\newblock Learning transferable visual models from natural language
  supervision.
\newblock In {\em International conference on machine learning}, pp.\
  8748--8763. PMLR.

\bibitem[\protect\citeauthoryear{Raghunathan, Steinhardt, and
  Liang}{Raghunathan et~al.}{2018}]{raghunathan2018certified}
Raghunathan, A., J.~Steinhardt, and P.~Liang 2018.
\newblock Certified defenses against adversarial examples.
\newblock In {\em International Conference on Learning Representations}.

\bibitem[\protect\citeauthoryear{Ramachandra and Busch}{Ramachandra and
  Busch}{2017}]{ramachandra2017presentation}
Ramachandra, R. and C.~Busch. 2017.
\newblock Presentation attack detection methods for face recognition systems: A
  comprehensive survey.
\newblock {\em ACM Computing Surveys (CSUR)\/}~{\em 50\/}(1): 1--37 .

\bibitem[\protect\citeauthoryear{Rana, Nobi, Murali, and Sung}{Rana
  et~al.}{2022}]{rana2022deepfake}
Rana, M.S., M.N. Nobi, B.~Murali, and A.H. Sung. 2022.
\newblock Deepfake detection: A systematic literature review.
\newblock {\em IEEE access\/}~10: 25494--25513 .

\bibitem[\protect\citeauthoryear{Ratha, Connell, and Bolle}{Ratha
  et~al.}{2001}]{ratha2001enhancing}
Ratha, N.K., J.H. Connell, and R.M. Bolle. 2001.
\newblock Enhancing security and privacy in biometrics-based authentication
  systems.
\newblock {\em IBM Systems Journal\/}~{\em 40\/}(3): 614--634 .

\bibitem[\protect\citeauthoryear{Rescorla}{Rescorla}{2001}]{rescorla2001ssl}
Rescorla, E. 2001.
\newblock Ssl and tls: designing and building secure systems.
\newblock {\em Pearson Education\/} .

\bibitem[\protect\citeauthoryear{Rombach, Blattmann, Lorenz, Esser, and
  Ommer}{Rombach et~al.}{2022}]{rombach2022high}
Rombach, R., A.~Blattmann, D.~Lorenz, P.~Esser, and B.~Ommer 2022.
\newblock High-resolution image synthesis with latent diffusion models.
\newblock In {\em Conference on Computer Vision and Pattern Recognition}, pp.\
  10684--10695.

\bibitem[\protect\citeauthoryear{Rossler, Cozzolino, Verdoliva, Riess, Thies,
  and Nie{\ss}ner}{Rossler et~al.}{2019}]{rossler2019faceforensics++}
Rossler, A., D.~Cozzolino, L.~Verdoliva, C.~Riess, J.~Thies, and M.~Nie{\ss}ner
  2019.
\newblock Faceforensics++: Learning to detect manipulated facial images.
\newblock In {\em International Conference on Computer Vision}, pp.\  1--11.

\bibitem[\protect\citeauthoryear{Saha, Subramanya, and Pirsiavash}{Saha
  et~al.}{2020}]{saha2020hidden}
Saha, A., A.~Subramanya, and H.~Pirsiavash 2020.
\newblock Hidden trigger backdoor attacks.
\newblock In {\em Proceedings of the AAAI conference on artificial
  intelligence}, Volume~34, pp.\  11957--11965.

\bibitem[\protect\citeauthoryear{Sandha, Noor, Anwar, and Srivastava}{Sandha
  et~al.}{2020}]{sandha2020time}
Sandha, S.S., J.~Noor, F.M. Anwar, and M.~Srivastava 2020.
\newblock Time awareness in deep learning-based multimodal fusion across
  smartphone platforms.
\newblock In {\em ACM/IEEE International Conference on Internet of Things
  Design and Implementation}, pp.\  149--156.

\bibitem[\protect\citeauthoryear{Sarafianos, Xu, and Kakadiaris}{Sarafianos
  et~al.}{2019}]{sarafianos2019adversarial}
Sarafianos, N., X.~Xu, and I.A. Kakadiaris 2019.
\newblock Adversarial representation learning for text-to-image matching.
\newblock In {\em Proceedings of the IEEE/CVF international conference on
  computer vision}, pp.\  5814--5824.

\bibitem[\protect\citeauthoryear{Shao, Wu, and Liu}{Shao
  et~al.}{2023}]{shao2023detecting}
Shao, R., T.~Wu, and Z.~Liu 2023.
\newblock Detecting and grounding multi-modal media manipulation.
\newblock In {\em Proceedings of the IEEE/CVF Conference on Computer Vision and
  Pattern Recognition}, pp.\  6904--6913.

\bibitem[\protect\citeauthoryear{Sharif, Bhagavatula, Bauer, and Reiter}{Sharif
  et~al.}{2019}]{sharif2019general}
Sharif, M., S.~Bhagavatula, L.~Bauer, and M.K. Reiter. 2019.
\newblock A general framework for adversarial examples with objectives.
\newblock {\em ACM Transactions on Privacy and Security\/}~{\em 22\/}(3): 1--30
  .

\bibitem[\protect\citeauthoryear{Shu, Nie, Huang, Yu, Goldstein, Anandkumar,
  and Xiao}{Shu et~al.}{2022}]{shu2022test}
Shu, M., W.~Nie, D.A. Huang, Z.~Yu, T.~Goldstein, A.~Anandkumar, and C.~Xiao.
  2022.
\newblock Test-time prompt tuning for zero-shot generalization in
  vision-language models.
\newblock {\em Advances in Neural Information Processing Systems\/}~35:
  14274--14289 .

\bibitem[\protect\citeauthoryear{Siarohin, Lathuili{\`e}re, Tulyakov, Ricci,
  and Sebe}{Siarohin et~al.}{2019}]{siarohin2019first}
Siarohin, A., S.~Lathuili{\`e}re, S.~Tulyakov, E.~Ricci, and N.~Sebe 2019.
\newblock First order motion model for image animation.
\newblock In {\em Advances in Neural Information Processing Systems}, pp.\
  7137--7147.

\bibitem[\protect\citeauthoryear{Singh and Pandey}{Singh and
  Pandey}{2018}]{singh2018revisiting}
Singh, V. and S.~Pandey 2018.
\newblock Revisiting cloud security threats: replay attack.
\newblock In {\em 2018 4th International Conference on Computing Communication
  and Automation (ICCCA)}, pp.\  1--6. IEEE.

\bibitem[\protect\citeauthoryear{Smith, Wiliem, and Lovell}{Smith
  et~al.}{2015}]{smith2015face}
Smith, D.F., A.~Wiliem, and B.C. Lovell. 2015.
\newblock Face recognition on consumer devices: Reflections on replay attacks.
\newblock {\em IEEE Transactions on Information Forensics and Security\/}~{\em
  10\/}(4): 736--745 .

\bibitem[\protect\citeauthoryear{Song, Yan, Luo, and Tan}{Song
  et~al.}{2022}]{song2022sardino}
Song, Q., Z.~Yan, W.~Luo, and R.~Tan 2022.
\newblock Sardino: Ultra-fast dynamic ensemble for secure visual sensing at
  mobile edge.
\newblock In {\em International Conference on Embedded Wireless Systems and
  Networks}, pp.\  24--35.

\bibitem[\protect\citeauthoryear{Srivatsan, Naseer, and Nandakumar}{Srivatsan
  et~al.}{2023}]{srivatsan2023flip}
Srivatsan, K., M.~Naseer, and K.~Nandakumar 2023.
\newblock Flip: Cross-domain face anti-spoofing with language guidance.
\newblock In {\em Proceedings of the IEEE/CVF International Conference on
  Computer Vision}, pp.\  19685--19696.

\bibitem[\protect\citeauthoryear{Sun, Liu, Liu, Li, and Chu}{Sun
  et~al.}{2023}]{sun2023rethinking}
Sun, Y., Y.~Liu, X.~Liu, Y.~Li, and W.S. Chu 2023.
\newblock Rethinking domain generalization for face anti-spoofing: Separability
  and alignment.
\newblock In {\em Conference on Computer Vision and Pattern Recognition}, pp.\
  24563--24574.

\bibitem[\protect\citeauthoryear{Szegedy, Zaremba, Sutskever, Bruna, Erhan,
  Goodfellow, and Fergus}{Szegedy et~al.}{2013}]{szegedy2013intriguing}
Szegedy, C., W.~Zaremba, I.~Sutskever, J.~Bruna, D.~Erhan, I.~Goodfellow, and
  R.~Fergus. 2013.
\newblock Intriguing properties of neural networks.
\newblock {\em arXiv preprint arXiv:1312.6199\/} .

\bibitem[\protect\citeauthoryear{Tan and Le}{Tan and
  Le}{2019}]{tan2019efficientnet}
Tan, M. and Q.~Le 2019.
\newblock Efficientnet: Rethinking model scaling for convolutional neural
  networks.
\newblock In {\em International conference on machine learning}, pp.\
  6105--6114. PMLR.

\bibitem[\protect\citeauthoryear{Tang, Zhou, Zhang, and Zhang}{Tang
  et~al.}{2018}]{tang2018face}
Tang, D., Z.~Zhou, Y.~Zhang, and K.~Zhang. 2018.
\newblock Face flashing: a secure liveness detection protocol based on light
  reflections.
\newblock {\em arXiv preprint arXiv:1801.01949\/} .

\bibitem[\protect\citeauthoryear{Thies, Zollh{\"o}fer, and Nie{\ss}ner}{Thies
  et~al.}{2019}]{thies2019deferred}
Thies, J., M.~Zollh{\"o}fer, and M.~Nie{\ss}ner. 2019.
\newblock Deferred neural rendering: Image synthesis using neural textures.
\newblock {\em ACM Transactions on Graphics\/}~{\em 38\/}(4): 1--12 .

\bibitem[\protect\citeauthoryear{Thullier, Bouchard, and Menelas}{Thullier
  et~al.}{2017}]{thullier2017text}
Thullier, F., B.~Bouchard, and B.A.J. Menelas. 2017.
\newblock A text-independent speaker authentication system for mobile devices.
\newblock {\em Cryptography\/}~{\em 1\/}(3): 16 .

\bibitem[\protect\citeauthoryear{Tolosana, Vera-Rodriguez, Fierrez, Morales,
  and Ortega-Garcia}{Tolosana et~al.}{2020}]{tolosana2020deepfakes}
Tolosana, R., R.~Vera-Rodriguez, J.~Fierrez, A.~Morales, and J.~Ortega-Garcia.
  2020.
\newblock Deepfakes and beyond: A survey of face manipulation and fake
  detection.
\newblock {\em Information Fusion\/}~64: 131--148 .

\bibitem[\protect\citeauthoryear{Trinh, Tsang, Rambhatla, and Liu}{Trinh
  et~al.}{2021}]{trinh2021interpretable}
Trinh, L., M.~Tsang, S.~Rambhatla, and Y.~Liu 2021.
\newblock Interpretable and trustworthy deepfake detection via dynamic
  prototypes.
\newblock In {\em Proceedings of the IEEE/CVF winter conference on applications
  of computer vision}, pp.\  1973--1983.

\bibitem[\protect\citeauthoryear{Tripathy, Kannala, and Rahtu}{Tripathy
  et~al.}{2020}]{tripathy2020icface}
Tripathy, S., J.~Kannala, and E.~Rahtu 2020.
\newblock Icface: Interpretable and controllable face reenactment using gans.
\newblock In {\em Winter Conference on Applications of Computer Vision}, pp.\
  3385--3394.

\bibitem[\protect\citeauthoryear{Walia, Rohilla, et~al.}{Walia
  et~al.}{2021}]{walia2021contemporary}
Walia, G.S., R.~Rohilla, et~al. 2021.
\newblock A contemporary survey of multimodal presentation attack detection
  techniques: Challenges and opportunities.
\newblock {\em SN Computer Science\/}~{\em 2\/}(1): 1--7 .

\bibitem[\protect\citeauthoryear{Wan, Guo, Escalera, Escalante, and Li}{Wan
  et~al.}{2023}]{wan2023advances}
Wan, J., G.~Guo, S.~Escalera, H.J. Escalante, and S.Z. Li. 2023.
\newblock {\em Advances in face presentation attack detection}.
\newblock Springer.

\bibitem[\protect\citeauthoryear{Wang, Yao, Shan, Li, Viswanath, Zheng, and
  Zhao}{Wang et~al.}{2019}]{wang2019neural}
Wang, B., Y.~Yao, S.~Shan, H.~Li, B.~Viswanath, H.~Zheng, and B.Y. Zhao 2019.
\newblock Neural cleanse: Identifying and mitigating backdoor attacks in neural
  networks.
\newblock In {\em IEEE Symposium on Security and Privacy}, pp.\  707--723.

\bibitem[\protect\citeauthoryear{Wang, Lu, Yang, and Lai}{Wang
  et~al.}{2022}]{wang2022patchnet}
Wang, C.Y., Y.D. Lu, S.T. Yang, and S.H. Lai 2022.
\newblock Patchnet: A simple face anti-spoofing framework via fine-grained
  patch recognition.
\newblock In {\em Proceedings of the IEEE/CVF Conference on Computer Vision and
  Pattern Recognition}, pp.\  20281--20290.

\bibitem[\protect\citeauthoryear{Wang, Guo, Shao, He, Chen, Xiao, Liu,
  Escalera, Escalante, Lei, et~al.}{Wang et~al.}{2023}]{wang2023wild}
Wang, D., J.~Guo, Q.~Shao, H.~He, Z.~Chen, C.~Xiao, A.~Liu, S.~Escalera, H.J.
  Escalante, Z.~Lei, et~al. 2023.
\newblock Wild face anti-spoofing challenge 2023: Benchmark and results.
\newblock In {\em Conference on Computer Vision and Pattern Recognition}, pp.\
  6379--6390.

\bibitem[\protect\citeauthoryear{Wang, Han, Shan, and Chen}{Wang
  et~al.}{2020}]{wang2020cross}
Wang, G., H.~Han, S.~Shan, and X.~Chen 2020.
\newblock Cross-domain face presentation attack detection via multi-domain
  disentangled representation learning.
\newblock In {\em Conference on Computer Vision and Pattern Recognition}, pp.\
  6678--6687.

\bibitem[\protect\citeauthoryear{Wang, Chen, Gui, Hu, Liu, and Wang}{Wang
  et~al.}{2020}]{wang2020once}
Wang, H., T.~Chen, S.~Gui, T.~Hu, J.~Liu, and Z.~Wang. 2020.
\newblock Once-for-all adversarial training: In-situ tradeoff between
  robustness and accuracy for free.
\newblock {\em Advances in Neural Information Processing Systems\/}~33:
  7449--7461 .

\bibitem[\protect\citeauthoryear{Wang, Wu, Ouyang, Han, Chen, Jiang, and
  Li}{Wang et~al.}{2022}]{wang2022m2tr}
Wang, J., Z.~Wu, W.~Ouyang, X.~Han, J.~Chen, Y.G. Jiang, and S.N. Li 2022.
\newblock M2tr: Multi-modal multi-scale transformers for deepfake detection.
\newblock In {\em International Conference on Multimedia Retrieval}, pp.\
  615--623.

\bibitem[\protect\citeauthoryear{Wang, Wang, Zhang, Owens, and Efros}{Wang
  et~al.}{2020}]{wang2020cnn}
Wang, S.Y., O.~Wang, R.~Zhang, A.~Owens, and A.A. Efros 2020.
\newblock Cnn-generated images are surprisingly easy to spot... for now.
\newblock In {\em Conference on Computer Vision and Pattern Recognition}, pp.\
  8695--8704.

\bibitem[\protect\citeauthoryear{Wang, Liu, Zheng, Ying, and Wen}{Wang
  et~al.}{2023}]{wang2023domain}
Wang, W., P.~Liu, H.~Zheng, R.~Ying, and F.~Wen. 2023.
\newblock Domain generalization for face anti-spoofing via negative data
  augmentation.
\newblock {\em IEEE Transactions on Information Forensics and Security\/} .

\bibitem[\protect\citeauthoryear{Wang, Yu, Chen, Hu, and Peng}{Wang
  et~al.}{2023}]{wang2023dynamic}
Wang, Y., K.~Yu, C.~Chen, X.~Hu, and S.~Peng 2023.
\newblock Dynamic graph learning with content-guided spatial-frequency relation
  reasoning for deepfake detection.
\newblock In {\em Proceedings of the IEEE/CVF Conference on Computer Vision and
  Pattern Recognition}, pp.\  7278--7287.

\bibitem[\protect\citeauthoryear{Wang, Bao, Zhou, Wang, and Li}{Wang
  et~al.}{2023}]{wang2023altfreezing}
Wang, Z., J.~Bao, W.~Zhou, W.~Wang, and H.~Li 2023.
\newblock Altfreezing for more general video face forgery detection.
\newblock In {\em Proceedings of the IEEE/CVF Conference on Computer Vision and
  Pattern Recognition}, pp.\  4129--4138.

\bibitem[\protect\citeauthoryear{Wang, Wang, Deng, and Guo}{Wang
  et~al.}{2022}]{wang2022face}
Wang, Z., Q.~Wang, W.~Deng, and G.~Guo. 2022.
\newblock Face anti-spoofing using transformers with relation-aware mechanism.
\newblock {\em IEEE Transactions on Biometrics, Behavior, and Identity
  Science\/}~{\em 4\/}(3): 439--450 .

\bibitem[\protect\citeauthoryear{Wang, Wang, Yu, Deng, Li, Gao, and Wang}{Wang
  et~al.}{2022}]{wang2022domain}
Wang, Z., Z.~Wang, Z.~Yu, W.~Deng, J.~Li, T.~Gao, and Z.~Wang 2022.
\newblock Domain generalization via shuffled style assembly for face
  anti-spoofing.
\newblock In {\em Proceedings of the IEEE/CVF Conference on Computer Vision and
  Pattern Recognition}, pp.\  4123--4133.

\bibitem[\protect\citeauthoryear{Wang, Yu, Zhao, Zhu, Qin, Zhou, Zhou, and
  Lei}{Wang et~al.}{2020}]{wang2020deep}
Wang, Z., Z.~Yu, C.~Zhao, X.~Zhu, Y.~Qin, Q.~Zhou, F.~Zhou, and Z.~Lei 2020.
\newblock Deep spatial gradient and temporal depth learning for face
  anti-spoofing.
\newblock In {\em Proceedings of the IEEE/CVF conference on computer vision and
  pattern recognition}, pp.\  5042--5051.

\bibitem[\protect\citeauthoryear{Wen, Han, and Jain}{Wen
  et~al.}{2015}]{wen2015face}
Wen, D., H.~Han, and A.K. Jain. 2015.
\newblock Face spoof detection with image distortion analysis.
\newblock {\em IEEE Transactions on Information Forensics and Security\/}~{\em
  10\/}(4): 746--761 .

\bibitem[\protect\citeauthoryear{Weng, Lee, and Wu}{Weng
  et~al.}{2020}]{weng2020trade}
Weng, C.H., Y.T. Lee, and S.H.B. Wu. 2020.
\newblock On the trade-off between adversarial and backdoor robustness.
\newblock {\em Advances in Neural Information Processing Systems\/}~33:
  11973--11983 .

\bibitem[\protect\citeauthoryear{Wiles, Koepke, and Zisserman}{Wiles
  et~al.}{2018}]{wiles2018x2face}
Wiles, O., A.~Koepke, and A.~Zisserman 2018.
\newblock X2face: A network for controlling face generation using images,
  audio, and pose codes.
\newblock In {\em European Conference on Computer Vision}, pp.\  670--686.

\bibitem[\protect\citeauthoryear{Wong, Rice, and Kolter}{Wong
  et~al.}{2020}]{wong2020fast}
Wong, E., L.~Rice, and J.Z. Kolter. 2020.
\newblock Fast is better than free: Revisiting adversarial training.
\newblock {\em arXiv preprint arXiv:2001.03994\/} .

\bibitem[\protect\citeauthoryear{Xiao, Chen, Jin, Wang, Nie, Liu, Anandkumar,
  Li, and Song}{Xiao et~al.}{2022}]{xiao2022densepure}
Xiao, C., Z.~Chen, K.~Jin, J.~Wang, W.~Nie, M.~Liu, A.~Anandkumar, B.~Li, and
  D.~Song 2022.
\newblock Densepure: Understanding diffusion models for adversarial robustness.
\newblock In {\em The Eleventh International Conference on Learning
  Representations}.

\bibitem[\protect\citeauthoryear{Xiao, Gao, Fu, Dong, Gao, Zhang, Zhou, and
  Zhu}{Xiao et~al.}{2021}]{xiao2021improving}
Xiao, Z., X.~Gao, C.~Fu, Y.~Dong, W.~Gao, X.~Zhang, J.~Zhou, and J.~Zhu 2021.
\newblock Improving transferability of adversarial patches on face recognition
  with generative models.
\newblock In {\em Conference on Computer Vision and Pattern Recognition}, pp.\
  11845--11854.

\bibitem[\protect\citeauthoryear{Xu, Xiong, and Xia}{Xu
  et~al.}{2021}]{xu2021improving}
Xu, X., Y.~Xiong, and W.~Xia 2021.
\newblock On improving temporal consistency for online face liveness detection
  system.
\newblock In {\em Proceedings of the IEEE/CVF International Conference on
  Computer Vision Workshops}, pp.\  824--833.

\bibitem[\protect\citeauthoryear{Xu, Zhou, Venkatesan, Swaminathan, and
  Majumder}{Xu et~al.}{2019}]{xu2019d}
Xu, X., X.~Zhou, R.~Venkatesan, G.~Swaminathan, and O.~Majumder 2019.
\newblock d-sne: Domain adaptation using stochastic neighborhood embedding.
\newblock In {\em Proceedings of the IEEE/CVF Conference on Computer Vision and
  Pattern Recognition}, pp.\  2497--2506.

\bibitem[\protect\citeauthoryear{Yan, Zhang, Fan, and Wu}{Yan
  et~al.}{2023}]{yan2023ucf}
Yan, Z., Y.~Zhang, Y.~Fan, and B.~Wu 2023.
\newblock Ucf: Uncovering common features for generalizable deepfake detection.
\newblock In {\em Proceedings of the IEEE/CVF International Conference on
  Computer Vision}, pp.\  22412--22423.

\bibitem[\protect\citeauthoryear{Yang, Kortylewski, Xie, Cao, and Yuille}{Yang
  et~al.}{2020}]{yang2020patchattack}
Yang, C., A.~Kortylewski, C.~Xie, Y.~Cao, and A.~Yuille 2020.
\newblock Patchattack: A black-box texture-based attack with reinforcement
  learning.
\newblock In {\em European Conference on Computer Vision}, pp.\  681--698.

\bibitem[\protect\citeauthoryear{Yang, Zhou, Chen, Guo, Ba, Xia, Cao, and
  Ren}{Yang et~al.}{2023}]{yang2023avoid}
Yang, W., X.~Zhou, Z.~Chen, B.~Guo, Z.~Ba, Z.~Xia, X.~Cao, and K.~Ren. 2023.
\newblock Avoid-df: Audio-visual joint learning for detecting deepfake.
\newblock {\em IEEE Transactions on Information Forensics and Security\/}~18:
  2015--2029 .

\bibitem[\protect\citeauthoryear{Yang, Luo, Bao, Gao, Gong, Zheng, Li, and
  Liu}{Yang et~al.}{2019}]{yang2019face}
Yang, X., W.~Luo, L.~Bao, Y.~Gao, D.~Gong, S.~Zheng, Z.~Li, and W.~Liu 2019.
\newblock Face anti-spoofing: Model matters, so does data.
\newblock In {\em Proceedings of the IEEE/CVF Conference on Computer Vision and
  Pattern Recognition}, pp.\  3507--3516.

\bibitem[\protect\citeauthoryear{Ye, Xu, Zhang, and Wu}{Ye
  et~al.}{2024}]{Ye2024}
Ye, M., X.~Xu, Q.~Zhang, and J.~Wu 2024.
\newblock Sharpness-aware optimization for real-world adversarial attacks for
  diverse compute platforms with enhanced transferability.
\newblock In {\em Proceedings of the IEEE/CVF Conference on Computer Vision and
  Pattern Recognition}.

\bibitem[\protect\citeauthoryear{Yin, Lu, Li, and Huang}{Yin
  et~al.}{2023}]{yin2023dynamic}
Yin, Q., W.~Lu, B.~Li, and J.~Huang. 2023.
\newblock Dynamic difference learning with spatio-temporal correlation for
  deepfake video detection.
\newblock {\em IEEE Transactions on Information Forensics and Security\/} .

\bibitem[\protect\citeauthoryear{Yu, Qin, Li, Zhao, Lei, and Zhao}{Yu
  et~al.}{2023}]{yu2022deep}
Yu, Z., Y.~Qin, X.~Li, C.~Zhao, Z.~Lei, and G.~Zhao. 2023.
\newblock Deep learning for face anti-spoofing: A survey.
\newblock {\em IEEE Transactions on Pattern Analysis and Machine
  Intelligence\/}~{\em 45\/}(5): 5609--5631 .

\bibitem[\protect\citeauthoryear{Yu, Zhao, Wang, Qin, Su, Li, Zhou, and
  Zhao}{Yu et~al.}{2020}]{yu2020searching}
Yu, Z., C.~Zhao, Z.~Wang, Y.~Qin, Z.~Su, X.~Li, F.~Zhou, and G.~Zhao 2020.
\newblock Searching central difference convolutional networks for face
  anti-spoofing.
\newblock In {\em Conference on Computer Vision and Pattern Recognition}, pp.\
  5295--5305.

\bibitem[\protect\citeauthoryear{Zakharov, Shysheya, Burkov, and
  Lempitsky}{Zakharov et~al.}{2019}]{zakharov2019few}
Zakharov, E., A.~Shysheya, E.~Burkov, and V.~Lempitsky 2019.
\newblock Few-shot adversarial learning of realistic neural talking head
  models.
\newblock In {\em International Conference on Computer Vision}, pp.\
  9459--9468.

\bibitem[\protect\citeauthoryear{Zhang, Tai, Yao, Meng, Ding, Wang, Li, Huang,
  and Ji}{Zhang et~al.}{2021}]{zhang2021aurora}
Zhang, J., Y.~Tai, T.~Yao, J.~Meng, S.~Ding, C.~Wang, J.~Li, F.~Huang, and
  R.~Ji. 2021.
\newblock Aurora guard: Reliable face anti-spoofing via mobile lighting system.
\newblock {\em arXiv preprint arXiv:2102.00713\/} .

\bibitem[\protect\citeauthoryear{Zhang, Liu, Wan, Liang, Guo, Escalera,
  Escalante, and Li}{Zhang et~al.}{2020}]{zhang2020casia}
Zhang, S., A.~Liu, J.~Wan, Y.~Liang, G.~Guo, S.~Escalera, H.J. Escalante, and
  S.Z. Li. 2020.
\newblock Casia-surf: A large-scale multi-modal benchmark for face
  anti-spoofing.
\newblock {\em IEEE Transactions on Biometrics, Behavior, and Identity
  Science\/}~{\em 2\/}(2): 182--193 .

\bibitem[\protect\citeauthoryear{Zhang, Ye, Huang, Ye, Cao, Zhang, and
  Yang}{Zhang et~al.}{2023}]{zhang2023understanding}
Zhang, X., H.~Ye, Z.~Huang, X.~Ye, Y.~Cao, Y.~Zhang, and M.~Yang. 2023.
\newblock Understanding the (in)security of cross-side face verification
  systems in mobile apps: A system perspective.
\newblock {\em IEEE Symposium on Security and Privacy\/}: 934--950.
\newblock \doi{10.1109/SP46215.2023.10179474} .

\bibitem[\protect\citeauthoryear{Zhang, Wei, Jiang, Zhang, Zuo, and Tian}{Zhang
  et~al.}{2023}]{zhang2023controlvideo}
Zhang, Y., Y.~Wei, D.~Jiang, X.~Zhang, W.~Zuo, and Q.~Tian. 2023.
\newblock Controlvideo: Training-free controllable text-to-video generation.
\newblock {\em arXiv preprint arXiv:2305.13077\/} .

\bibitem[\protect\citeauthoryear{Zhang, Yin, Li, Yin, Yan, Shao, and Liu}{Zhang
  et~al.}{2020}]{zhang2020celeba}
Zhang, Y., Z.~Yin, Y.~Li, G.~Yin, J.~Yan, J.~Shao, and Z.~Liu 2020.
\newblock Celeba-spoof: Large-scale face anti-spoofing dataset with rich
  annotations.
\newblock In {\em European Conference on Computer Vision}, pp.\  70--85.

\bibitem[\protect\citeauthoryear{Zhang, Yan, Liu, Lei, Yi, and Li}{Zhang
  et~al.}{2012}]{zhang2012face}
Zhang, Z., J.~Yan, S.~Liu, Z.~Lei, D.~Yi, and S.Z. Li 2012.
\newblock A face antispoofing database with diverse attacks.
\newblock In {\em International Conference on Biometrics}, pp.\  26--31.

\bibitem[\protect\citeauthoryear{Zhao, Wang, Hu, Chen, Liu, and Tang}{Zhao
  et~al.}{2023}]{zhao2023istvt}
Zhao, C., C.~Wang, G.~Hu, H.~Chen, C.~Liu, and J.~Tang. 2023.
\newblock Istvt: interpretable spatial-temporal video transformer for deepfake
  detection.
\newblock {\em IEEE Transactions on Information Forensics and Security\/}~18:
  1335--1348 .

\bibitem[\protect\citeauthoryear{Zhao, Zhou, Chen, Wei, Zhang, and Yu}{Zhao
  et~al.}{2021}]{zhao2021multi}
Zhao, H., W.~Zhou, D.~Chen, T.~Wei, W.~Zhang, and N.~Yu 2021.
\newblock Multi-attentional deepfake detection.
\newblock In {\em Proceedings of the IEEE/CVF conference on computer vision and
  pattern recognition}, pp.\  2185--2194.

\bibitem[\protect\citeauthoryear{Zhao, Xu, Xu, Ding, Xiong, and Xia}{Zhao
  et~al.}{2021}]{zhao2021learning}
Zhao, T., X.~Xu, M.~Xu, H.~Ding, Y.~Xiong, and W.~Xia 2021.
\newblock Learning self-consistency for deepfake detection.
\newblock In {\em International Conference on Computer Vision}, pp.\
  15023--15033.

\bibitem[\protect\citeauthoryear{Zheng, Liu, Dai, Li, Zou, and Xiong}{Zheng
  et~al.}{2023}]{zheng2023learning}
Zheng, G., Y.~Liu, W.~Dai, C.~Li, J.~Zou, and H.~Xiong 2023.
\newblock Learning causal representations for generalizable face anti spoofing.
\newblock In {\em ICASSP 2023-2023 IEEE International Conference on Acoustics,
  Speech and Signal Processing (ICASSP)}, pp.\  1--5. IEEE.

\bibitem[\protect\citeauthoryear{Zhou, Zhang, Yao, Lu, Yi, Ding, and Ma}{Zhou
  et~al.}{2023}]{zhou2023instance}
Zhou, Q., K.Y. Zhang, T.~Yao, X.~Lu, R.~Yi, S.~Ding, and L.~Ma 2023.
\newblock Instance-aware domain generalization for face anti-spoofing.
\newblock In {\em Conference on Computer Vision and Pattern Recognition}, pp.\
  20453--20463.

\bibitem[\protect\citeauthoryear{Zhou, Zhang, Yao, Yi, Ding, and Ma}{Zhou
  et~al.}{2022}]{zhou2022adaptive}
Zhou, Q., K.Y. Zhang, T.~Yao, R.~Yi, S.~Ding, and L.~Ma 2022.
\newblock Adaptive mixture of experts learning for generalizable face
  anti-spoofing.
\newblock In {\em Proceedings of the 30th ACM international conference on
  multimedia}, pp.\  6009--6018.

\bibitem[\protect\citeauthoryear{Zhou, Zhang, Yao, Yi, Sheng, Ding, and
  Ma}{Zhou et~al.}{2022}]{zhou2022generative}
Zhou, Q., K.Y. Zhang, T.~Yao, R.~Yi, K.~Sheng, S.~Ding, and L.~Ma 2022.
\newblock Generative domain adaptation for face anti-spoofing.
\newblock In {\em European Conference on Computer Vision}, pp.\  335--356.

\bibitem[\protect\citeauthoryear{Zhou, Wang, Liang, and Shen}{Zhou
  et~al.}{2021}]{zhou2021face}
Zhou, T., W.~Wang, Z.~Liang, and J.~Shen 2021.
\newblock Face forensics in the wild.
\newblock In {\em Proceedings of the IEEE/CVF conference on computer vision and
  pattern recognition}, pp.\  5778--5788.

\bibitem[\protect\citeauthoryear{Zhou and Lim}{Zhou and
  Lim}{2021}]{zhou2021joint}
Zhou, Y. and S.N. Lim 2021.
\newblock Joint audio-visual deepfake detection.
\newblock In {\em Proceedings of the IEEE/CVF International Conference on
  Computer Vision}, pp.\  14800--14809.

\bibitem[\protect\citeauthoryear{Zhu, Li, Wang, Xu, and Sun}{Zhu
  et~al.}{2021}]{zhu2021one}
Zhu, Y., Q.~Li, J.~Wang, C.Z. Xu, and Z.~Sun 2021.
\newblock One shot face swapping on megapixels.
\newblock In {\em Conference on Computer Vision and Pattern Recognition}, pp.\
  4834--4844.

\bibitem[\protect\citeauthoryear{Zi, Chang, Chen, Ma, and Jiang}{Zi
  et~al.}{2020}]{zi2020wilddeepfake}
Zi, B., M.~Chang, J.~Chen, X.~Ma, and Y.G. Jiang 2020.
\newblock Wilddeepfake: A challenging real-world dataset for deepfake
  detection.
\newblock In {\em Proceedings of the 28th ACM international conference on
  multimedia}, pp.\  2382--2390.

\end{thebibliography}

\end{document}